%% file: main.tex
\documentclass{article}

\PassOptionsToPackage{numbers, compress}{natbib}

\usepackage[final]{neurips_2024}

\usepackage[utf8]{inputenc} 
\usepackage[T1]{fontenc}    
\usepackage{url}            
\usepackage{booktabs}       
\usepackage{amsfonts}       
\usepackage{nicefrac}       
\usepackage{microtype}      
\usepackage{xcolor}         
\usepackage{cancel}

\input{packages.tex}

\input{macros.tex}

\usepackage[hidelinks]{hyperref}       
\usepackage[capitalize]{cleveref}

\definecolor{ourblue}{rgb}{0.368,0.507,0.71}
\definecolor{ourgreen}{rgb}{0.56,0.692,0.195}
\definecolor{ourred}{rgb}{0.923,0.386,0.209}
\definecolor{url}{HTML}{d95225}
\hypersetup{
	final,
	colorlinks,
	linkcolor=ourred,
	citecolor=ourblue,
	urlcolor=url,
}

\title{The GAN is dead; long live the GAN!\\A Modern Baseline GAN}

\author{%
  Yiwen Huang\\
  Brown University \\
  \And
  Aaron Gokaslan \\
  Cornell University \\
  \And
  Volodymyr Kuleshov \\
  Cornell University \\
  \And
  James Tompkin \\
  Brown University \\
}

\begin{document}

\maketitle

\input{tex/abstract}
\input{tex/introduction}
\input{tex/loss}
\input{tex/roadmap}
\input{tex/experiments}
\input{tex/discussion}

\paragraph{Acknowledgements.}
The authors thank Xinjie Jayden Yi for contributing to the proof and Yu Cheng for helpful discussion. For compute, the authors thank Databricks Mosaic Research. Yiwen Huang was supported by a Brown University Division of Research Seed Award, and James Tompkin was supported by NSF CAREER 2144956. Volodymyr Kuleshov
was supported by NSF CAREER 2145577 and NIH MIRA 1R35GM15124301.

{\small
\bibliographystyle{neurips_2024}
\bibliography{main}
}

\appendix
\clearpage
\section*{Appendices}

\input{tex/appendix}

\clearpage
\input{tex/checklist}

\end{document}

%% file: packages.tex
\usepackage{booktabs}       
\usepackage{array}
\usepackage{multirow}

\usepackage{graphicx}
\usepackage{siunitx}
\usepackage{xspace}

\usepackage{todonotes}
\usepackage{comment}
\usepackage{amsmath} 
\usepackage{placeins}

\usepackage{amsfonts}       
\usepackage{nicefrac}       
\usepackage{microtype}      
\usepackage[super]{nth}
\usepackage{floatrow}

\usepackage[inline,shortlabels]{enumitem}
\usepackage{wrapfig}
\usepackage{tikz}
\usetikzlibrary{calc,spy}
\DeclareMathOperator\supp{supp}
\usepackage{cancel}
\usepackage{tabularray}
\usepackage{rotating}
\usepackage{pdflscape}
\usepackage{caption}
\usepackage{wrapfig}
\usepackage{tablefootnote}

\usepackage{adjustbox}
\usepackage[hang,flushmargin]{footmisc}

\setcitestyle{square}
\setcitestyle{citesep={,}}

\makeatletter
\DeclareRobustCommand\onedot{\futurelet\@let@token\@onedot}
\def\@onedot{\ifx\@let@token.\else.\null\fi\xspace}

\def\eg{\emph{e.g}\onedot} 
\def\ie{\emph{i.e}\onedot} 
 
\def\etc{\emph{etc}\onedot} 
\def\wrt{w.r.t\onedot} 
 
\def\etal{\emph{et al}\onedot}
\makeatother

\newfloatcommand{capbtabbox}{table}[][\FBwidth]

\newcommand{\upperRomannumeral}[1]{\uppercase\expandafter{\romannumeral#1}}

%% file: macros.tex
\newcommand{\modelName}{R3GAN\xspace}

%% file: tex/abstract.tex
\begin{abstract}
There is a widely-spread claim that GANs are difficult to train, and GAN architectures in the literature are littered with empirical tricks. We provide evidence against this claim and build a modern GAN baseline in a more principled manner. First, we derive a well-behaved regularized relativistic GAN loss that addresses issues of mode dropping and non-convergence that were previously tackled via a bag of ad-hoc tricks. We analyze our loss mathematically and prove that it admits local convergence guarantees, unlike most existing relativistic losses. Second, this loss allows us to discard all ad-hoc tricks and replace outdated backbones used in common GANs with modern architectures. Using StyleGAN2 as an example, we present a roadmap of simplification and modernization that results in a new minimalist baseline---\modelName (``Re-GAN''). Despite being simple, our approach surpasses StyleGAN2 on FFHQ, ImageNet, CIFAR, and Stacked MNIST datasets, and compares favorably against state-of-the-art GANs and diffusion models.\\
Code: \href{https://www.github.com/brownvc/R3GAN}{https://www.github.com/brownvc/R3GAN}
\end{abstract}

%% file: tex/introduction.tex
\section{Introduction}
\label{sec:intro}
Generative adversarial networks~(GANs) let us generate high-quality images in a single forward pass. 
However, the original objective in Goodfellow~\etal~\cite{gan}, is notoriously difficult to optimize due to its minimax nature. This leads to a fear that training might diverge at any point due to instability, and a fear that generated images might lose diversity through mode collapse. While there has been progress in GAN objectives~\cite{wgan-gp,rgan,rpgan,r1,r1r2}, practically, the effects of brittle losses are still regularly felt. This notoriety has had a lasting negative impact on GAN research.

A complementary issue---partly motivated by this instability---is that existing popular GAN backbones like StyleGAN~\cite{sg1,sg2,sg2ada,sg3} use many poorly-understood empirical tricks with little theory. 
For instance, StyleGAN uses a gradient penalized non-saturating loss~\cite{r1} to increase stability (affecting sample diversity), but then employs a minibatch standard deviation trick~\cite{pggan} to increase sample diversity.
Without tricks, the StyleGAN backbone still resembles DCGAN~\cite{dcgan} from 2015, yet it is still the common backbone of SOTA GANs such as GigaGAN~\cite{gigagan} and StyleGAN-T~\cite{sg-t}.
Advances in GANs have been conservative compared to other generative models such as diffusion models~\cite{ddpm,sde,edm,edm2}, where modern computer vision techniques such as multi-headed self attention~\cite{trans} and backbones such as preactivated ResNet~\cite{resnet2}, U-Net~\cite{unet} and vision transformers (ViTs)~\cite{vit} are the norm. 
Given outdated backbones, it is not surprising that there is a widely-spread belief that GANs do not scale in terms of quantitative metrics like Frechet Inception Distance~\cite{fid}. 

We reconsider this situation: we show that by combining progress in objectives into a regularized training loss, GANs gain improved training stability, which allows us to upgrade GANs with modern backbones. 
First, we propose a novel objective that augments the relativistic pairing GAN loss (RpGAN; \cite{rgan}) with zero-centered gradient penalties~\cite{r1,r1r2}, improving stability~\cite{wgan-gp,r1r2,r1}. 
We show mathematically that gradient-penalized RpGAN enjoys the same guarantee of local convergence as regularized classic GANs, and that removing our regularization scheme induces non-convergence. 

Once we have a well-behaved loss, none of the GAN tricks are necessary~\cite{pggan,sg2}, and we are free to engineer a modern SOTA backbone architecture. We strip StyleGAN of all its features, identify those that are essential, then borrow new architecture designs from modern ConvNets and transformers~\cite{convnext,metaformer}. Briefly, we find that proper ResNet design~\cite{resnet2,mobnet}, initialization~\cite{fixup}, and resampling~\cite{sg1,sg2,sg3,blurpool} are important, along with grouped convolution~\cite{resnext,xception} and no normalization~\cite{sg2,edm2,wgan-gp,esrgan,nfnet}. This leads to a design that is simpler than StyleGAN and improves FID performance for the same network capacity (2.75 vs.~3.78 on FFHQ-256).

In summary, our work first argues mathematically that GANs need not be tricky to train via an improved regularized loss.
Then, it empirically develops a simple GAN baseline that, without any tricks, compares favorably by FID to StyleGAN~\cite{sg1,sg2,sg3}, other SOTA GANs~\cite{biggan,vitgan,ddgan}, and diffusion models~\cite{ddpm,sde,lsgm} across FFHQ, ImageNet, CIFAR, and Stacked MNIST datasets. 

%% file: tex/loss.tex
\section{Serving Two Masters: Stability and Diversity with RpGAN \texorpdfstring{$+ R_1+R_2$}{R-1R-2}}
\label{sec:loss}

In defining a GAN objective, we tackle two challenges: stability and diversity. Some previous work deals with stability~\cite{sg1,sg2,sg3} and other previous work deals with mode collapse \cite{rgan}. To make progress in both, we combine a stable method with a simple regularizer that is grounded by theory.

\subsection{Traditional GAN}
A traditional GAN~\cite{gan,nowozin2016f} is formulated as a minimax game between a discriminator (or critic) $D_\psi$ and a generator $G_\theta$. Given real data $x\sim p_\mathcal{D}$ and fake data $x\sim p_\theta$ produced by $G_\theta$, the most general form of a GAN is given by:
\begin{align}
\begin{split}
\label{eq:gan}
\mathcal{L}(\theta,\psi)=\mathbb{E}_{z\sim p_z}\left[f\left(  D_\psi(G_\theta(z))\right)\right]+\mathbb{E}_{x\sim p_\mathcal{D}}\left[f\left( -D_\psi(x) \right)\right]
\end{split}
\end{align}
\noindent where $G$ tries to minimize $\mathcal{L}$ while $D$ tries to maximize it. The choice of $f$ is flexible~\cite{lsgan,hingegan}. In particular, $f(t) = -\log(1+e^{-t})$ recovers the classic GAN by Goodfellow~\etal~\cite{gan}. For the rest of this work, this will be our choice of $f$~\cite{nowozin2016f}.

It has been shown that Equation~\ref{eq:gan} has convex properties when $p_\theta$ can be optimized directly~\cite{gan,rpgan}. However, in practical implementations, the empirical GAN loss typically shifts fake samples beyond the decision boundary set by $D$, as opposed to directly updating the density function $p_\theta$. This deviation leads to a significantly more challenging problem, characterized by susceptibility to two prevalent failure scenarios: mode collapse/dropping\footnote{While mode collapse and mode dropping are technically distinct issues, they are used interchangeably in this context to describe the common problem where $\supp(p_\theta)$ does not comprehensively cover $\supp(p_\mathcal{D})$. Mode collapse refers to the generator producing a limited diversity of samples (i.e., one image for the entire distribution), whereas mode dropping involves the generator failing to represent certain modes of the data distribution (ignoring entire subsets of the training distribution).} and non-convergence.

\subsection{Relativistic \texorpdfstring{$f$-GAN}{f-GAN}}

We employ a slightly different minimax game named relativistic pairing GAN (RpGAN) by Jolicoeur-Martineau~\etal~\cite{rgan} to address mode dropping. The general RpGAN is defined as:
\begin{equation}
\label{eq:rpgan}
\mathcal{L}(\theta,\psi)=\mathbb{E}_{\substack{z\sim p_z\\x\sim p_\mathcal{D}}}\left[f\left(  D_\psi(G_\theta(z))-D_\psi(x) \right)\right]
\end{equation}
Although Eq.~\ref{eq:rpgan} differs only slightly from Eq.~\ref{eq:gan}, evaluating this critic difference has a fundamental impact on the landscape of $\mathcal{L}$. Since Eq.~\ref{eq:gan} merely requires $D$ to separate real and fake data, in the scenario where all real and fake data can be separated by a single decision boundary, the empirical GAN loss encourages $G$ to simply move all fake samples barely past this single boundary---this degenerate solution is what we observe as mode collapse/dropping. Sun~\etal~\cite{rpgan} characterize such degenerate solutions as bad local minima in the landscape of $\mathcal{L}$, and show that Eq.~\ref{eq:gan} has \emph{exponentially many} bad local minima. The culprit is the existence of a single decision boundary that naturally arises when real and fake data are considered in isolation. RpGAN introduces a simple solution by coupling real and fake data,~\ie a fake sample is critiqued by its realness \emph{relative to} a real sample, which effectively maintains a decision boundary in the neighborhood of \emph{each} real sample and hence forbids mode dropping. Sun~\etal~\cite{rpgan} show that the landscape of Eq.~\ref{eq:rpgan} contains no local minima that correspond to mode dropping solutions, and that every basin is a global minimum.

\subsection{Training Dynamics of RpGAN}
Although the RpGAN landscape result~\cite{rpgan} allows us to address mode dropping, the training dynamics of RpGAN have yet to be studied. The ultimate goal of Eq.~\ref{eq:rpgan} is to find the equilibrium $(\theta^*,\psi^*)$ such that $p_{\theta^*}=p_\mathcal{D}$ and $D_{\psi^*}$ is constant everywhere on $p_\mathcal{D}$. Sun~\etal~\cite{rpgan} show that $\theta^*$ is globally reachable along a non-increasing trajectory in the landscape of Eq.~\ref{eq:rpgan} under reasonable assumptions. However, the existence of such a trajectory does not necessarily mean that gradient descent will find it. Jolicoeur-Martineau~\etal show empirically that unregularized RpGAN does not perform well~\cite{rgan}. 

\vspace{1ex}
\noindent \textbf{Proposition~\upperRomannumeral{1}.} (Informal) \emph{Unregularized RpGAN does not always converge using gradient descent.}
\vspace{1ex}

\noindent We confirm this proposition with a proof in Appendix B. 
We show analytically that RpGAN does not converge for certain types of $p_\mathcal{D}$, such as ones that approach a delta distribution. Thus, further regularization is necessary to fill in the missing piece of a well-behaved loss.

\paragraph{Zero-centered gradient penalties.}
To tackle RpGAN non-convergence, we explore gradient penalties as the solution since it is proven that zero-centered gradient penalties (0-GP) facilitate convergent training for classic GANs~\cite{r1}. The two most commonly-used 0-GPs are $R_1$ and $R_2$:
\begin{equation}
\begin{aligned}
R_1(\psi)&=\frac{\gamma}{2}\mathbb{E}_{x\sim p_\mathcal{D}}\left[\left\| \nabla_x D_\psi \right \|^2\right]\\ 
R_2(\theta,\psi)&=\frac{\gamma}{2}\mathbb{E}_{x\sim p_\theta}\hspace{0.06cm}\left[\left\| \nabla_x D_\psi \right \|^2\right]
\end{aligned}
\end{equation}
$R_1$ penalizes the gradient norm of $D$ on real data, and $R_2$ penalizes the gradient norm of $D$ on fake data. Analysis on the training dynamics of GANs has thus far focused on local convergence~\cite{nagarajan2017gradient,gannum,r1},~\ie, whether the training at least converges when $(\theta,\psi)$ are in a neighborhood of $(\theta^*,\psi^*)$. In such a scenario, the convergence behavior can be analyzed~\cite{nagarajan2017gradient,gannum,r1} by examining the spectrum of the Jacobian of the gradient vector field $\left(-\nabla_\theta\mathcal{L},\nabla_\psi\mathcal{L} \right )$ at $(\theta^*,\psi^*)$. The key insight here is that when $G$ already produces the true distribution, we want $\nabla_x D=0$, so that $G$ is not pushed away from its optimal state, and thus the training does not oscillate. $R_1$ and $R_2$ impose such a constraint when $p_\theta=p_\mathcal{D}$. This also explains why earlier attempts at gradient penalties, such as the one-centered gradient penalty (1-GP) in WGAN-GP~\cite{wgan-gp}, fail to achieve convergent training~\cite{r1} as they still encourage $D$ to have a non-zero slope when $G$ has reached optimality.

Since the same insight also applies to RpGAN, 
we extend our previous analysis and show that:

\vspace{1ex}
\noindent \textbf{Proposition~\upperRomannumeral{2}.} (Informal) \emph{RpGAN with $R_1$ or $R_2$ regularization is locally convergent subject to similar assumptions as in} Mescheder~\etal~\cite{r1}.
\vspace{1ex}

In Appendix C, our proof similarly analyzes the eigenvalues of the Jacobian of the regularized RpGAN gradient vector field at $(\theta^*,\psi^*)$. We show that all eigenvalues have a negative real part; thus, regularized RpGAN is convergent in a neighborhood of $(\theta^*,\psi^*)$ for small enough learning rates~\cite{r1}.

\paragraph{Discussion.}
Another line of work~\cite{r1r2} links $R_1$ and $R_2$ to instance noise~\cite{instancenoise} as its analytical approximation. Roth et al.~\cite{r1r2} showed that for the classic GAN~\cite{gan} by Goodfellow~\etal, $R_1$ approximates convolving $p_\mathcal{D}$ with the density function of $\mathcal{N}(0, \gamma I)$, up to additional weighting and a Laplacian error term. $R_2$ likewise approximates convolving $p_\theta$ with $\mathcal{N}(0, \gamma I)$ up to similar error terms. The Laplacian error terms from $R_1$, $R_2$ cancel when $D_\psi$ approaches $D_{\psi^*}$. We do not extend Roth~\etal's proof~\cite{r1r2} to RpGAN; however, this approach might provide complimentary insights to our work, which follows the strategy of Mescheder~\etal~\cite{r1}.

\subsection{A Practical Demonstration}


We experiment with how well-behaved our loss is on StackedMNIST~\cite{pacgan} which consists of 1000 uniformly-distributed modes. The network is a small ResNet~\cite{resnet2} for $G$ and $D$ without any normalization layers~\cite{bn,gn,ln,in}.
Through the use of a pretrained MNIST classifier, we can explicitly measure how many modes of $p_\mathcal{D}$ are recovered by $p_\theta$. Furthermore, we can estimate the reverse KL divergence between the fake and real samples $D_\text{KL}\left(p_\theta\parallel p_\mathcal{D} \right)$ via the KL divergence between the categorical distribution of $p_\theta$ and the true uniform distribution.
\begin{figure}
\begin{floatrow}
\ffigbox{%
  \centering
  \includegraphics[width=0.48\textwidth]{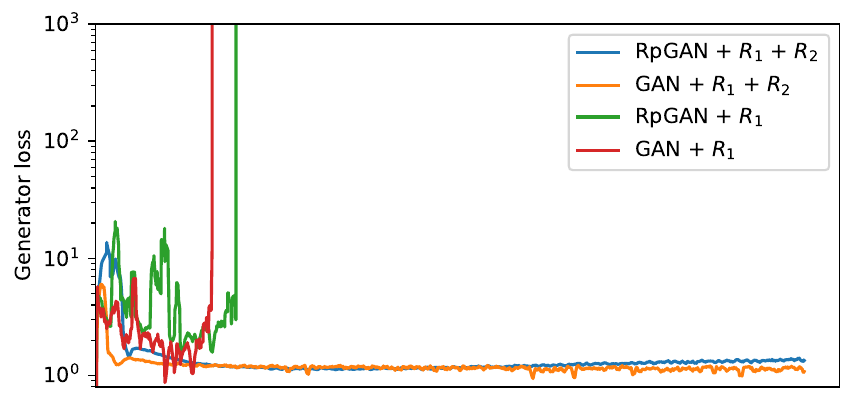}%
}{%
\caption{Generator $G$ loss for different objectives over training. Regardless of which objective is used, training diverges with only $R_1$ and succeeded with both $R_1$ and $R_2$. Convergence failure with only $R_1$ was noted by Lee et al.~\cite{vitgan}.}
\label{fig:mnist_loss_curve}
}
\capbtabbox{%
\centering
    \begin{tabular}{ l rrr }
        \toprule
        Loss & \# modes$\uparrow$ & $D_\text{KL}$$\downarrow$ \\
        \midrule
        RpGAN $+ R_1+R_2$ & $\mathbf{1000}$ & $\mathbf{0.0781}$ \\
        GAN $+ R_1+R_2$ & $693$ & $0.9270$ \\
        RpGAN $+ R_1$ & Fail & Fail \\
        GAN $+ R_1$ & Fail & Fail \\
        \bottomrule
    \end{tabular}
}{%
    \caption{StackedMNIST~\cite{pacgan} result for each loss function. The maximum possible mode coverage is 1000. ``Fail'' indicates that training diverged early on.}
    \label{tab:loss}
}
\end{floatrow}
\end{figure}

A conventional GAN loss with $R_1$, as used by Mescheder et al.~\cite{r1} and the StyleGAN series~\cite{sg1, sg2, sg3}, diverges quickly (Fig.~\ref{fig:mnist_loss_curve}). Next, while theoretically sufficient for local convergence, RpGAN with only $R_1$ regularization is also unstable and diverges quickly\footnote{Varying $\gamma$ from 0.1 to 100 does not stabilize training.}. In each case, the gradient of $D$ on fake samples explodes when training diverges. With both $R_1$ and $R_2$, training becomes stable for both the classic GAN and RpGAN. Now stable, we can see that the classic GAN suffers from mode dropping, whereas RpGAN achieves full mode coverage (Tab.~\ref{tab:loss}) and reduces $D_\text{KL}$ from 0.9270 to 0.0781. As a point of contrast, StyleGAN~\cite{sg1,sg2,sg2ada,sg3} uses the minibatch standard deviation trick to reduce mode dropping, improving mode coverage from 857 to 881 on StackedMNIST\footnote{These numbers are from Karras~\etal~\cite{pggan}, Table 4. "857" corresponds to a low-capacity version of a progressive GAN and "881" adds the minibatch standard deviation trick. Further comparisons via loss curves are difficult since progressive GAN is a substantially different model than the small ResNet we use for this experiment.} and with barely any improvement on $D_\text{KL}$~\cite{pggan}.

$R_1$ alone is not sufficient for globally-convergent training. While a theoretical analysis of this is difficult, our small demonstration still provides insights into the assumptions of our convergence proof. 
In particular, the assumption that $(\theta,\psi)$ are sufficiently close to $(\theta^*,\psi^*)$ is highly unlikely early in training. In this scenario, if $D$ is sufficiently powerful, regularizing $D$ solely on real data is not likely to have much effect on $D$'s behavior on fake data and so training can fail due to an ill-behaved $D$ on fake data. This observation has been made by previous studies~\cite{r1gradexp, r1gradexpcvpr} specifically for empirical GAN training, that regularizing an empirical discriminator with only $R_1$ leads to gradient explosion on fake data due to the memorization of real samples.


Thus, the practical solution is to regularize $D$ on both real and fake data. The benefit of doing so can be viewed from the insight of Roth~\etal~\cite{r1r2}: that applying $R_1$ and $R_2$ in conjunction smooths both $p_\mathcal{D}$ and $p_\theta$ which makes learning easier than only smoothing $p_\mathcal{D}$. We also find empirically that with both $R_1$ and $R_2$ in place, $D$ tends to satisfy $\mathbb{E}_{x\sim p_\mathcal{D}}\left[\left\| \nabla_x D \right \|^2\right]\approx\mathbb{E}_{x\sim p_\theta}\left[\left\| \nabla_x D \right \|^2\right]$ even early in the training. Jolicoeur-Martineau~\etal~\cite{ganmmc} show that in this case $D$ becomes a maximum margin classifier---but if only one regularization term is applied, this does not hold. Additionally, having roughly the same gradient norm on real and fake data potentially reduces discriminator overfitting, as Fang~\etal~\cite{diggan} observe that the gradient norm on real and fake data diverges when $D$ starts to overfit.


%% file: tex/roadmap.tex

\section{A Roadmap to a New Baseline --- \modelName}
\label{sec:roadmap}



The well-behaved RpGAN + $R_1$ + $R_2$ loss alleviates GAN optimization problems, and lets us proceed to build a minimalist baseline---\modelName---with recent network backbone advances in mind~\cite{convnext,metaformer}. Rather than simply state the new approach, we will draw out a roadmap from the StyleGAN2 baseline~\cite{sg2ada}. This model (Config A; identical to~\cite{sg2ada}) consists of a VGG-like~\cite{vgg} backbone for $G$, a ResNet $D$, a few techniques that facilitate style-based generation, and many tricks that serve as patches to the weak backbone. Then, we remove all non-essential features of StyleGAN2 (Config B), apply our loss function (Config C), and gradually modernize the network backbone (Config D-E).

We evaluate each configuration on FFHQ $256\times256$~\cite{sg1}. Network capacity is kept roughly the same for all configurations---both $G$ and $D$ have about 25\ M trainable parameters. Each configuration is trained until $D$ sees 5\ M real images. We inherit training hyperparameters (\eg, optimizer settings, batch size, EMA decay length) from Config A unless otherwise specified. We tune the training hyperparameters for our final model and show the converged result in Sec.~\ref{sec:exp}.


\vspace{-0.3cm}
\paragraph{Minimum baseline (Config B).}
\input{tex/table_stylegan_roadmap}

We strip away all StyleGAN2 features, retaining only the raw network backbone and basic image generation capability. The features fall into three categories:
\begin{itemize}[leftmargin=10pt,itemsep=0pt,topsep=0pt]
\item Style-based generation: mapping network~\cite{sg1}, style injection~\cite{sg1}, weight modulation/demodulation~\cite{sg2}, noise injection~\cite{sg1}.
\end{itemize}\quad 
\begin{itemize}[leftmargin=10pt,itemsep=0pt,topsep=0pt]
\vspace{-0.25cm} 
\item Image manipulation enhancements: mixing regularization~\cite{sg1}, path length regularization~\cite{sg2}.
\item Tricks: $z$ normalization~\cite{pggan}, minibatch stddev~\cite{pggan}, equalized learning rate~\cite{pggan}, lazy regularization~\cite{sg2}.
\end{itemize}

Following~\cite{sgxl,sg-t}, we reduce the dimension of $z$ to 64. The absence of equalized learning rate necessitates a lower learning rate, reduced from 2.5$\times$10\textsuperscript{-3} to 5$\times$10\textsuperscript{-5}. Despite a higher FID of 12.46 than Config-A, this simplified baseline produces reasonable sample quality and stable training. We compare this with DCGAN~\cite{dcgan}, an early attempt at image generation. Key differences include:
\begin{enumerate}[label=\alph*), noitemsep,topsep=0pt,leftmargin=24pt]
\item Convergent training objective with $R_1$ regularization.\label{item:convergent} 
\item Smaller learning rate, avoiding momentum optimizer (Adam $\beta_1=0$).\label{item:learning_rate} 
\item No normalization layer in $G$ or $D$.\label{item:normalization} 
\item Proper resampling via bilinear interpolation instead of strided (transposed) convolution.\label{item:resampling} 
\item Leaky ReLU in both $G$ and $D$, no tanh in the output layer of $G$.\label{item:activation} 
\item 4$\times$4 constant input for $G$, output skips for $G$, ResNet $D$.\label{item:input} 
\end{enumerate}

\textbf{Experimental findings from StyleGAN.} Violating \ref{item:convergent}, \ref{item:learning_rate}, or \ref{item:normalization} often leads to training failures.
Gidel~\etal~\cite{ganmomentum} show that \emph{negative} momentum can improve GAN training dynamics. Since optimal negative momentum is another challenging hyperparameter, we do not use any momentum to avoid worsening GAN training dynamics. Studies suggest normalization layers harm generative models~\cite{sg2,edm2}. Batch normalization~\cite{bn} often cripples training due to dependencies across multiple samples, and is incompatible with $R_1$, $R_2$, or RpGAN that assume independent handling of each sample. Weaker data-independent normalizations~\cite{sg2,edm2} might help; we leave this for future work. Early GANs may succeed despite violating \ref{item:convergent} and \ref{item:normalization}, possibly constituting a full-rank solution~\cite{r1} to Eq.~\ref{eq:gan}.

Violations of \ref{item:resampling} or \ref{item:activation} do not significantly impair training stability but negatively affect sample quality. Improper transposed convolution can cause checkerboard artifacts, unresolved even with subpixel convolution~\cite{subpixel} or carefully tuned transposed convolution unless a low-pass filter is applied. Interpolation methods avoid this issue, varying from nearest neighbor~\cite{pggan} to Kaiser filters~\cite{sg3}. We use bilinear interpolation for simplicity. For activation functions, smooth approximations of (leaky) ReLU, such as Swish~\cite{swish}, GELU~\cite{gelu}, and SMU~\cite{smu}, worsen FID. PReLU~\cite{prelu} marginally improves FID but increases VRAM usage, so we use leaky ReLU.

All subsequent configurations adhere to \ref{item:convergent} through \ref{item:activation}. Violation of \ref{item:input} is acceptable as it pertains to the network backbone of StyleGAN2~\cite{sg2}, modernized in Config D and E.

\vspace{-0.3cm}
\paragraph{Well-behaved loss function (Config C).}
We use the loss function proposed in Section~\ref{sec:loss} and this reduces FID to 11.65. We hypothesize that the network backbone in Config B is the limiting factor. 

\begin{figure*}[t]%
\centering\footnotesize%
 \includegraphics[width=\linewidth,clip,trim={3.5cm 11.5cm 0cm 0cm}]{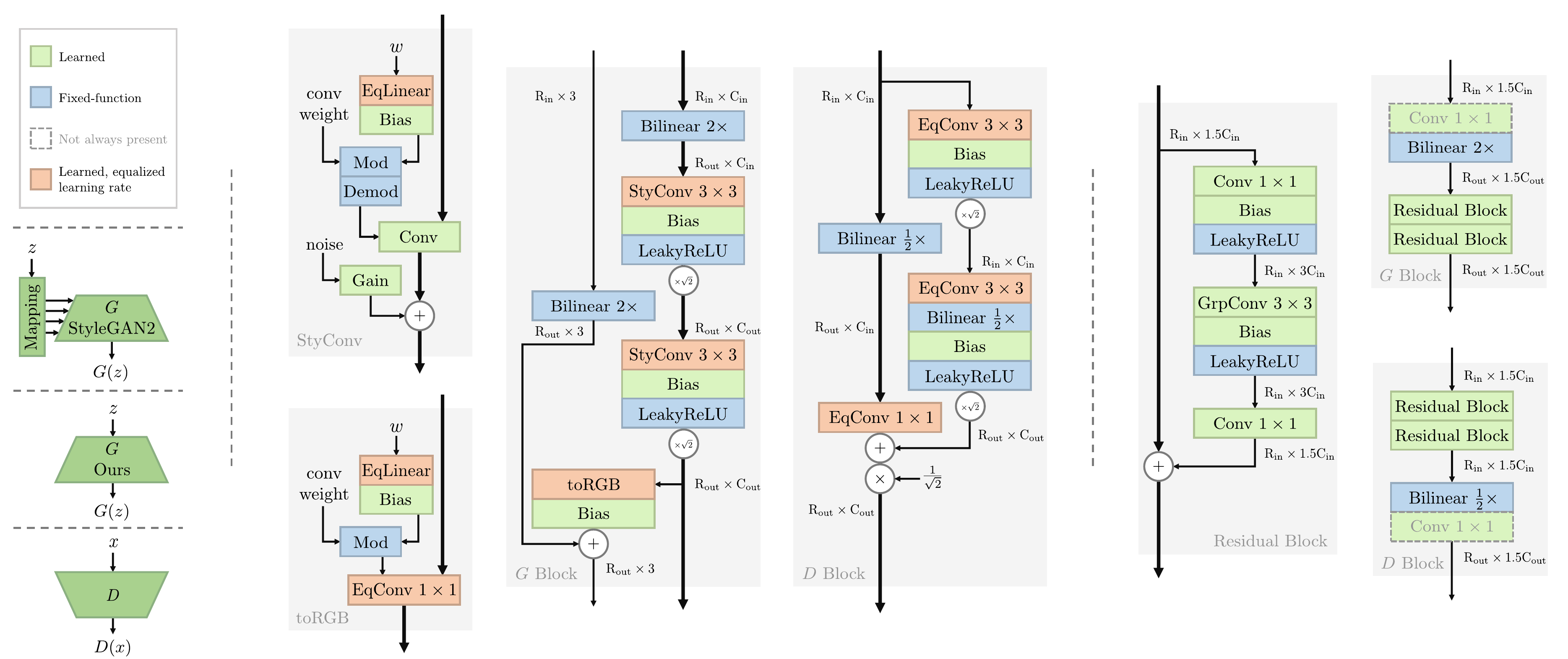}\\%
\makebox[0.144\linewidth]{(a) Overall view}\hfill%
\makebox[0.562\linewidth]{(b) StyleGAN2 architecture blocks~\cite{sg2} (Config A)}\hfill%
\makebox[0.294\linewidth]{(c) Ours (Config E)}\hfill%
\vspace{0.5mm}
\caption{\label{fig:network}%
\textbf{Architecture comparison.}
For image generation, $G$ and $D$ are often both deep ConvNets with either partially or fully symmetric architectures.
\textbf{(a)}
StyleGAN2~\cite{sg2} $G$ uses a network to map noise vector $z$ to an intermediate style space $\mathcal{W}$. We use a traditional generator as style mapping is not necessary for a minimal working model.
\textbf{(b)}
StyleGAN2's building blocks have intricate layers but are themselves simple, with a ConvNet architecture from 2015~\cite{alexnet,vgg,resnet}. ResNet's identity mapping principle is also violated in the discriminator.
\textbf{(c)}
We remove tricks and modernize the architecture. Our design has clean layers with a more powerful ConvNet architecture.
}%
\vspace{-0.3cm}%
\end{figure*}

\vspace{-0.3cm}
\paragraph{General network modernization (Config D).}
First, we apply the 1-3-1 bottleneck ResNet architecture~\cite{resnet,resnet2} to both $G$ and $D$. This is the direct ancestor of all modern vision backbones~\cite{convnext,metaformer}. 
We also incorporate principles discovered in Config B and various modernization efforts from ConvNeXt~\cite{convnext}. We categorize the roadmap of ConvNeXt as follows:
\begin{enumerate}[label=\roman*., itemsep=1pt,leftmargin=12pt,topsep=0pt,parsep=1pt]
    \item Consistently beneficial: \begin{enumerate*}[label=\theenumi\arabic*), ref=\arabic*, before=\unskip{ }, itemjoin={{, }}, itemjoin*={{, and }}]
        \item\label{item:i1} increased width with depthwise convolution
        \item\label{item:i2} inverted bottleneck
        \item\label{item:i3} fewer activation functions
        \item\label{item:i4} separate resampling layers.
    \end{enumerate*}
    \item Negligible performance gain: \begin{enumerate*}[label=\theenumi\arabic*), ref=\arabic*, before=\unskip{ }, itemjoin={{, }}, itemjoin*={{, and }}]
        \item\label{item:ii1} large kernel depthwise conv.~with fewer channels
        \item\label{item:ii2} swap ReLU with GELU
        \item\label{item:ii3} fewer normalization layers
        \item\label{item:ii4} swap batch norm.~with layer norm.
    \end{enumerate*}
    \item Irrelevant to our setting: \begin{enumerate*}[label=\theenumi\arabic*), ref=\arabic*, before=\unskip{ }, itemjoin={{, }}, itemjoin*={{, and }}]
        \item\label{item:iii1} \hspace{-0.1cm}improved training recipe
        \item\label{item:iii2} \hspace{-0.1cm}stage ratio
        \item\label{item:iii3} \hspace{-0.1cm}`patchify' stem.
    \end{enumerate*}
\end{enumerate}

We aim to apply i) to our model, specifically i.\ref{item:i3} and i.\ref{item:i4} for the classic ResNet, while reserving i.\ref{item:i1} and i.\ref{item:i2} for Config E. Many aspects of ii) were introduced merely to mimic vision transformers~\cite{swin,vit} without yielding significant improvements~\cite{convnext}. ii.\ref{item:ii3} and ii.\ref{item:ii4} are inapplicable due to our avoidance of normalization layers following principle \ref{item:normalization}. ii.\ref{item:ii2} contradicts our finding that GELU deteriorates GAN performance, thus we use leaky ReLU per principle \ref{item:activation}. Liu~\etal emphasize large conv.~kernels (ii.\ref{item:ii1})~\cite{convnext}, but this results in slightly worse performance compared to wider 3$\times$3 conv.~layers, so we do not adopt this ConvNeXt design choice.

\paragraph{Neural network architecture details.} Given i.\ref{item:i3}, i.\ref{item:i4}, and principles \ref{item:normalization}, \ref{item:resampling}, and \ref{item:activation}, we can replace the StyleGAN2 backbone with a modernized ResNet. We use a fully symmetric design for $G$ and $D$ with 25\ M parameters each, comparable to Config-A. The architecture is minimalist: each resolution stage has one transition layer and two residual blocks. The transition layer consists of bilinear resampling and an optional 1$\times$1 conv.~for changing spatial size and feature map channels. The residual block includes five operations: Conv1$\times$1$\rightarrow$ Leaky ReLU $\rightarrow$ Conv3$\times$3$\rightarrow$ Leaky ReLU $\rightarrow$ Conv1$\times$1, with the final Conv1$\times$1 having no bias term. For the 4$\times$4 resolution stage, the transition layer is replaced by a basis layer for $G$ and a classifier head for $D$. The basis layer, similar to StyleGAN~\cite{sg1,sg2}, uses 4$\times$4 learnable feature maps modulated by $z$ via a linear layer. The classifier head uses a global 4$\times$4 depthwise conv.~to remove spatial extent, followed by a linear layer to produce $D$'s output. We maintain the width ratio for each resolution stage as in Config A, making the stem width 3$\times$ as wide due to the efficient 1$\times$1 conv. The 3$\times$3 conv.~in the residual block has a compression ratio of 4, following~\cite{resnet,resnet2}, making the bottleneck width 0.75$\times$ as wide as Config A.
To avoid variance explosion due to the lack of normalization, we employ fix-up initialization~\cite{fixup}: 
We zero-initialize the last convolutional layer in each residual block and scale down the initialization of the other two convolutional layers in the block by $L^{-0.25}$, where $L$ is the number of residual blocks. We avoid other fix-up tricks, such as excessive bias terms and a learnable multiplier.







\paragraph{Bottleneck modernization (Config E).}
Now that we have settled on the overall architecture, we investigate how the residual block can be modernized, specifically i.\ref{item:i1}) and i.\ref{item:i2}). First, we explore i.\ref{item:i1} and replace the 3$\times$3 convolution in the residual block with a grouped convolution. We set the group size to 16 rather than 1 (\ie depthwise convolution as in ConvNeXt) as depthwise convolution is highly inefficient on GPUs and is not much faster than using a larger group size. With grouped convolution, we can reduce the bottleneck compression ratio to two given the same model size. This increases the width of the bottleneck to 1.5$\times$ as wide as Config A. 
Finally, we notice that the compute cost of grouped convolution is negligible compared to 1$\times$1 convolution, and so we seek to enhance the capacity of grouped convolution. We apply i.\ref{item:i2}), which inverts the bottleneck width and the stem width, and which doubles the width of grouped convolutions without any increase in model size. Figure~\ref{fig:network} depicts our final design, which reflects modern CNN architectures.

%% file: tex/table_stylegan_roadmap.tex
\begin{wraptable}[21]{r}{8cm}
\vspace{-0.5cm}
\resizebox{1\linewidth}{!}{
\begin{tabular}{ l r c c c } 
\toprule
   & \multicolumn{1}{l}{Configuration}  & FID$\downarrow$                    & G \#params              & D \#params               \\ 
\midrule
A  & \multicolumn{1}{l}{StyleGAN2}  & 7.516                  & 24.767M                  & 24.001M                   \\
\midrule
B  & \multicolumn{1}{l}{Stripped StyleGAN2}                                                                                                                                                                                                                                                                                                                                                                                                                 &                        &                          &                           \\ 
   & \begin{tabular}[c]{@{}r@{}}\textcolor{red}{- $z$ normalization}\\ \textcolor{red}{- Minibatch stddev}\\ \textcolor{red}{- Equalized learning rate}\textcolor{red}{}\\\textcolor{red}{- Mapping network}\\ \textcolor{red}{- Style injection}\\ \textcolor{red}{- Weight mod / demod}\\ \textcolor{red}{- Noise injection}\\ \textcolor{red}{- Mixing regularization}\\ \textcolor{red}{- Path length regularization}\\ \textcolor{red}{- Lazy regularization}\textcolor{blue}{}\end{tabular} & \multirow{2}{*}{12.46} & \multirow{2}{*}{18.890M} & \multirow{2}{*}{23.996M}  \\ 
\midrule
C  & \multicolumn{1}{l}{Well-behaved Loss}                       &                        &                          &                           \\ 
 & \textcolor[rgb]{0,0.502,0.502}{+ RpGAN loss}                                                                                                                                                                                                                                                                                                                                                                                                                   & 11.77                                           & \multirow{2}{*}{18.890M} & \multirow{2}{*}{23.996M}                           \\ 
   & \textcolor[rgb]{0,0.502,0.502}{+ $R_2$ gradient penalty}                                                                                                                                                                                                                                                                                                                                                                                                & \multirow{1}{*}{11.65} &                          &                           \\ 
\midrule
D  & \multicolumn{1}{l}{ConvNeXt-ify pt.~1}                                                                                                                                                                                                                                                                                                                                                                                                                 &                        &                          &                           \\ 
 & \textcolor[rgb]{0,0.502,0.502}{+ ResNet-ify G $\&$ D}                                                                                                                                                                                                                                                                                                                                                                                                   & 10.17                  & 23.400M                  & \multirow{2}{*}{23.282M}  \\
   & \textcolor{red}{- Output skips}                                                                                                                                                                                                                                                                                                                                                                                                                         & 9.950 & 23.378M &                           \\ 
\midrule
E  & \multicolumn{1}{l}{ConvNeXt-ify pt.~2}                              &                        &                          &                           \\
 & \textcolor[rgb]{0,0.502,0.502}{+ ResNeXt-ify G $\&$ D}                                                                                                                                                                                                                                                                                                                                                                                                  & 7.507                  & 23.188M                  & 23.091M                   \\ 
   & \textcolor[rgb]{0,0.502,0.502}{+ Inverted bottleneck}                                                                                                                                         & 7.045 & 23.058M & 23.010M  \\ 
\bottomrule
\end{tabular}
}
\vspace{-0.25cm}
\caption{Effect of our simplification and modernization efforts evaluted on FFHQ-256.} 
\label{tab:roadmap}
\end{wraptable}

%% file: tex/experiments.tex
\vspace{-0.2cm}
\section{Experiments Details}
\label{sec:exp}

\vspace{-0.2cm}
\subsection{Roadmap Insights on FFHQ-256\texorpdfstring{~\cite{sg1}}{}}
\label{sub:arc-experiments}
\vspace{-0.1cm}
As per Table~\ref{tab:roadmap}, Config A (vanilla StyleGAN2) achieves an FID of 7.52 using the official implementation on FFHQ-256. Config B with all tricks removed achieves an FID of 12.46---performance drops as expected. 
Config C, with a well-behaved loss, achieves an FID of 11.65. But, now training is sufficiently stable to improve the architecture.

Config D, which improves $G$ and $D$ based on the classic ResNet and ConvNeXt findings, achieves an FID of 9.95. The output skips of the StyleGAN2 generator are no longer useful given our new architecture; including them produces a worse FID of 10.17. Karras~\etal find that the benefit of output skips is mostly related to gradient magnitude dynamics~\cite{sg3}, and this has been addressed by our ResNet architecture. For StyleGAN2, Karras~\etal conclude that a ResNet architecture is harmful to $G$~\cite{sg2}, but this is not true in our case as their ResNet implementation is considerably different from ours: 1) Karras~\etal use one 3-3 residual block for each resolution stage, while we have a separate transition layer and two 1-3-1 residual blocks; 2) i.3) and i.4) are violated as they do not have a linear residual block~\cite{mobnet} and the transition layer is placed on the skip branch of the residual block rather than the stem; 3) the essential principle of ResNet~\cite{resnet}---identity mapping~\cite{resnet2}---is violated as Karras~\etal divide the output of the residual block by $\sqrt{2}$ to avoid variance explosion due to the absence of a proper initialization scheme.

For Config E, we conduct two experiments that ablate i.\ref{item:i1} (increased width with depthwise conv.) and i.\ref{item:i2} (an inverted bottleneck). We add GroupedConv and reduce the bottleneck compression ratio to two given the same model size. Each bottleneck is now 1.5$\times$ the width of Config A, and the FID drops to 7.51, surpassing the performance of StyleGAN2. By inverting the stem and the bottleneck dimensions to enhance the capacity of GroupedConv, our final model achieves an FID of 7.05, exceeding StyleGAN2.

\begin{wraptable}[12]{r}{6.5cm}
\vspace{-1.25cm}
\centering
\caption{StackedMNIST 1000-mode coverage.}
\vspace{-0.4cm}
\resizebox{0.8\linewidth}{!}{
\begin{tblr}{
  cell{2}{2} = {c},
  cell{2}{3} = {c},
  cell{3}{2} = {c},
  cell{3}{3} = {c},
  cell{4}{2} = {c},
  cell{4}{3} = {c},
  cell{5}{2} = {c},
  cell{5}{3} = {c},
  cell{6}{2} = {c},
  cell{6}{3} = {c},
  cell{7}{2} = {c},
  cell{7}{3} = {c},
  cell{8}{2} = {c},
  cell{8}{3} = {c},
  cell{9}{2} = {c},
  cell{9}{3} = {c},
  cell{10}{2} = {c},
  cell{10}{3} = {c},
  cell{11}{2} = {c},
  cell{11}{3} = {c},
  cell{12}{2} = {c},
  cell{12}{3} = {c},
  hline{2,12} = {1-3}{},
}
Model     & \# modes$\uparrow$ & $D_\text{KL}$$\downarrow$            &  \\
DCGAN~\cite{dcgan}     & 99            & 3.40\phantom{0}&  \\
VEEGAN~\cite{srivastava2017veegan}    & 150           & 2.95\phantom{0}&  \\
WGAN-GP~\cite{wgan-gp}& 959           & 0.73\phantom{0}&  \\
PacGAN~\cite{pacgan}    & 992           & 0.28\phantom{0}&  \\
StyleGAN2~\cite{sg2} & 940           & 0.42\phantom{0}&  \\
PresGAN~\cite{presgan}   & \textbf{1000} & 0.12\phantom{0}&  \\
Adv. DSM~\cite{advsm}  & \textbf{1000} & 1.49\phantom{0}&  \\
VAEBM~\cite{vaebm}     & \textbf{1000} & 0.087          &  \\
DDGAN~\cite{ddgan}     & \textbf{1000} & 0.071          &  \\
MEG~\cite{meg}       & \textbf{1000} & 0.031          &  \\
Ours---Config E     & \textbf{1000} & \textbf{0.029} &  
\end{tblr}
}
\label{tab:stackedmnist}
\end{wraptable}%

\subsection{Mode Recovery --- StackedMNIST\texorpdfstring{~\cite{metz2016unrolled}}{}} 
\vspace{-0.1cm}
We repeat the earlier experiment in 1000-mode convergence on StackedMNIST (unconditional generation), but this time with our updated architecture and with comparisons to SOTA GANs and likelihood-based methods (Tab.~\ref{tab:stackedmnist}, Fig.~\ref{fig:stacked-mnist}). 
One advantage brought up of likelihood-based models such as diffusion over GANs is that they achieve mode coverage~\cite{adm}. We find that most GANs struggle to find all modes. But, PresGAN~\cite{presgan}, DDGAN~\cite{ddgan}, and our approach are successful. Further, our method outperforms all other tested GAN models in term of KL divergence.

\subsection{FID --- FFHQ-256\texorpdfstring{~\cite{sg1}}{} (Optimized)}
\vspace{-0.1cm}
We train Config E model until convergence and with optimized hyperparameters and training schedule on FFHQ at 256$\times$256 (unconditional generation) (Tab.~\ref{tab:ffhq256}, Figs.~\ref{fig:ffhq-256-teaser} and~\ref{fig:ffhq-256}). 
Please see our supplemental material for training details.
Our model outperforms existing StyleGAN methods, plus four more recent diffusion-based methods. On this common dataset experimental setting, many methods (not listed here) use the bCR~\cite{zhao2021improved} trick---this has only been shown to improve performance on FFHQ-256 (not even at different resolutions of FFHQ)~\cite{zhao2021improved, zhang2022styleswin}. We do not use this trick. 

\subsection{FID --- FFHQ-64\texorpdfstring{~\cite{edm}}{}}
\vspace{-0.1cm}
To compare with EDM~\cite{edm} directly, we evaluate our model on FFHQ at 64$\times$64 resolution. For this, we remove the two highest resolution stages of our 256$\times$256 model, resulting in a generator that is less than half the number of parameters as EDM. Despite this, our model outperforms EDM on this dataset and needs one function evaluation only (Tab.~\ref{tab:ffhq64}).

\begin{figure}
\begin{floatrow}
    \capbtabbox{%
        \centering
        \resizebox{\linewidth}{!}{
        \begin{tblr}{
          column{2,3} = {r},
          cell{1}{2} = {c},
          cell{1}{3} = {c},
          hline{2,5,9,10} = {-}{},
        }
        Model       & NFE$\downarrow$ & FID$\downarrow$  \\
        StyleGAN2~\cite{sg2}   & 1               & 3.78 \\
        StyleGAN3-T~\cite{sg3} & 1               & 4.81 \\
        StyleGAN3-R~\cite{sg3} & 1               & 3.92 \\
        LDM~\cite{rombach2022high} & 200               & 4.98\\
        ADM (DDIM)~\cite{adm,compdiff} & 500               & 8.41\\
        ADM (DPM-Solver)~\cite{adm,compdiff} & 500               & 8.40\\
        Diffusion Autoencoder~\cite{diffae,compdiff} & 500               & 5.81\\
        Ours---Config E  & 1               & 2.75 \\
        \emph{With ImageNet feature leakage~\cite{kynkaanniemi2022role}:} & & \\
        PolyINR*~\cite{singh2023polynomial} & 1               & 2.72 \\
        StyleGAN-XL*~\cite{sgxl} & 1               & 2.19 \\
        StyleSAN-XL*~\cite{takida2024san} & 1               & 1.68 \\
        \end{tblr}
        }
    }{%
        \caption{
        \label{tab:ffhq256}FFHQ-256. * denotes models that leak ImageNet features.}
    }
    \capbtabbox{%
        \centering
        \resizebox{0.85\linewidth}{!}{
        \begin{tblr}{
          column{2} = {r},
          column{3} = {r},
          hline{2,5,8} = {-}{},
        }
        Model         & NFE$\downarrow$ & FID$\downarrow$ \\
        StyleGAN2~\cite{sg2,anycostgan}     & 1               & 3.32            \\
        MSG-GAN~\cite{karnewar2020msg,anycostgan}       & 1               & 2.7             \\
        Anycost GAN~\cite{anycostgan}   & 1               & 2.52            \\
        VE~\cite{sde,edm}            & 79              & 25.95           \\
        VP~\cite{sde,edm}            & 79              & 3.39            \\
        EDM~\cite{edm}           & 79              & 2.39            \\
        Ours—Config E & 1               & 1.95 \\
        \end{tblr}
        }
    }{%
        \caption{\label{tab:ffhq64}FFHQ-64.}
    }
\end{floatrow}
\vspace{-0.25cm}
\end{figure}

\subsection{FID --- CIFAR-10~\cite{krizhevsky2009learning}} \vspace{-0.1cm}

\begin{wraptable}[14]{r}{6.5cm}
\vspace{-0.75cm}
\centering
\caption{\label{tab:cifar10}CIFAR-10 performance.}
\vspace{-0.4cm}
\resizebox{0.9\linewidth}{!}{
    \begin{tblr}{
          column{2,3} = {r},
          cell{1}{2} = {c},
          cell{1}{3} = {c},
          hline{2,9,13} = {-}{},
        }
        Model               & NFE$\downarrow$ & FID$\downarrow$ \\
        BigGAN~\cite{biggan}              & 1               & 14.73 \\
        TransGAN~\cite{trans}            & 1               & 9.26 \\
        ViTGAN~\cite{vitgan}              & 1               & 6.66 \\
        DDGAN~\cite{ddgan}               & 4               & 3.75 \\
        Diffusion StyleGAN2~\cite{diffusiongan} & 1               & 3.19 \\
        StyleGAN2 + ADA~\cite{sg2ada}     & 1               & 2.42 \\
        StyleGAN3-R + ADA~\cite{sg3,studio}   & 1               & 10.83 \\
        DDPM~\cite{ddpm}               & 1000            & 3.21 \\
        DDIM~\cite{ddim}                & 50             & 4.67 \\
        VE~\cite{sde,edm}                  & 35              & 3.11 \\
        VP~\cite{sde,edm}                  & 35              & 2.48 \\
        Ours---Config E     & 1               & 1.96 \\
        \hline
        \emph{With ImageNet feature leakage~\cite{kynkaanniemi2022role}:} & & \\
        StyleGAN-XL*~\cite{sgxl}       & 1               & 1.85 \\
        \end{tblr}
}
\end{wraptable}

\begin{wrapfigure}[12]{r}{6.5cm}
    \vspace{-0.4cm}
    \centering
    \includegraphics[width=\linewidth,clip,trim={0 0 0 2cm}]{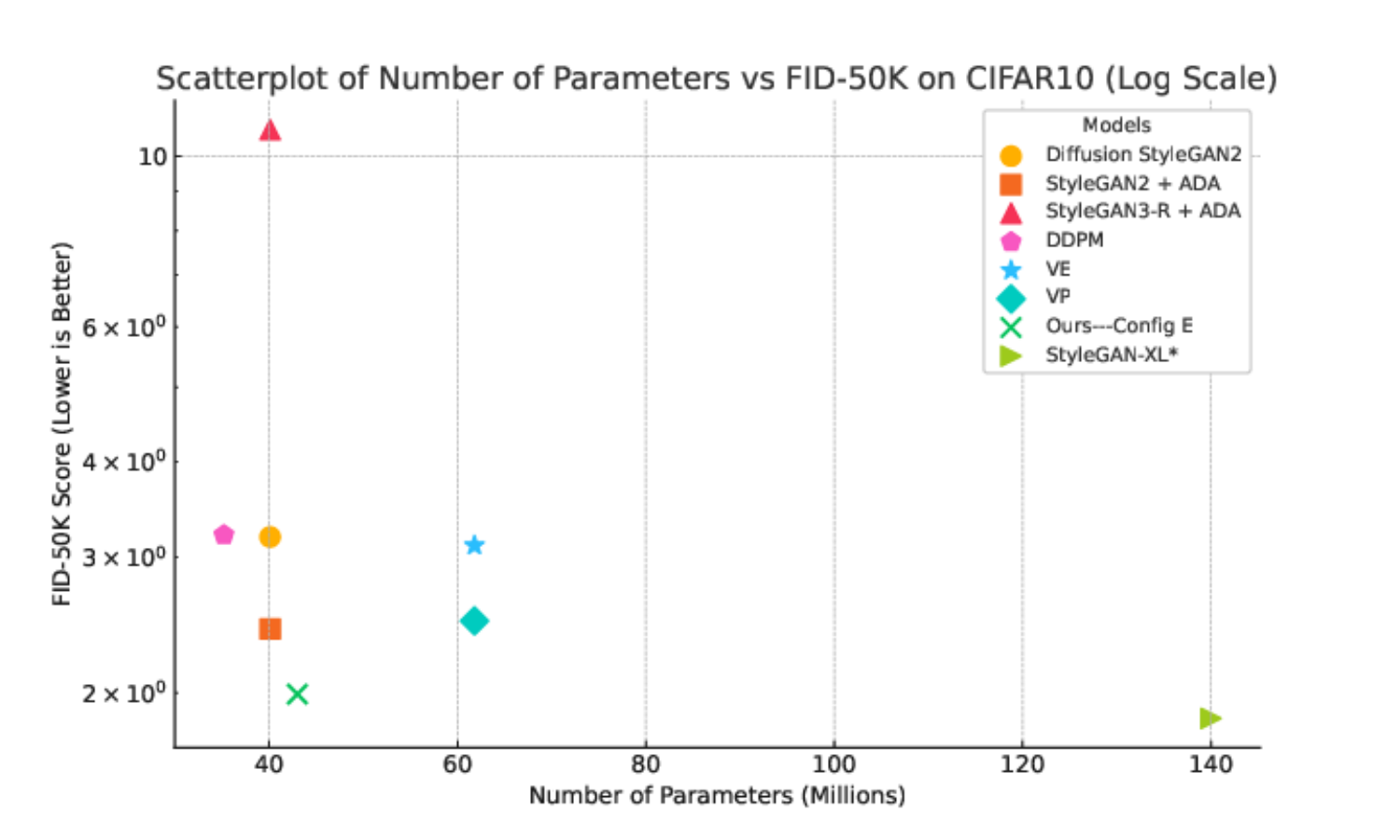}
    \caption{Millions of parameters vs.~FID-50K (log scale) on CIFAR-10. Lower is better.}
    \label{fig:fid-50k-vs-params-cifar-10}
\end{wrapfigure}

We train Config E model until convergence and with optimized hyperparameters and training schedule on CIFAR-10 (conditional generation) (Tab.~\ref{tab:cifar10}, Fig.~\ref{fig:cifar10}). Our method outperforms many other GANs by FID even though the model has relatively small capacity. For instance, StyleGAN-XL~\cite{sgxl} has 18\ M parameters in the generator and 125\ M parameters in the discriminator, while our model has a 40\ M parameters between the generator and discriminator combined (Fig.~\ref{fig:fid-50k-vs-params-cifar-10}). Compared to diffusion models like LDM or ADM, GAN inference is significantly cheaper as it requires only one network function evaluation compared to the tens or hundreds of network function evaluations for diffusion models without distillation. 

Many state-of-the-art GANs are derived from Projected GAN~\cite{sauer2021projected}, including StyleGAN-XL~\cite{sgxl} and the concurrent work of StyleSAN-XL~\cite{takida2024san}. These methods use a pre-trained ImageNet classifier in the discriminator. Prior work has shown that a pre-trained ImageNet discriminator can leak ImageNet features into the model~\cite{kynkaanniemi2022role}, causing the model to perform better when evaluating on FID since it relies on a pre-trained ImageNet classifier for the loss. But, this does not improve results in perceptual studies~\cite{kynkaanniemi2022role}. Our model produces its low FID without any ImageNet pre-training.




\subsection{FID --- ImageNet-32~\cite{chrabaszcz2017downsampled}}
\label{sec:imagenet32-fid-explain}
We train Config E model until convergence and with optimized hyperparameters and training schedule on ImageNet-32 (conditional generation). We compare against recent GAN models and recent diffusion models in Table~\ref{tab:imagenet32}.
We adjust the number of parameters in the generator of our model to match StyleGAN-XL~\cite{sgxl}'s generator (84M parameters). Specifically, we make the model significantly wider to match. Our method achieves comparable FID despite using a 60\% smaller discriminator (Tab.~\ref{tab:imagenet32}) and despite not using a pre-trained ImageNet classifier.

\vspace{-0.1cm}
\subsection{FID --- ImageNet-64~\cite{chrabaszcz2017downsampled}}
We evaluate our model on ImageNet-64 to test its scalability. We stack another resolution stage on our ImageNet-32 model, resulting in a generator of 104\ M parameters. This model is nearly 3$\times$ smaller than diffusion-like models~\cite{adm,edm,cm,icm} that rely on the ADM backbone, which contains about 300\ M parameters. Despite the smaller model size and that our model generates samples in one step, it outperforms larger diffusion models with many NFEs on FID (Tab.~\ref{tab:imagenet64}).

\vspace{-0.1cm}
\subsection{Recall}
We evaluate the recall~\cite{precrecall} of our model on each dataset to quantify sample diversity. In general, our model achieves a recall that is similar to or marginally worse than the diffusion model counterpart, yet superior to existing GAN models. For CIFAR-10, the recall of our model peaked at 0.57; as a point of comparison, StyleGAN-XL~\cite{sgxl} has a worse recall of 0.47 despite its lower FID. For FFHQ, we obtain a recall of 0.53 at 64$\times$64 and 0.49 at 256$\times$256, whereas StyleGAN2~\cite{sg2} achieved a recall of 0.43 on FFHQ-256. Our ImageNet-32 model achieved a recall of 0.63; comparable to ADM~\cite{adm}. Our ImageNet-64 model achieved recall 0.59. While this is slightly worse than $\approx$0.63 that many diffusion models achieve, it is better than BigGAN-deep~\cite{biggan} which achieved a recall of 0.48.

\begin{figure}
    \begin{floatrow}
        \capbtabbox{%
        \centering
        \resizebox{0.9\linewidth}{!}{
        \begin{tblr}{
          column{2} = {r},
          column{3} = {r},
          cell{8}{1} = {c=3}{},
          hline{2,7-8} = {-}{},
        }
    Model                                                       & NFE$\downarrow$  & FID$\downarrow$                        \\ 
    DDPM++~\cite{kim2021soft}                  & 1000 & 8.42                                   \\
    VDM~\cite{kingma2021variational}           & 1000 & 7.41                                   \\
    MSGAN~\cite{karnewar2020msg,ning2023input} & 1    & 12.3                                   \\
    ADM~\cite{adm}                             & 1000 & 3.60                                   \\
    DDPM-IP~\cite{ning2023input}               & 1000 & 2.87                                   \\
    Ours—Config E               & 1    & 1.27   \\
    \textit{With ImageNet feature leakage~\cite{kynkaanniemi2022role}:}    \\
    StyleGAN-XL*~\cite{sgxl}                   & 1    & 1.10                                  
    \end{tblr}
        }
    }{%
        \caption{\label{tab:imagenet32}ImageNet-32.}
    }
    \capbtabbox{
        \centering
        \resizebox{0.9\linewidth}{!}{
        \begin{tblr}{
          column{2} = {r},
          column{3} = {r},
          cell{1}{2} = {c},
          cell{1}{3} = {c},
          cell{12}{1} = {c=3}{},
          hline{2-3,11-12} = {-}{},
        }
        Model         & NFE$\downarrow$ & FID$\downarrow$ \\
        BigGAN-deep~\cite{biggan}\phantom{xx}   & 1               & 4.06            \\
        DDPM~\cite{ddpm}          & 250             & 11.0            \\
        DDIM~\cite{ddim}          & 50              & 13.7            \\
        ADM~\cite{adm}           & $^\S$250             & 2.91            \\
        EDM~\cite{edm}           & 79              & 2.23            \\
        CT~\cite{cm}            & 2               & 11.1            \\
        CD~\cite{cm}            & 3               & 4.32            \\
        iCT-deep~\cite{icm}      & 2               & 2.77            \\
        DMD~\cite{dmd}           & 1               & 2.62            \\
        Ours—Config E & 1               & 2.09            \\
        \emph{With ImageNet feature leakage~\cite{kynkaanniemi2022role}:}          &                 &                 \\
        StyleGAN-XL*~\cite{sgxl}   & 1               & 1.52            
        \end{tblr}
        }
    }
    {
        \caption{\label{tab:imagenet64}ImageNet-64.\hspace{-0.1cm} {\small \S:\hspace{-0.05cm}deterministic sampling.}}
    }
    \end{floatrow}
    \vspace{-0.25cm}
\end{figure}

\input{figures/qualitative/ffhq-256-teaser}

%% file: figures/qualitative/ffhq-256-teaser.tex
\begin{figure}[h!]
    \newlength{\imgsize}
    \setlength{\imgsize}{0.10\linewidth} 
    
    \newcommand{\qualitativeimg}[1]{%
        \includegraphics[width=\imgsize]{figures/qualitative/ffhq-256-000139623/image-#1.jpg}%
    }
    
    \setlength{\tabcolsep}{0pt} 
    \renewcommand{\arraystretch}{0} 
    
    \centering
    \begin{tabular}{cccccccc} 
        \qualitativeimg{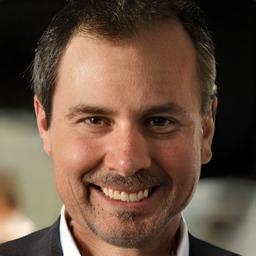} & \qualitativeimg{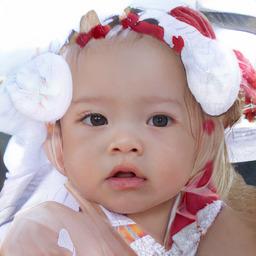} & \qualitativeimg{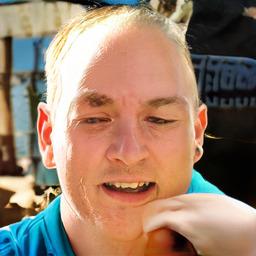} & \qualitativeimg{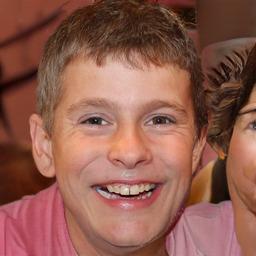} & \qualitativeimg{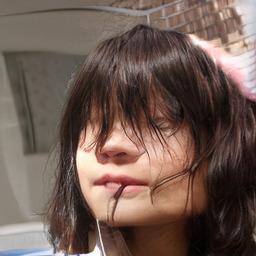} & \qualitativeimg{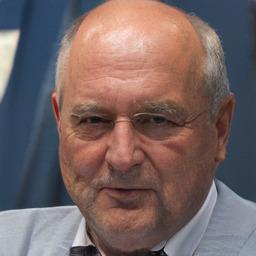} & \qualitativeimg{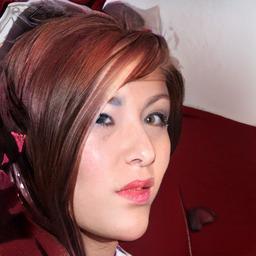} & \qualitativeimg{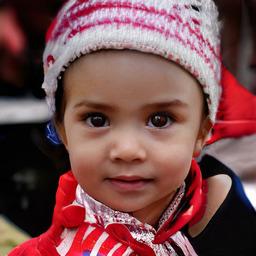} \\
        \qualitativeimg{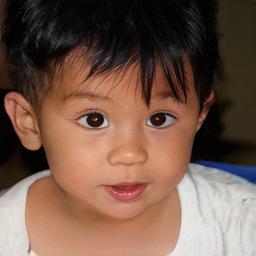} & \qualitativeimg{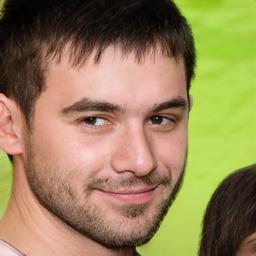} & \qualitativeimg{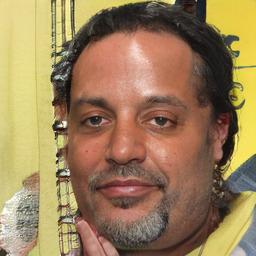} & \qualitativeimg{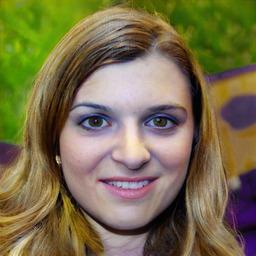} & \qualitativeimg{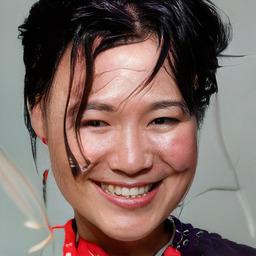} & \qualitativeimg{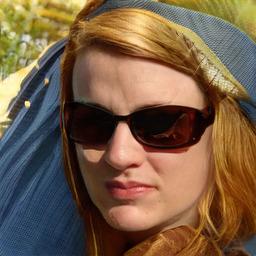} & \qualitativeimg{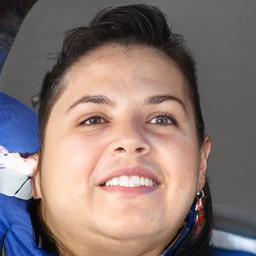} & \qualitativeimg{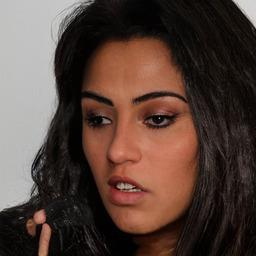} \\
        \qualitativeimg{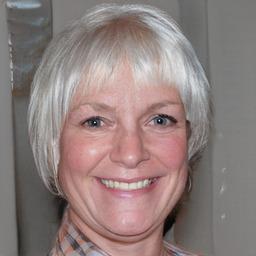} & \qualitativeimg{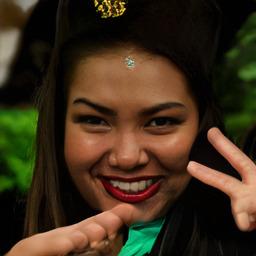} & \qualitativeimg{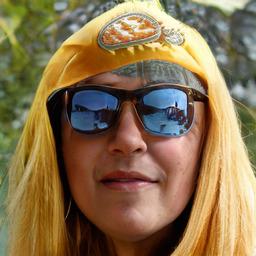} & \qualitativeimg{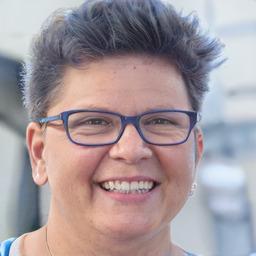} & \qualitativeimg{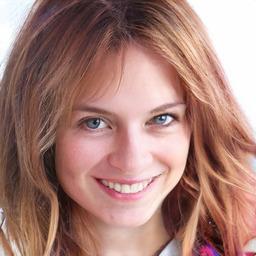} & \qualitativeimg{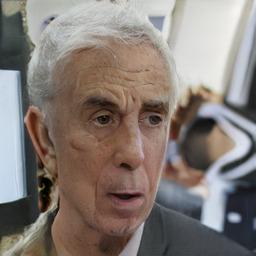} & \qualitativeimg{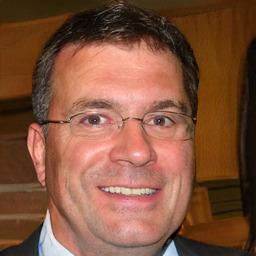} & \qualitativeimg{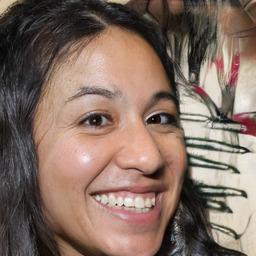} \\
        \qualitativeimg{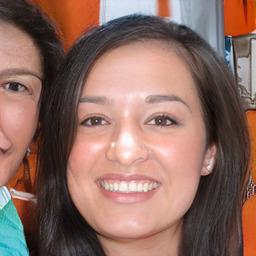} & \qualitativeimg{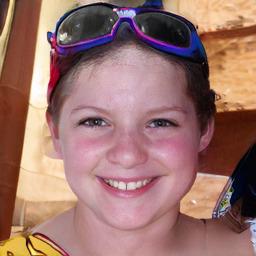} & \qualitativeimg{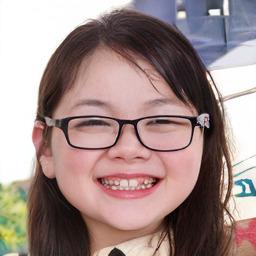} & \qualitativeimg{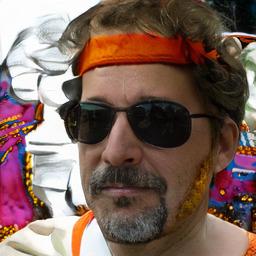} & \qualitativeimg{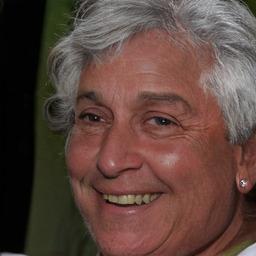} & \qualitativeimg{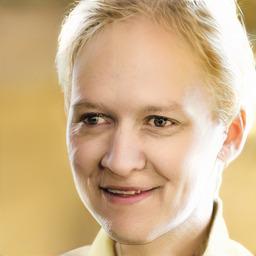} & \qualitativeimg{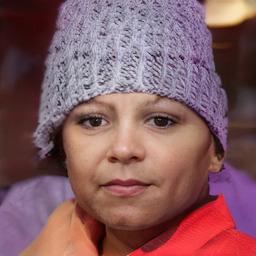} & \qualitativeimg{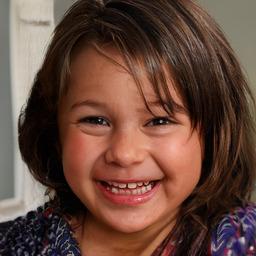} \\
        \qualitativeimg{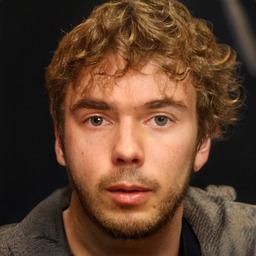} & \qualitativeimg{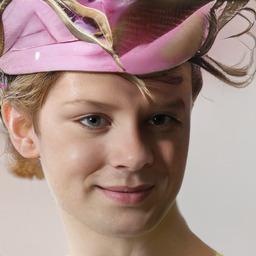} & \qualitativeimg{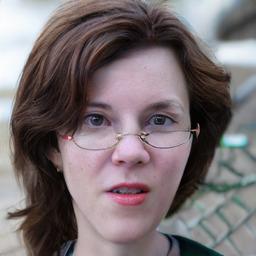} & \qualitativeimg{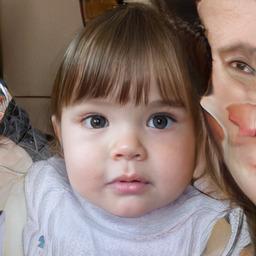} & \qualitativeimg{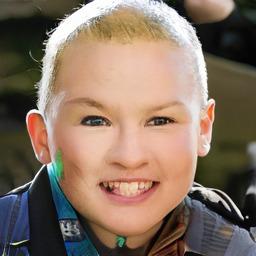} & \qualitativeimg{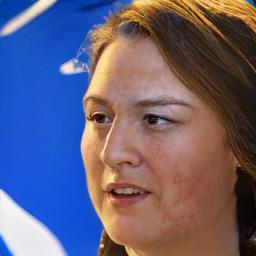} & \qualitativeimg{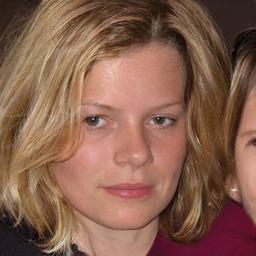} & \qualitativeimg{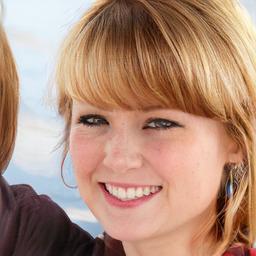} \\
        \qualitativeimg{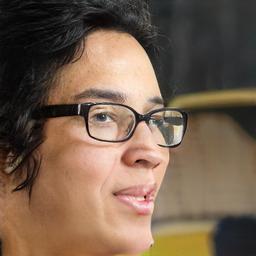} & \qualitativeimg{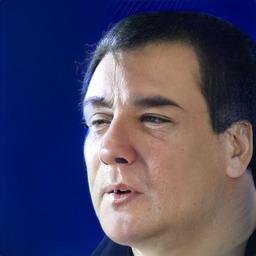} & \qualitativeimg{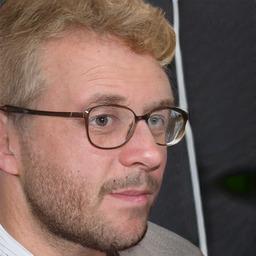} & \qualitativeimg{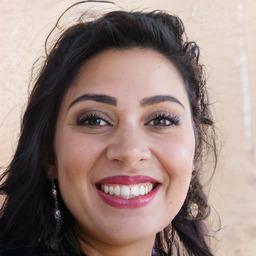} & \qualitativeimg{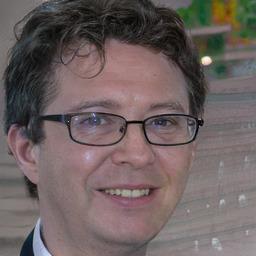} & \qualitativeimg{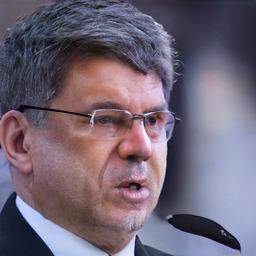} & \qualitativeimg{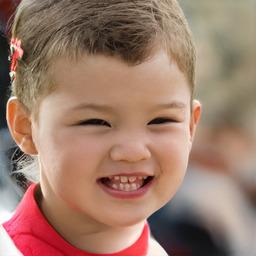} & \qualitativeimg{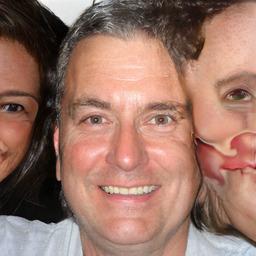} \\
        \qualitativeimg{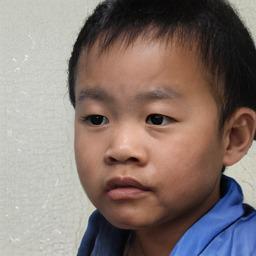} & \qualitativeimg{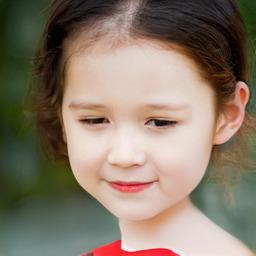} & \qualitativeimg{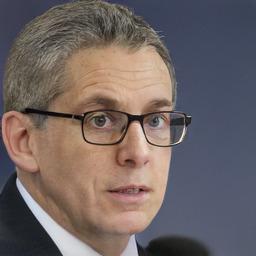} & \qualitativeimg{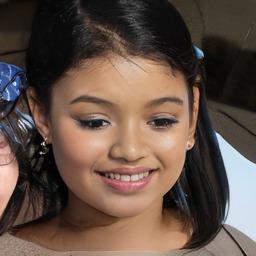} & \qualitativeimg{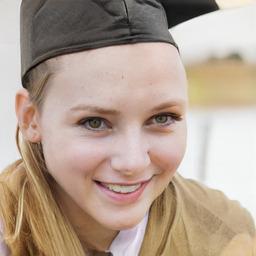} & \qualitativeimg{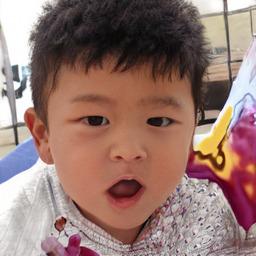} & \qualitativeimg{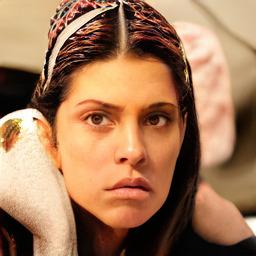} & \qualitativeimg{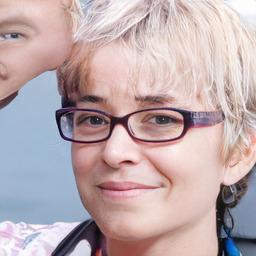} \\
        \qualitativeimg{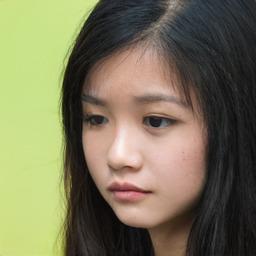} & \qualitativeimg{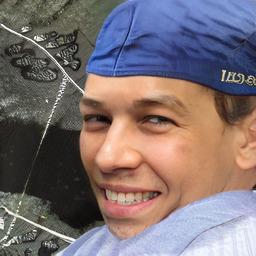} & \qualitativeimg{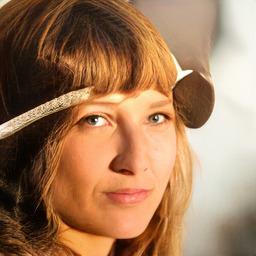} & \qualitativeimg{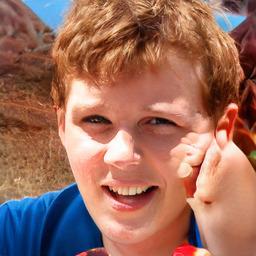} & \qualitativeimg{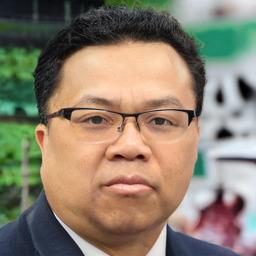} & \qualitativeimg{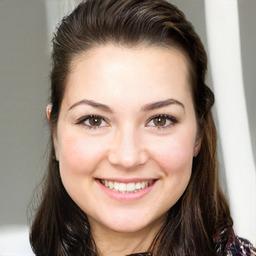} & \qualitativeimg{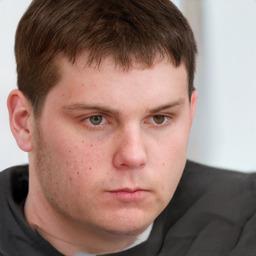} & \qualitativeimg{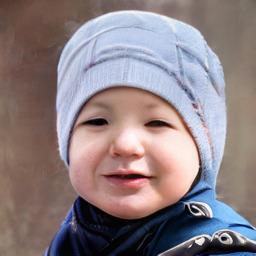} \\
    \end{tabular}
    \caption{Qualitative examples of sample generation from our Config E on FFHQ-256.}
    \label{fig:ffhq-256-teaser}
\end{figure}

%% file: tex/discussion.tex
\vspace{-0.2cm}
\section{Discussion and Limitations}
\vspace{-0.2cm}

We have shown that a simplification of GANs is possible for image generation tasks, built upon a more stable RpGAN$+ R_1 + R_2$ objective with mathematically-demonstrated convergence properties that still provides diverse output. This stability is what lets us re-engineer a modern network architecture without the tricks of previous methods, producing the \modelName model with competitive FID on the common datasets of Stacked-MNIST, FFHQ, CIFAR-10, and ImageNet as an empirical demonstration of the mathematical benefits.

The focus of our work is to elucidate the essential components of a minimum GAN for image generation. 
As such, we prioritize simplicity over functionality---we do not claim to beat the performance of every existing model on every dataset or task; merely to provide a new simple baseline that converges easily.
While this makes our model a possible backbone for future GANs, it also means that it is not suitable to apply our model directly to downstream applications such as image editing or controllable generation, as our model lacks dedicated features for easy image inversion or disentangled image synthesis. 
For instance, we remove style injection functionality from StyleGAN even though this has a clear use.
We also omitted common techniques that have been shown in previous literature to improve FID considerably. 
Examples include some form of adaptive normalization modulated by the latent code~\cite{adm,edm,sg1,zhang2022styleswin,dit,wang2023infodiffusion,sahoo2023diffusion}, and using multiheaded self attention at lower resolution stages~\cite{adm,edm,edm2}. 
We aim to explore these techniques in a future study. 

Further, our work is limited in its evaluation of the scalability of \modelName models. While they show promising results on 64$\times$64 ImageNet, we are yet to verify the scalability on higher resolution ImageNet data or large-scale text to image generation tasks \cite{gokaslan2024commoncanvas}.

Finally, as a method that can improve the quality of generative models, it would be amiss not to mention that generative models---especially of people---can cause direct harm (e.g., through personalized deep fakes) and societal harm through the spread of disinformation (e.g., fake influencers). 

\vspace{-0.2cm}
\section{Conclusion}
\vspace{-0.2cm}

This work introduced \modelName, a new baseline GAN that features increased stability, leverages modern architectures, and does not require ad-hoc tricks that are commonplace in existing GAN models.
Central to our approach is a regularized relativistic loss that provably features local convergence and that improves the stability of GAN training. This stable loss enables us to ablate various tricks that were previously necessary in GANs, and incorporate in their place modern deep architectures. The resulting streamlined baseline achieves competitive performance to SOTA models within its parameter size class. We anticipate that our backbone will help to drive future GAN research.


%% file: tex/appendix.tex
\section{Local convergence}
Following~\cite{r1}, GAN training can be formulated as a dynamical system where the update operator is given by $F_h(\theta,\psi)=(\theta,\psi)+hv(\theta,\psi)$. $h$ is the learning rate and $v$ denotes the gradient vector field:
\begin{equation}
v(\theta,\psi)=\begin{pmatrix}
 -\nabla_\theta\mathcal{L}(\theta,\psi) \\
 \nabla_\psi\mathcal{L}(\theta,\psi)
\end{pmatrix}
\end{equation}
Mescheder et al.~\cite{gannum} showed that local convergence near $(\theta^*,\psi^*)$ can be analyzed by examining the spectrum of the Jacobian $\textbf{J}_{F_h}$ at the equilibrium: if the Jacobian has eigenvalues with absolute value bigger than 1, then training does not converge. On the other hand, if all eigenvalues have absolute value smaller than 1, then training will converge to $(\theta^*,\psi^*)$ at a linear rate. If all eigenvalues have absolute value equal to 1, the convergence behavior is undetermined.

Given some calculations~\cite{r1}, we can show that the eigenvalues of the Jacobian of the update operator $\lambda_{\textbf{J}_{F_h}}$ can be determined by $\lambda_{\textbf{J}_v}$:
\begin{equation}
\lambda_{\textbf{J}_{F_h}}=1+h\lambda_{\textbf{J}_v}\ .
\end{equation}
That is, given small enough $h$~\cite{r1}, the training dynamics can instead be examined using $\lambda_{\textbf{J}_v}$,~\ie, the eigenvalues of the Jacobian of the gradient vector field. If all $\lambda_{\textbf{J}_v}$ have a negative real part, the training will locally converge to $(\theta^*,\psi^*)$ at a linear rate. On the other hand, if some $\lambda_{\textbf{J}_v}$ have a positive real part, the training is not convergent. If all $\lambda_{\textbf{J}_v}$ have a zero real part, the convergence behavior is inconclusive.

\section{DiracRpGAN: A demonstration of non-convergence}
\paragraph{Summary.} To obtain DiracRpGAN, we apply Eq.~\ref{eq:rpgan} to the DiracGAN~\cite{r1} problem setting. After simplification, DiracRpGAN and DiracGAN are different only by a constant. They have the same gradient vector field, therefore all proofs are identical to Mescheder~\etal~\cite{r1}.

\paragraph{Definition B.1.} \emph{The DiracRpGAN consists of a (univariate) generator distribution $p_{\theta} = \delta_{\theta}$ and a linear discriminator $D_{\psi}(x) = \psi \cdot x$. The true data distribution $p_{\mathcal{D}}$ is given by a Dirac distribution concentrated at 0.}

In this setup, the RpGAN training objective is given by:
\begin{equation}
\label{eq:diracrpgan}
    \mathcal{L}(\theta, \psi) = f(\psi \theta)\ .
\end{equation}
We can now show analytically that DiracRpGAN does not converge without regularzation.

\paragraph{Lemma B.2.} \emph{The unique equilibrium point of the training objective in Eq.~\ref{eq:diracrpgan} is given by $\theta = \psi = 0$. Moreover, the Jacobian of the gradient vector field at the equilibrium point has the two eigenvalues $\pm f'(0)i$ which are both on the imaginary axis.}

The gradient vector field $v$ of Eq.~\ref{eq:diracrpgan} is given by:
\begin{equation}
    v(\theta, \psi) =
    \begin{pmatrix}
    -\nabla_{\theta} \mathcal{L}(\theta, \psi) \\
    \nabla_{\psi} \mathcal{L}(\theta, \psi)
    \end{pmatrix} =
    \begin{pmatrix}
    -\psi f'(\psi \theta) \\
    \theta f'(\psi \theta)
    \end{pmatrix}
\end{equation}
and the Jacobian of $v$:
\begin{equation}
    \textbf{J}_v =
    \begin{pmatrix}
    -\psi^2 f''(\psi\theta) & -f'(\psi\theta) - \psi\theta f''(\psi\theta) \\
    f'(\psi\theta) + \psi\theta f''(\psi\theta) & \theta^2 f''(\psi\theta)
    \end{pmatrix}
\end{equation}
Evaluating $\textbf{J}_v$ at the equilibrium point $\theta = \psi = 0$ gives us:
\begin{equation}
    \textbf{J}_v \biggr\rvert_{(0,0)} =
    \begin{pmatrix}
    0 & -f'(0) \\
    f'(0) & 0
    \end{pmatrix}
\end{equation}
Therefore, the eigenvalues of $\textbf{J}_v$ are $\lambda_{1/2} = \pm f'(0) i$, both of which have a real part of 0. Thus, the convergence of DiracRpGAN is inconclusive and further analysis is required.

\paragraph{Lemma B.3.} \emph{The integral curves of the gradient vector field $v(\theta, \psi)$ do not converge to the equilibrium point. More specifically, every integral curve $(\theta(t), \psi(t))$ of the gradient vector field $v(\theta, \psi)$ satisfies $\theta(t)^2 + \psi(t)^2 = const$ for all $t \in [0, \infty)$.}

Let $R(\theta, \psi) = \frac{1}{2} (\theta^2 + \psi^2)$, then:
\begin{align}
    \frac{\mathrm{d}}{\mathrm{d}t} R(\theta(t), \psi(t)) \nonumber &= -\theta(t) \psi(t) f'(\theta(t) \psi(t)) + \psi(t) \theta(t) f'(\theta(t) \psi(t)) \nonumber \\
    &= 0\ .
\end{align}
We see that the distance between $(\theta, \psi)$ and the equilibrium point $(0,0)$ stays constant. Therefore, training runs in circles and never converges.

Next, we investigate the convergence behavior of DiracRpGAN with regularization. For DiracRpGAN, both $R_1$ and $R_2$ can be reduced to the following form:
\begin{equation}
    R(\psi) = \frac{\gamma}{2} \psi^2
\end{equation}

\paragraph{Lemma B.4.} \emph{The eigenvalues of the Jacobian of the gradient vector field for the gradient-regularized DiracRpGAN at the equilibrium point are given by
\begin{equation}
\label{eq:ev}
    \lambda_{1/2} = -\frac{\gamma}{2} \pm \sqrt{\frac{\gamma^2}{4}-f'(0)}
\end{equation}
In particular, for $\gamma > 0$ all eigenvalues have a negative real part. Hence, gradient descent is locally convergent for small enough learning rates.}

With regularization, the gradient vector field becomes
\begin{equation}
    \Tilde{v} (\theta, \psi) =
    \begin{pmatrix}
        -\psi f'(\psi\theta) \\
        \theta f'(\psi\theta) - \gamma\psi
    \end{pmatrix}
\end{equation}
the Jacobian of $\Tilde{v}$ is then given by
\begin{equation}
    \textbf{J}_{\Tilde{v}} =
    \begin{pmatrix}
    -\psi^2 f''(\psi\theta) & -f'(\psi\theta) - \psi\theta f''(\psi\theta) \\
    f'(\psi\theta) + \psi\theta f''(\psi\theta) & \theta^2 f''(\psi\theta) - \gamma
    \end{pmatrix}
\end{equation}
evaluating the Jacobian at $\theta = \psi = 0$ yields
\begin{equation}
    \textbf{J}_{\Tilde{v}} \biggr\rvert_{(0, 0)} =
    \begin{pmatrix}
        0 & -f'(0) \\
        f'(0) & -\gamma
    \end{pmatrix}
\end{equation}
given some calculations, we arrive at Eq.\ref{eq:ev}.

\section{General Convergence Results}
\paragraph{Summary.} The proofs are largely the same as Mescheder~\etal~\cite{r1}. We use the same proving techniques, and only slightly modify the assumptions and proof details to adapt Mescheder~\etal's effort to RpGAN. Like in~\cite{r1}, our proofs do not rely on unrealistic assumptions such as $\supp p_\mathcal{D}=\supp p_\theta$.

\subsection{Assumptions}
We closely follow~\cite{r1} but modify the assumptions wherever necessary to tailor the proofs for RpGAN. Like in~\cite{r1}, we also consider the realizable case where there exists $\theta$ such that $G_\theta$ produces the true data distribution.

\paragraph{Assumption~\upperRomannumeral{1}.}
\label{a:1}
\emph{We have $p_{\theta^*}=p_\mathcal{D}$, and $D_{\psi^*}=C$ in some local neighborhood of $\supp p_\mathcal{D}$, where $C$ is some arbitrary constant.}

\noindent Since RpGAN is defined on critic difference rather than raw logits, we no longer require $D_{\psi^*}$ to produce 0 on $\supp p_\mathcal{D}$, instead any constant $C$ would suffice.

\paragraph{Assumption~\upperRomannumeral{2}.}
\label{a:2}
\emph{We have $f^\prime(0)\neq0$ and $f^{\prime\prime}(0)<0$.}

\noindent This assumption is the same as in~\cite{r1}. The choice $f(t) = -\log(1+e^{-t})$ adopted in the main text satisfies this assumption.

As discussed in~\cite{r1}, there generally is not a single equilibrium point $(\theta^*,\psi^*)$, but a submanifold of equivalent equilibria corresponding to different parameterizations of the same function. It is therefore necessary to represent the equilibrium as \emph{reparameterization manifolds} $\mathcal{M}_G$ and $\mathcal{M}_D$. We modify the reparameterization $h$ as follows:
\begin{equation}
\label{eq:h}
h(\psi)=\mathbb{E}_{\substack{x\sim p_\mathcal{D}\\y\sim p_\mathcal{D}}}\left[\left | D_\psi(x)-D_\psi(y)\right |^2   +  \left \| \nabla_x D_\psi(x) \right \|^2\right]
\end{equation}
to account for the fact that $D_{\psi^*}$ is now allowed to have any constant value on $\supp p_\mathcal{D}$. The \emph{reparameterization manifolds} are then given by:
\begin{align}
&\mathcal{M}_G=\{\theta\,\rvert\,p_\theta=p_\mathcal{D}\} \\
&\mathcal{M}_D=\{\psi\,\rvert\,h(\psi)=0 \}
\end{align}
We assume the same regularity properties as in~\cite{r1} for $\mathcal{M}_G$ and $\mathcal{M}_D$ near the equilibrium. To state these assumptions, we need:
\begin{equation}
g(\theta)=\mathbb{E}_{x\sim p_\theta}\left[\nabla_\psi D_\psi\rvert_{\psi=\psi^*}\right]
\end{equation}
which leads to:
\paragraph{Assumption~\upperRomannumeral{3}.}
\label{a:3}
\emph{There are $\epsilon$-balls $B_\epsilon(\theta^*)$ and $B_\epsilon(\psi^*)$ around $\theta^*$ and $\psi^*$ so that $\mathcal{M}_G\,\cap\,B_\epsilon(\theta^*)$ and $\mathcal{M}_D\,\cap\,B_\epsilon(\psi^*)$ define $\mathcal{C}^1$-manifolds. Moreover, the following holds}:
\begin{enumerate}[label=(\roman*)]
\item \emph{if $v\in\mathbb{R}^n$ is not in $\mathcal{T}_{\psi^*}\mathcal{M}_D$, then $\partial_v^2h(\psi^*)\neq0$}.
\item \emph{if $w\in\mathbb{R}^m$ is not in $\mathcal{T}_{\theta^*}\mathcal{M}_G$, then $\partial_wg(\theta^*)\neq0$}.
\end{enumerate}

These two conditions have exactly the same meanings as in~\cite{r1}: the first condition indicates the geometry of $\mathcal{M}_D$ can be locally described by the second derivative of $h$. The second condition implies that $D$ is strong enough that it can detect any deviation from the equilibrium generator distribution. This is the only assumption we have about the expressiveness of $D$.

\subsection{Convergence}
We can now show the general convergence result for gradient penalized RpGAN, consider the gradient vector field with either $R_1$ or $R_2$ regularization:
\begin{equation}
\label{eq:vreg}
\tilde{v}_i(\theta,\psi)=\begin{pmatrix}
-\nabla_\theta\mathcal{L}(\theta,\psi)\\ 
\nabla_\psi\mathcal{L}(\theta,\psi)-\nabla_\psi R_i(\theta,\psi)
\end{pmatrix}
\end{equation}
note that the convergence result can also be trivially extended to the case where both $R_1$ and $R_2$ are applied. We omit the proof for this case as it is redundant once the convergence with either regularization is proven.

\paragraph{Theorem.} \emph{Assume Assumption~\upperRomannumeral{1},~\upperRomannumeral{2} and~~\upperRomannumeral{3} hold for $(\theta^*,\psi^*)$. For small enough learning rates, gradient descent for $\tilde{v}_1$ and $\tilde{v}_2$ are both convergent to $\mathcal{M}_G\times\mathcal{M}_D$ in a neighborhood of $(\theta^*,\psi^*)$. Moreover, the rate of convergence is at least linear}.

We extend the convergence proof by Mescheder~\etal~\cite{r1} to our setting. We first prove lemmas necessary to our main proof.

\paragraph{Lemma C.2.1.} \emph{Assume $J\in\mathbb{R}^{(n+m)\times(n+m)}$ is of the following form}:
\begin{equation}
J=\begin{pmatrix}
0 & -B^\top\\ 
B & -Q
\end{pmatrix}
\end{equation}
\emph{where $Q\in\mathbb{R}^{m\times m}$ is a symmetric positive definite matrix and $B\in\mathbb{R}^{m\times n}$ has full column rank. Then all eigenvalues $\lambda$ of $J$ satisfy $\Re(\lambda)< 0$}.

\emph{Proof.} See Mescheder~\etal~\cite{r1}, Theorem A.7.

\paragraph{Lemma C.2.2.} \emph{The gradient of $\mathcal{L}(\theta,\psi)$~\wrt $\theta$ and $\psi$ are given by}:
\begin{align}
&\nabla_\theta\mathcal{L}(\theta,\psi)=\mathbb{E}_{\substack{z\sim p_z\\x\sim p_\mathcal{D}}}[f^\prime(D_\psi(G_\theta(z))-D_\psi(x)) \left[\nabla_\theta G_\theta(z)\right]^\top\nabla_xD_\psi(G_\theta(z))] \\
\label{grad_psi}
&\nabla_\psi\mathcal{L}(\theta,\psi)=\mathbb{E}_{\substack{z\sim p_z\\x\sim p_\mathcal{D}}}[f^\prime(D_\psi(G_\theta(z))-D_\psi(x)) (\nabla_\psi D_\psi(G_\theta(z))-\nabla_\psi D_\psi(x))]
\end{align}
\emph{Proof.} This is just the chain rule.

\paragraph{Lemma C.2.3.} \emph{Assume that $(\theta^*,\psi^*)$ satisfies Assumption~\upperRomannumeral{1}. The Jacobian of the gradient vector field $v(\theta,\psi)$ at $(\theta^*,\psi^*)$ is then}
\begin{equation}
\textbf{J}_v\biggr\rvert_{(\theta^*,\psi^*)}=\begin{pmatrix}
0 & -K^\top_{DG}\\ 
K_{DG} & K_{DD}
\end{pmatrix}
\end{equation}
\emph{the terms $K_{DD}$ and $K_{DG}$ are given by}
\begin{align}
\label{kdd}
&K_{DD}=f^{\prime\prime}(0)\mathbb{E}_{\substack{x\sim p_\mathcal{D}\\y\sim p_\mathcal{D}}}[(\nabla_\psi D_{\psi^*}(x)-\nabla_\psi D_{\psi^*}(y))(\nabla_\psi D_{\psi^*}(x)-\nabla_\psi D_{\psi^*}(y))^\top] \\
\label{kdg}
&K_{DG}=f^\prime(0)\nabla_\theta\mathbb{E}_{x\sim p_\theta}[\nabla_\psi D_{\psi^*}(x)]\,\rvert_{\theta=\theta^*}
\end{align}

\emph{Proof.} Note that
\begin{equation}
\textbf{J}_v\biggr\rvert_{(\theta^*,\psi^*)}=\begin{pmatrix}
-\nabla^2_\theta\mathcal{L}(\theta^*,\psi^*) & -\nabla^2_{\theta,\psi}\mathcal{L}(\theta^*,\psi^*) \\ 
\nabla^2_{\theta,\psi}\mathcal{L}(\theta^*,\psi^*) & \nabla^2_\psi\mathcal{L}(\theta^*,\psi^*)
\end{pmatrix}
\end{equation}
By Assumption~\upperRomannumeral{1}, $D_{\psi^*}=C$ in some neighborhood of $\supp p_\mathcal{D}$. Therefore we also have $\nabla_x D_{\psi^*}=0$ and $\nabla^2_x D_{\psi^*}=0$ for $x\in\supp p_\mathcal{D}$. Using these two conditions, we see that $\nabla^2_\theta\mathcal{L}(\theta^*,\psi^*)=0$.

To see Eq.\ref{kdd} and Eq.\ref{kdg}, simply take the derivatives of Eq.\ref{grad_psi} and evaluate at $(\theta^*,\psi^*)$.

\paragraph{Lemma C.2.4.} \emph{The gradient $\nabla_\psi R_i(\theta,\psi)$ of the regularization terms $R_i$, $i\in\{1,2\}$,~\wrt $\psi$ are}
\begin{align}
&\nabla_\psi R_1(\theta,\psi)=\gamma\mathbb{E}_{x\sim p_\mathcal{D}}[\nabla_{\psi,x}D_\psi\nabla_xD_\psi] \\
&\nabla_\psi R_2(\theta,\psi)=\gamma\mathbb{E}_{x\sim p_\theta}[\nabla_{\psi,x}D_\psi\nabla_xD_\psi]
\end{align}

\emph{Proof.} See Mescheder~\etal~\cite{r1}, Lemma D.3.

\paragraph{Lemma C.2.5.} \emph{The second derivatives $\nabla^2_\psi R_i(\theta^*,\psi^*)$ of the regularization terms $R_i$, $i\in\{1,2\}$,~\wrt $\psi$ at $(\theta^*,\psi^*)$ are both given by}
\begin{equation}
L_{DD}=\gamma\mathbb{E}_{x\sim p_\mathcal{D}}[AA^\top]
\end{equation}
\emph{where $A=\nabla_{\psi,x}D_{\psi^*}$. Moreover, both regularization terms satisfy $\nabla_{\theta,\psi}R_i(\theta^*,\psi^*)=0$.}

\emph{Proof.} See Mescheder~\etal~\cite{r1}, Lemma D.4.

Given Lemma C.2.3, Lemma C.2.5 and Eq.\ref{eq:vreg}, we can now show that the Jacobian of the regularized gradient field at the equilibrium point is given by
\begin{equation}
\textbf{J}_{\tilde{v}}\biggr\vert_{(\theta^*,\psi^*)}=\begin{pmatrix}
0 & -K_{DG}^\top\\ 
K_{DG} & M_{DD}
\end{pmatrix}
\end{equation}
where $M_{DD}=K_{DD}-L_{DD}$. To prove our main theorem, we need to examine $\textbf{J}_{\tilde{v}}$ when restricting it to the space orthogonal to $\mathcal{T}_{(\theta^*,\psi^*)}\mathcal{M}_G\times\mathcal{M}_D$.

\paragraph{Lemma C.2.6.} \emph{Assume Assumptions~\upperRomannumeral{2} and~\upperRomannumeral{3} hold. If $v\neq0$ is not in $\mathcal{T}_{\psi^*}\mathcal{M}_D$, then $v^\top M_{DD}v<0$.}

\emph{Proof.} By Lemma C.2.3 and Lemma C.2.5, we have
\begin{align}
&v^\top K_{DD}v=f^{\prime\prime}(0)\mathbb{E}_{\substack{x\sim p_\mathcal{D}\\y\sim p_\mathcal{D}}}\left[((\nabla_\psi D_{\psi^*}(x)-\nabla_\psi D_{\psi^*}(y))^\top v)^2\right] \\
&v^\top L_{DD}v=\gamma\mathbb{E}_{x\sim p_\mathcal{D}}\left[\left\|Av\right \|^2\right]
\end{align}
By Assumption~\upperRomannumeral{2}, we have $f^{\prime\prime}(0)<0$. Therefore $v^\top M_{DD}v\leq0$. Suppose $v^\top M_{DD}v=0$, this implies
\begin{equation}
(\nabla_\psi D_{\psi^*}(x)-\nabla_\psi D_{\psi^*}(y))^\top v=0\,\,\,\,\text{and}\,\,\,\,Av=0
\end{equation}
for all $(x,y)\in\supp p_\mathcal{D}\times\supp p_\mathcal{D}$. Recall the definition of $h(\psi)$ from Eq.\ref{eq:h}. Using the fact that $D_{\psi^*}=C$ and $\nabla_x D_{\psi^*}=0$ for $x\in\supp p_\mathcal{D}$, we see that the Hessian of $h(\psi)$ at $\psi^*$ is
\begin{equation}
\nabla^2_\psi h(\psi^*)=2\mathbb{E}_{\substack{x\sim p_\mathcal{D}\\y\sim p_\mathcal{D}}}[(\nabla_\psi D_{\psi^*}(x)-\nabla_\psi D_{\psi^*}(y))(\nabla_\psi D_{\psi^*}(x)-\nabla_\psi D_{\psi^*}(y))^\top+AA^\top]
\end{equation}
The second directional derivative $\partial^2_v h(\psi)$ is therefore
\begin{equation}
\partial^2_v h(\psi)=2\mathbb{E}_{\substack{x\sim p_\mathcal{D}\\y\sim p_\mathcal{D}}}\left[\left| (\nabla_\psi D_{\psi^*}(x)-\nabla_\psi D_{\psi^*}(y))^\top v\right|^2 + \left\|Av\right\|^2\right]=0
\end{equation}
By Assumption~\upperRomannumeral{3}, this can only hold if $v\in\mathcal{T}_{\psi^*}\mathcal{M}_D$.

\paragraph{Lemma C.2.7.} \emph{Assume Assumption~\upperRomannumeral{3} holds. If $w\neq0$ is not in $\mathcal{T}_{\theta^*}\mathcal{M}_G$, then $K_{DG}w\neq0$.}

\emph{Proof.} See Mescheder~\etal~\cite{r1}, Lemma D.6.

\emph{Proof for the main theorem.} Given previous lemmas, by choosing local coordinates $\theta(\alpha,\gamma_G)$ and $\psi(\beta,\gamma_D)$ for $\mathcal{M}_G$ and $\mathcal{M}_D$ such that $\theta^*=0$, $\psi^*=0$ as well as
\begin{align}
\mathcal{M}_G=\mathcal{T}_{\theta^*}\mathcal{M}_G=\{0\}^k\times\mathbb{R}^{n-k} \\
\mathcal{M}_D=\mathcal{T}_{\psi^*}\mathcal{M}_D=\{0\}^l\times\mathbb{R}^{m-l}
\end{align}
our proof is \emph{exactly} the same as Mescheder~\etal~\cite{r1}, Theorem 4.1.

\newpage
\section{Hyperparameters, training configurations, and compute}
We implement our models on top of the official StyleGAN3 code base. While the loss function and the models are implemented from scratch, we reuse support code from the existing implementation whenever possible. This includes exponential moving average (EMA) of generator weights~\cite{pggan}, non-leaky data augmentation~\cite{sg2ada}, and metric evaluation~\cite{sg3}.

\vspace{-0.1cm}
\paragraph{Training schedule.}
To speed up the convergence early in training, we specify a cosine schedule for the following hyperparameters before they reach their target values: 
\begin{itemize}[parsep=2pt,topsep=0pt,itemsep=0pt]
    \item Learning rate
    \item $\gamma$ for $R_1$ and $R_2$ regularization
    \item Adam $\beta_2$
    \item EMA half-life
    \item Augmentation probability
\end{itemize}
We call this early training stage the burn-in phase. Burn-in length and schedule for each hyperparameter are listed in Table~\ref{tab:hyperparam} for each experiment. A schedule for the EMA half-life can already be found in Karras~\etal~\cite{sg2ada}, albeit they use a linear schedule. A lower initial Adam $\beta_2$ is crucial to the initial large learning rate as it allows the optimizer to adapt to the gradient magnitude change much quicker. We use a large initial $\gamma$ to account for that early in training: $p_\theta$ and $p_\mathcal{D}$ are far apart and a large $\gamma$ smooths both distributions more aggressively which makes learning easier. Augmentation is not necessary until $D$ starts to overfit later on; thus, we set the initial augmentation probability to 0.

\vspace{-0.1cm}
\paragraph{Dataset augmentation.}
We apply horizontal flips and non-leaky augmentation~\cite{sg2ada} to all datasets where augmentation is enabled. Following~\cite{sg2ada}, we include pixel blitting, geometric transformations, and color transforms in the augmentation pipeline. We additionally include cutout augmentation which works particularly well with our model, although it does not seem to have much effect on StyleGAN2. We also find it beneficial to apply color transforms less often and thus set their probability multiplier to 0.5 while retaining the multiplier 1 for other types of augmentations. As previously mentioned, we apply a fixed cosine schedule to the augmentation probability rather than adjusting it adaptively as in~\cite{sg2ada}. We did not observe any performance degradation with this simplification.

\vspace{-0.1cm}
\paragraph{Network capacity.}
We keep the capacity distribution for each resolution the same as in~\cite{sg2ada,sg3}. We place two residual blocks per resolution which makes our model roughly 3$\times$ as deep, 1.5--3$\times$ as wide as StyleGAN2 while maintaining the same model size on CIFAR-10 and FFHQ. For the ImageNet model, we double the number of channels which results in roughly 4$\times$ as many parameters as the default StyleGAN2 configuration.

\vspace{-0.1cm}
\paragraph{Mixed precision training.}
We apply mixed precision training as in~\cite{sg2ada,sg3} where all parameters are stored in FP32, but cast to lower precision along with the activation maps for the 4 highest resolutions. We notice that using FP16 as the low precision format cripples the training of our model. However, we see no problem when using BFloat16 instead.

\vspace{-0.1cm}
\paragraph{Class conditioning.}
For class conditional models, we follow the same conditioning scheme as in~\cite{sg2ada}. For $G$, the conditional latent code $z^\prime$ is the concatenation of $z$ and the embedding of the class label $c$, specifically $z^\prime=\text{concat}(z,\text{embed}(c))$. For $D$, we use a projection discriminator~\cite{cgans} which evaluates the dot product of the class embedding and the feature vector $D^\prime(x)$ produced by the last layer of $D$, concretely $D(x)=\text{embed}(c)\cdot D^\prime(x)^\top$. We do not employ any normalization-based conditioning such as AdaIN~\cite{sg1}, AdaGN~\cite{adm,edm}, AdaBN~\cite{biggan} or AdaLN~\cite{dit} for simplicity, even though they improve FID considerably.

\vspace{-0.1cm}
\paragraph{Stacked MNIST.}
We base this model off of the CIFAR-10 model but without class conditioning. We disable all data augmentation and shorten the burn-in phase considerably. We use a constant learning rate and did not observe any benefit of using a lower learning rate later in the training.

\vspace{-0.1cm}
\paragraph{Compute resources.}
We train the Stacked MNIST and CIFAR-10 models on an $8\times$ NVIDIA L40 node. Training took 7 hours for Stacked MNIST and 4 days for CIFAR-10. The FFHQ model was trained on an $8\times$ NVIDIA A6000 f0r roughly 3 weeks. The ImageNet model was trained on NVIDIA A100/H100 clusters and training took one day on 32 H100s (about 5000 H100 hours).

\input{tex/table_hyperparameters}

\section{Negative Results and Future Work}
Following the convention of Brock~\etal~\cite{biggan}, we report alternative design choices that did not make to our final model. Either because they failed to produce any quantitative improvement or because they would considerably complicate our minimalist design which might be better suited for future study.
\begin{itemize}
    \item We tried to apply GELU~\cite{gelu}, Swish~\cite{swish}, and SMU~\cite{smu} to $G$ and $D$ and found that doing so deteriorates FID considerably. We did not try on $G$ xor $D$. We posit two independent factors:
    \begin{itemize}
        \item ConvNeXt in general does not benefit much from GELU (and possibly similar activations). Table 10 and Table 11 in~\cite{convnext}: replacing ReLU with GELU gives little performance gain to ConvNeXt-T and virtually no performance gain to ConvNeXt-B.
        \item In the context of GANs, GELU and Swish have the same problem as ReLU: that they have little gradient in the negative interval. Since G is updated from the gradient of D, having these activation functions in D could sparsify the gradient of D and as a result G will not receive as much useful information from D compared to using leaky ReLU.
    \end{itemize}
    This does not explain the strange case of SMU~\cite{smu}: SMU is a smooth approximation of leaky ReLU and does not have the sparse gradient problem. It is unclear why it also underperformed and future work awaits.
    \item We tried adding group normalization~\cite{groupnorm} to $G$ and $D$ and it did not improve FID or training stability. We do not claim that all forms of normalizations are harmful. Our claim in principle c) only extends to normalization layers that explicitly standardizes the mean and standard deviation of the activation maps. This has been verified by prior studies~\cite{sg2,edm2,litevae}. The harm of normalization layers extends to the adjacent field of image restoration~\cite{edsr,esrgan}. To the best of our knowledge, EDM2~\cite{edm2} is currently the strongest diffusion UNet and it does not use normalization layers. However, it does apply normalization to the trainable weights and this improves performance considerably. We expect that applying the normalization techniques in EDM2 would improve our model's performance.
    \item We tried removing the activation function after the 3$\times$3 grouped convolution in each residual block as modern architectures~\cite{convnext,metaformer} typically do not apply non-linearity after depthwise convolution. This worsened FID performance.
    \item We tried Pixel-Shuffle/Unshuffle~\cite{pixshuffle} for changing the resolution of the activation maps, and found that without low-pass filtering, this led to high frequency artifacts similar to checkerboard artifacts even though Pixel-Shuffle does not have the uneven overlap problem that transposed convolution does. Note that bilinear resampling is equivalent to applying channel duplication/averaging with Pixel-Shuffle/Unshuffle in conjunction with a [1, 2, 1] low-pass kernel. It might be interesting in future studies to explore inplace resampling filters that apply a low-pass filtered Pixel-Shuffle/Unshuffle operation on top of a learned function that changes the number of channels.
    \item We tried scaling up our model size. We found that allocating more model capacity to lower resolution stages generally did not improve FID, but contributed to more rapid overfitting. Increasing model capacity at higher resolution stages always improves FID in our experiments, however scaling up higher resolution stages is very computationally expensive. Capacity distribution for each resolution stage of the model might be an important topic to explore in future studies.
    \item For model simplicity, we did not conduct any experiment with a transformer architecture or attention mechanism in general. We are interested to see whether adding attention blocks to a convolutional network (similar to BigGAN~\cite{biggan} and diffusion UNet~\cite{ddpm,edm,edm2}) or using a pure transformer architecture (similar to DiT~\cite{dit}) will result in stronger performance. Given the impressive results of EDM2~\cite{edm2} (which uses UNet), it seems the argument has not yet settled for generative modeling.
    \item We experimented with Adam $\beta_2=0.999$ following common practice in supervised learning and diffusion models, and found that doing so led to stability issues on our ImageNet models. We expect that introducing proper normalization to our model will resolve this problem.
    \item We tried mixed precision training with IEEE FP16 as this is the low precision format used in StyleGAN2-ADA~\cite{sg2ada}, StyleGAN3~\cite{sg3}, and EDM2~\cite{edm2}. This crippled the training of our model and switching to BFloat16 fixed the problem. We expect that introducing proper normalization to our model will allow us to use IEEE FP16 which offers more precision than BFloat16.
    \item We tried lazy regularization~\cite{sg2} in our early experiments where $R_1$ and $R_2$ were applied once every 8 minibatches. This led to slightly worse FID performance on real world datasets like FFHQ and CIFAR-10. However, it resulted in complete convergence failure on Stacked MNIST and several two dimensional toy datasets (line, circle, 25 Gaussians,~\etc), indicating potential concerns regarding the mathematical legitimacy of this trick.
\end{itemize}
\newpage
\section{Qualitative Results}

\input{figures/qualitative/stacked-mnist-64x64}
\FloatBarrier
\input{figures/qualitative/ffhq-256-8x8}
\FloatBarrier
{
\begin{figure}[h!]
    \setlength{\imgsize}{0.2\linewidth} 
    
    \setlength{\tabcolsep}{0pt} 
    \renewcommand{\arraystretch}{0} 

    \centering
    
    \includegraphics[width=\linewidth]{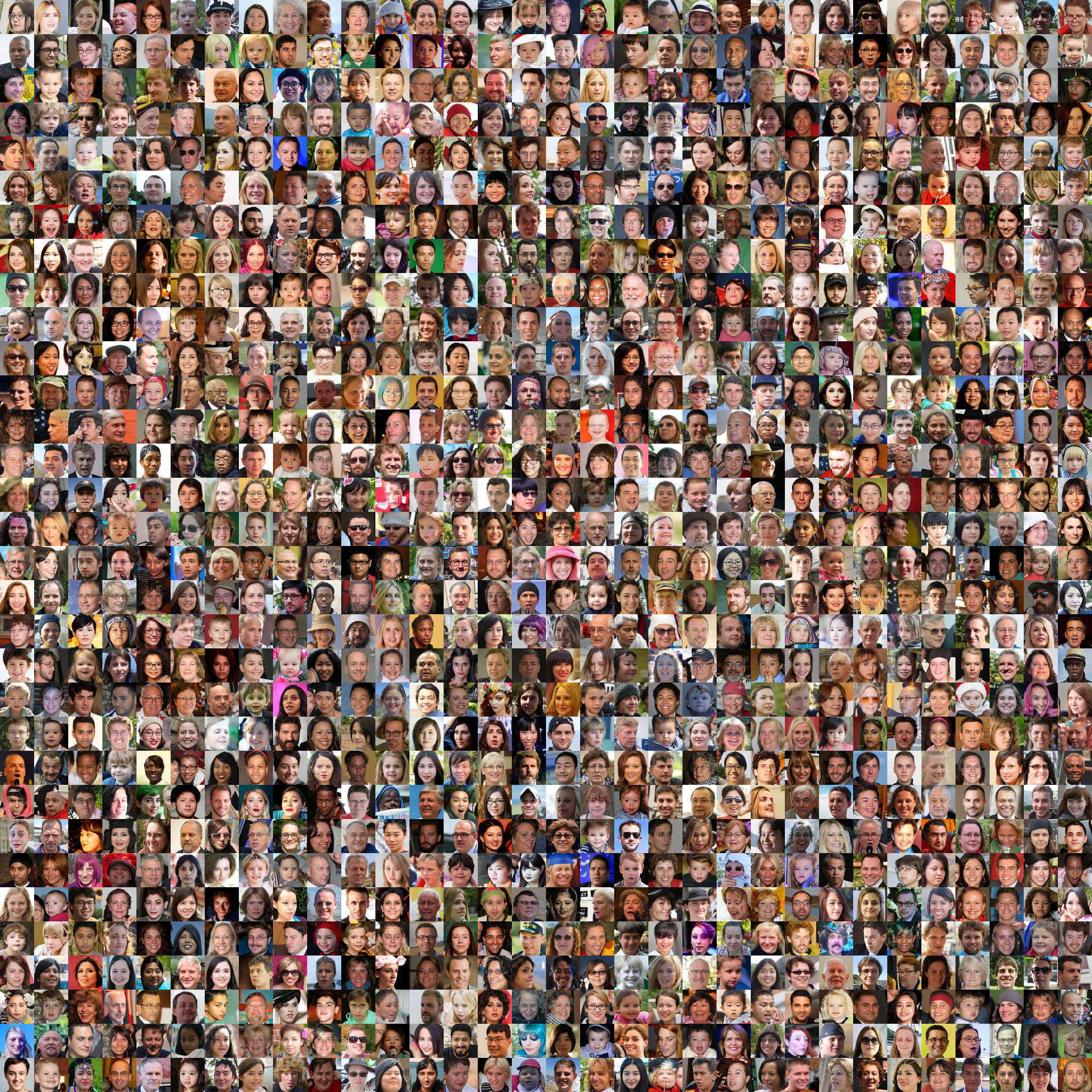}
    \caption{Qualitative examples of sample generation from our Config E on FFHQ-64.}
    \label{fig:ffhq64}
\end{figure}
}
\FloatBarrier
\input{figures/qualitative/cifar-10-32x32}
\FloatBarrier
\input{figures/qualitative/imgnet-32-32x32}
\FloatBarrier
{
\begin{figure}[ht!]
    \setlength{\imgsize}{0.2\linewidth} 
    
    \setlength{\tabcolsep}{0pt} 
    \renewcommand{\arraystretch}{0} 

    \centering
    
    \includegraphics[width=\linewidth]{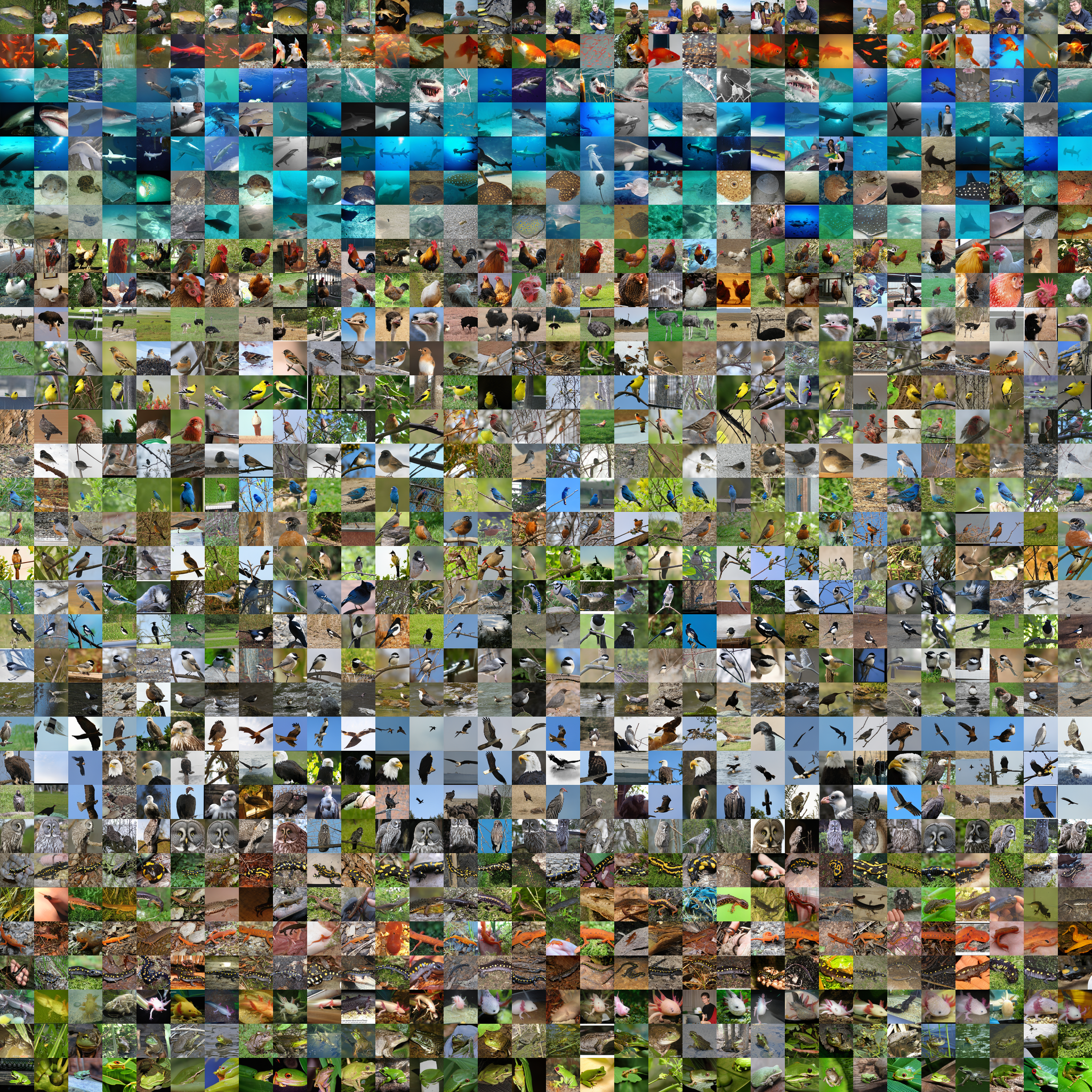}
    \caption{Qualitative examples of sample generation from our Config E on ImageNet-64.}
    \label{fig:imgnet-64}
\end{figure}
}
\FloatBarrier

\clearpage
\section{Training Curves}
\vspace{-0.25cm}
\begin{figure}[h!]
    \centering    
    \includegraphics[width=0.32\linewidth]{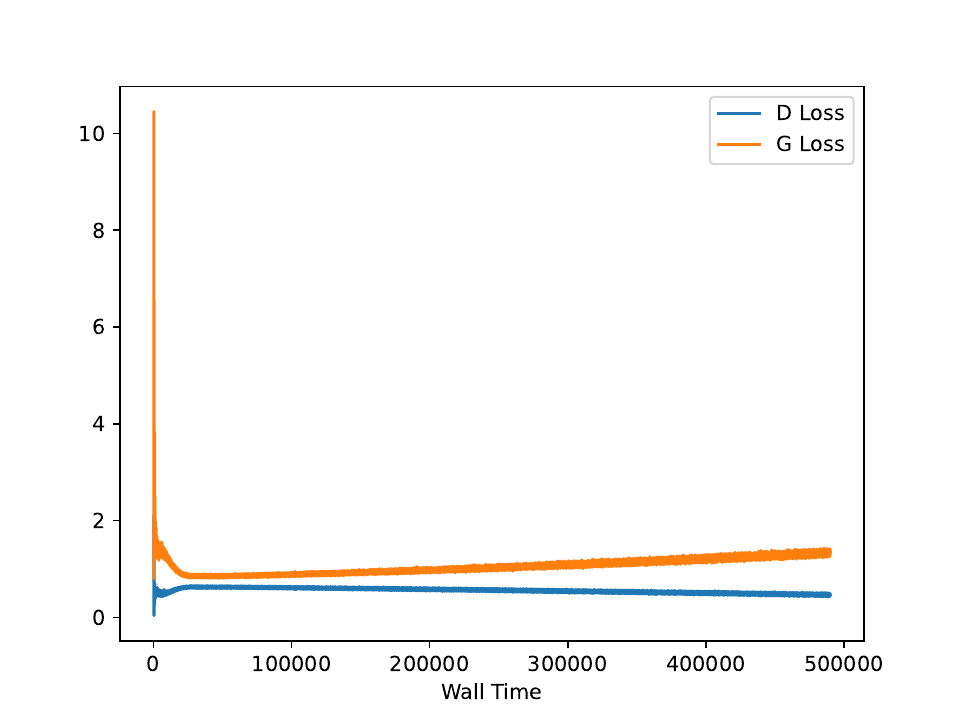}%
    \includegraphics[width=0.32\linewidth]{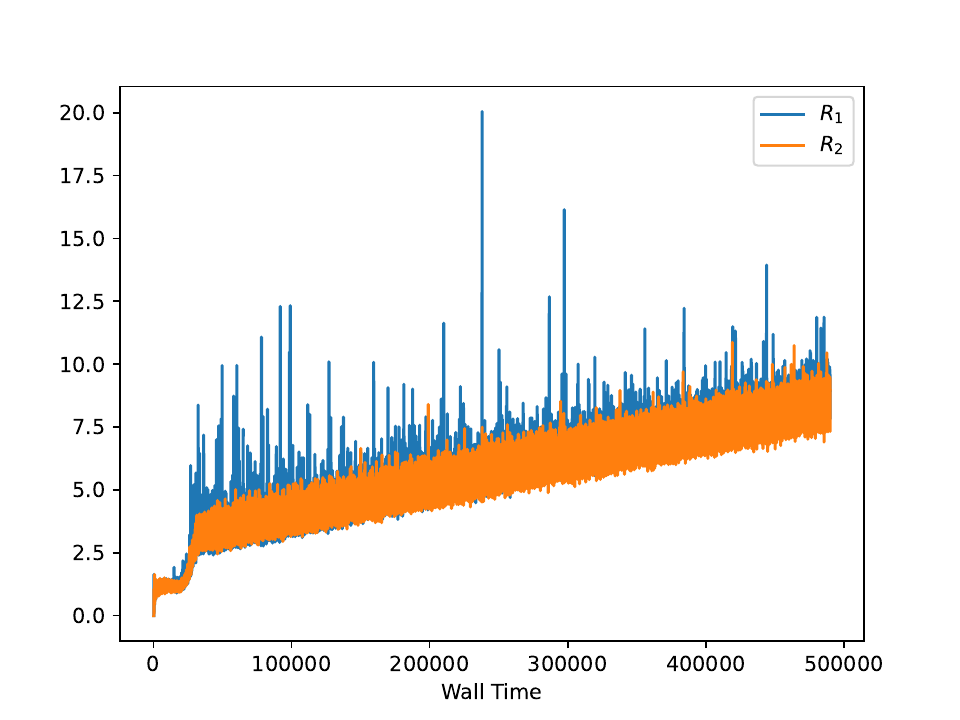}%
    \includegraphics[width=0.32\linewidth]{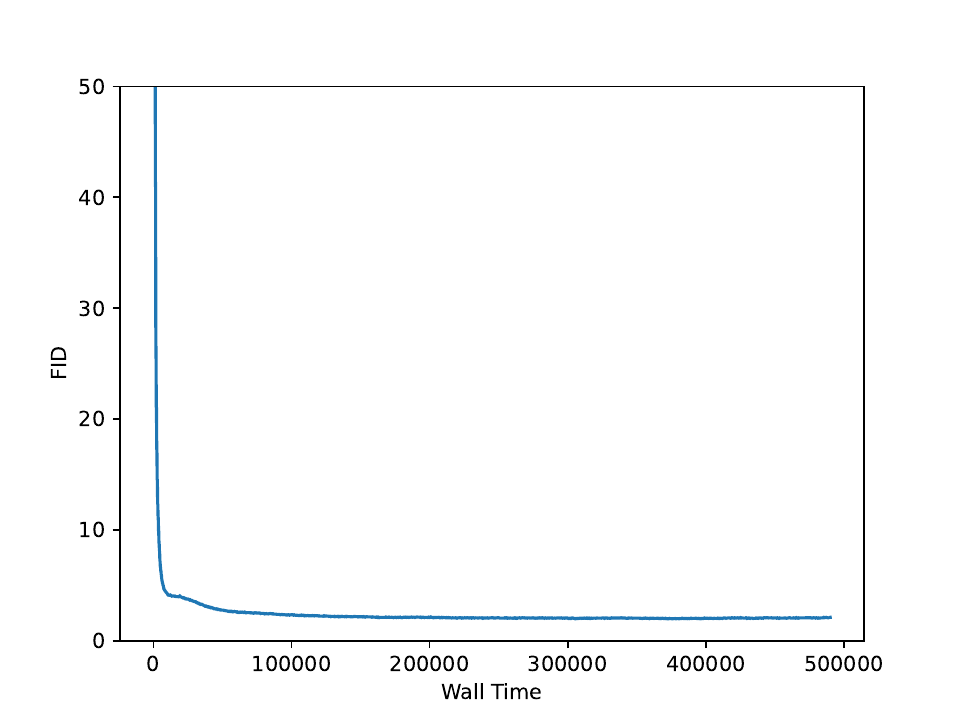}
    \caption{CIFAR-10 training curves.}
    \label{fig:cifar10_trainingcurve}
    \vspace{-0.3cm}
\end{figure}

\begin{figure}[h!]
    \centering
    \includegraphics[width=0.32\linewidth]{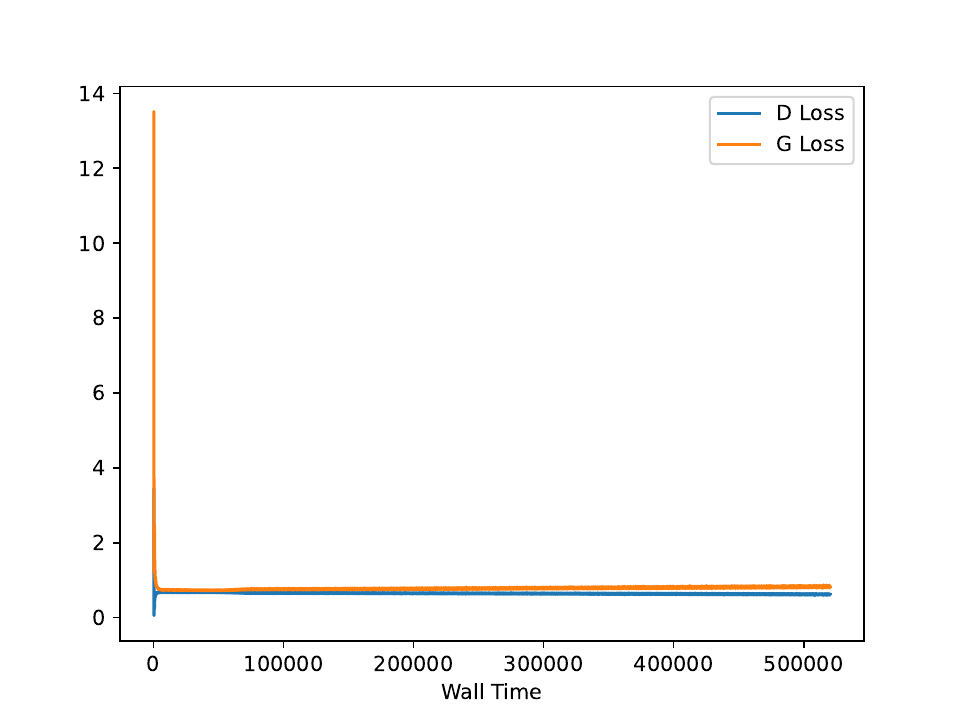}%
    \includegraphics[width=0.32\linewidth]{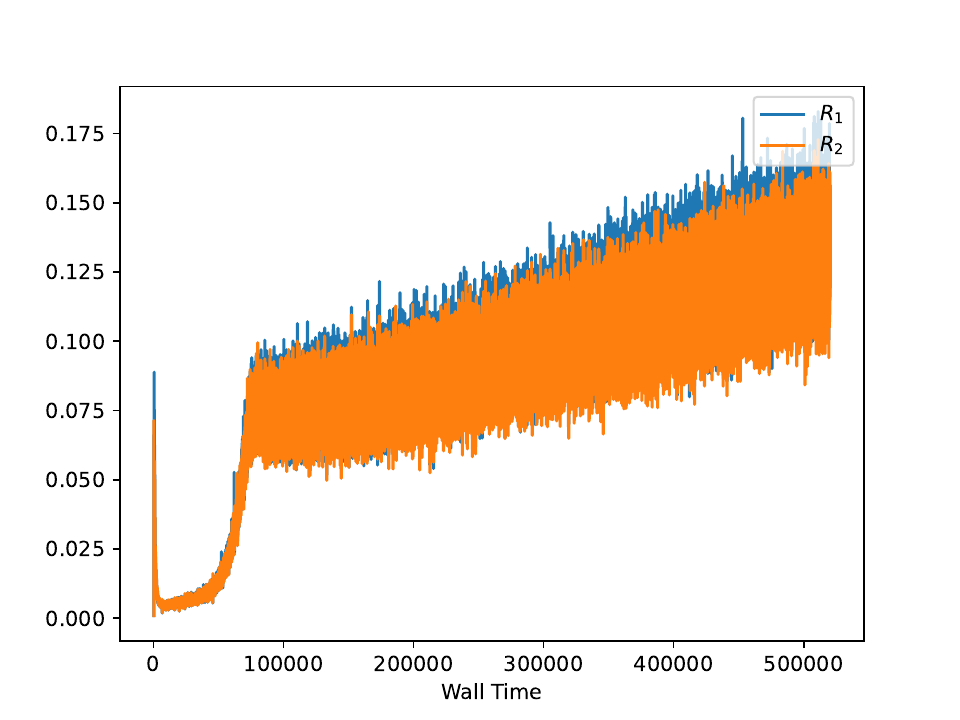}%
    \includegraphics[width=0.32\linewidth]{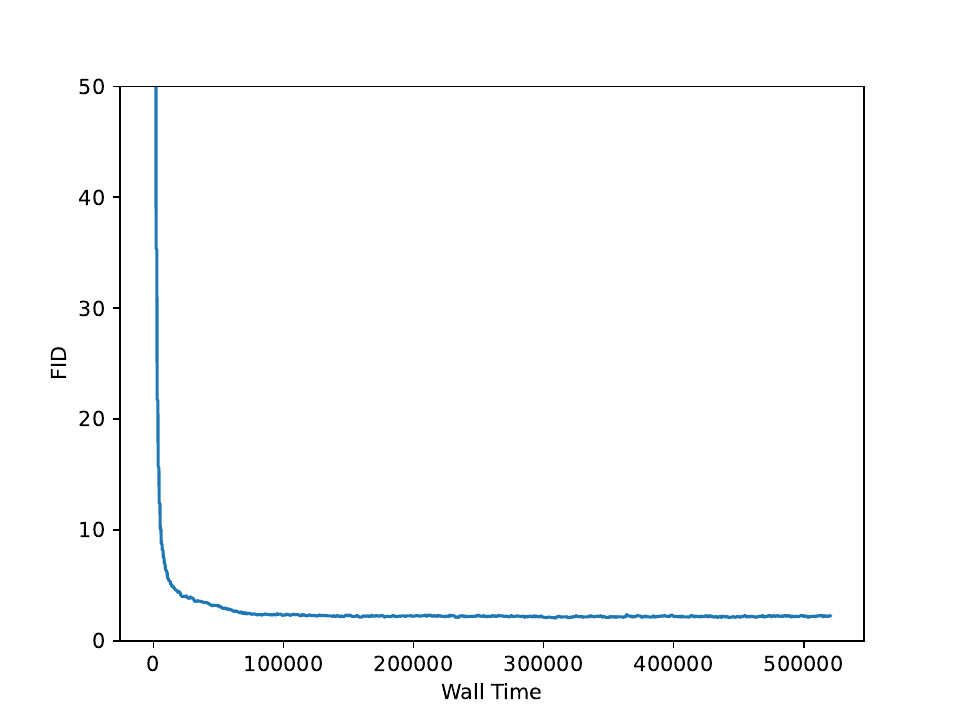}
    \caption{FFHQ-64 training curves.}
    \label{fig:ffhq64_trainingcurve}
    \vspace{-0.3cm}
\end{figure}

\begin{figure}[h!]
    \centering
    \includegraphics[width=0.32\linewidth]{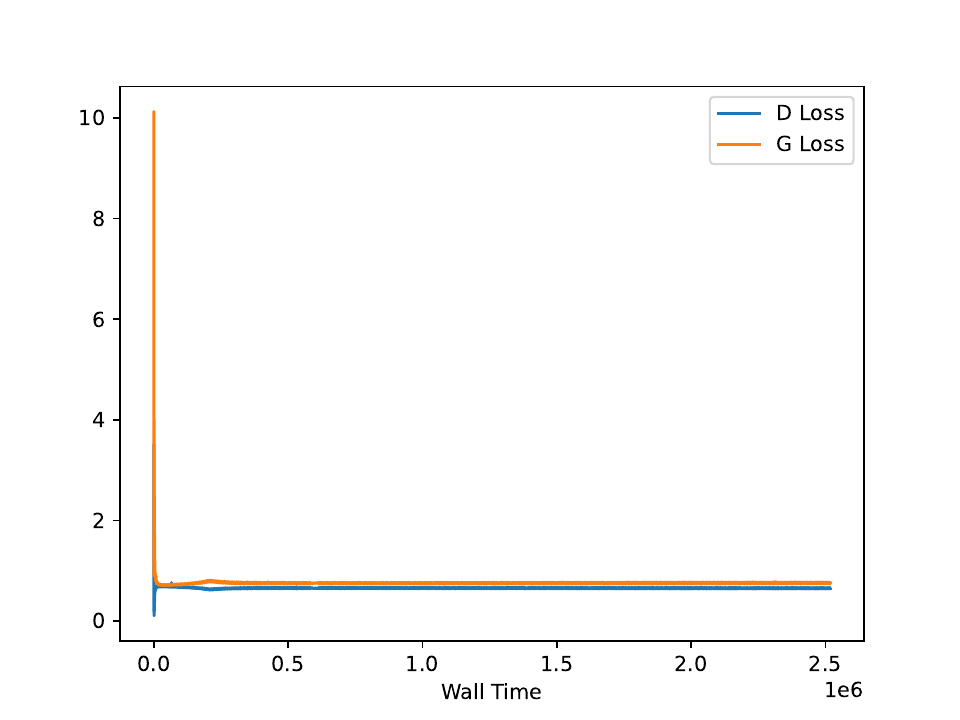}%
    \includegraphics[width=0.32\linewidth]{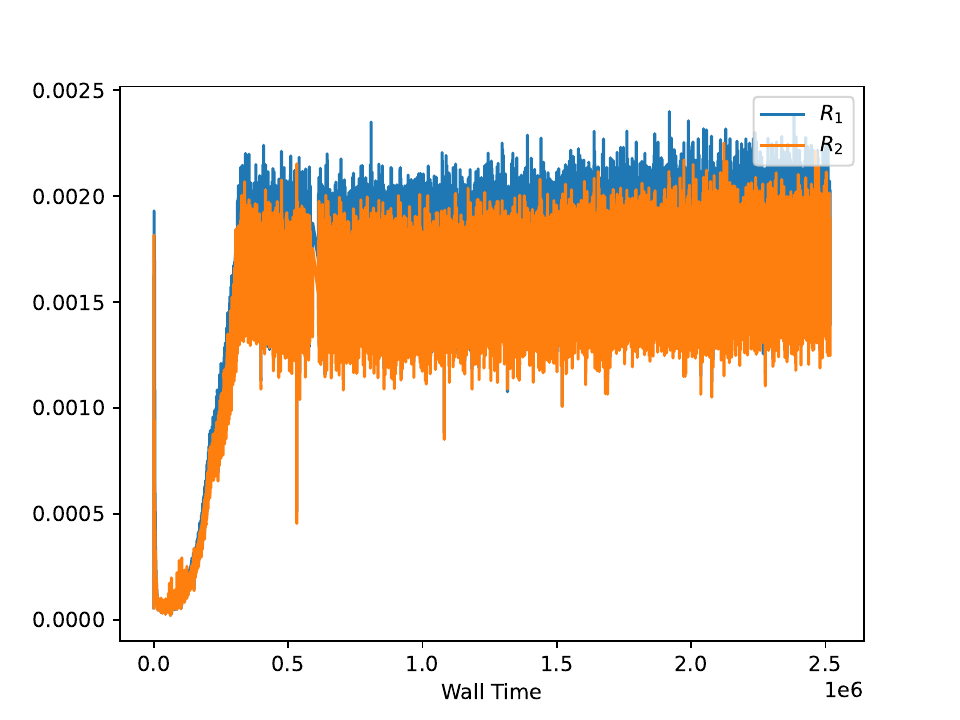}%
    \includegraphics[width=0.32\linewidth]{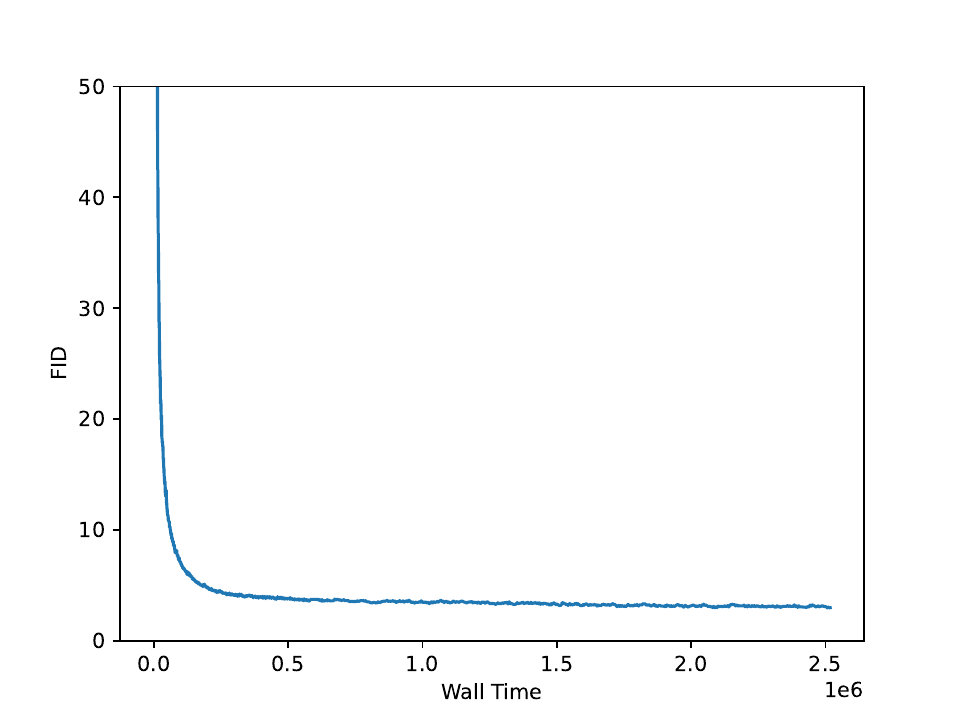}
    \caption{FFHQ-256 training curves.}
    \label{fig:ffhq256_trainingcurve}
    \vspace{-0.3cm}
\end{figure}

\begin{figure}[h!]
    \centering    
    \includegraphics[width=0.32\linewidth]{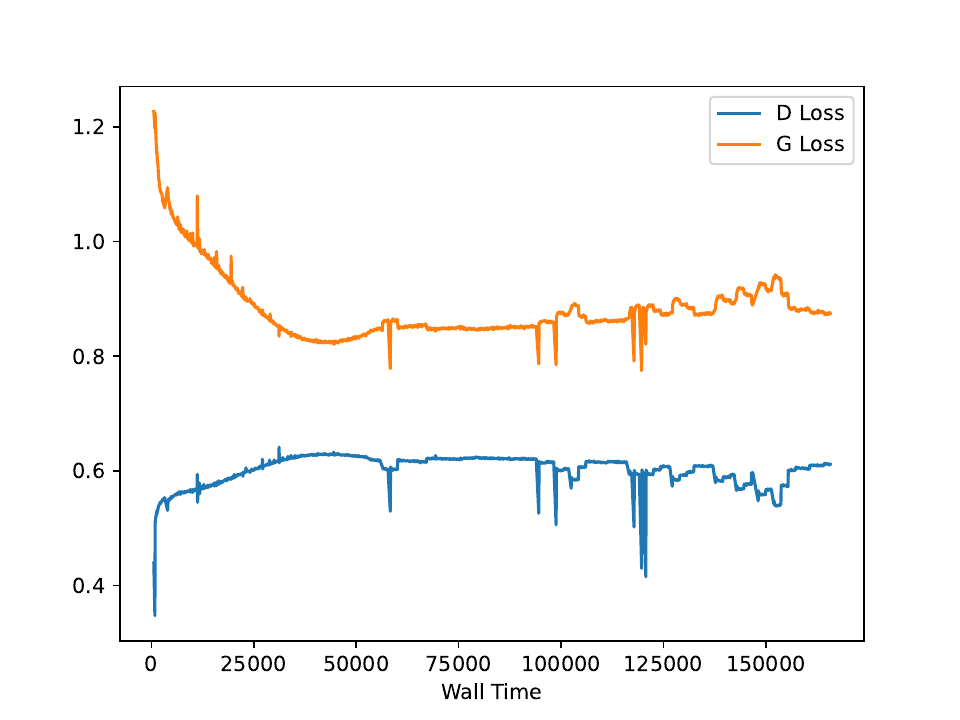}%
    \includegraphics[width=0.32\linewidth]{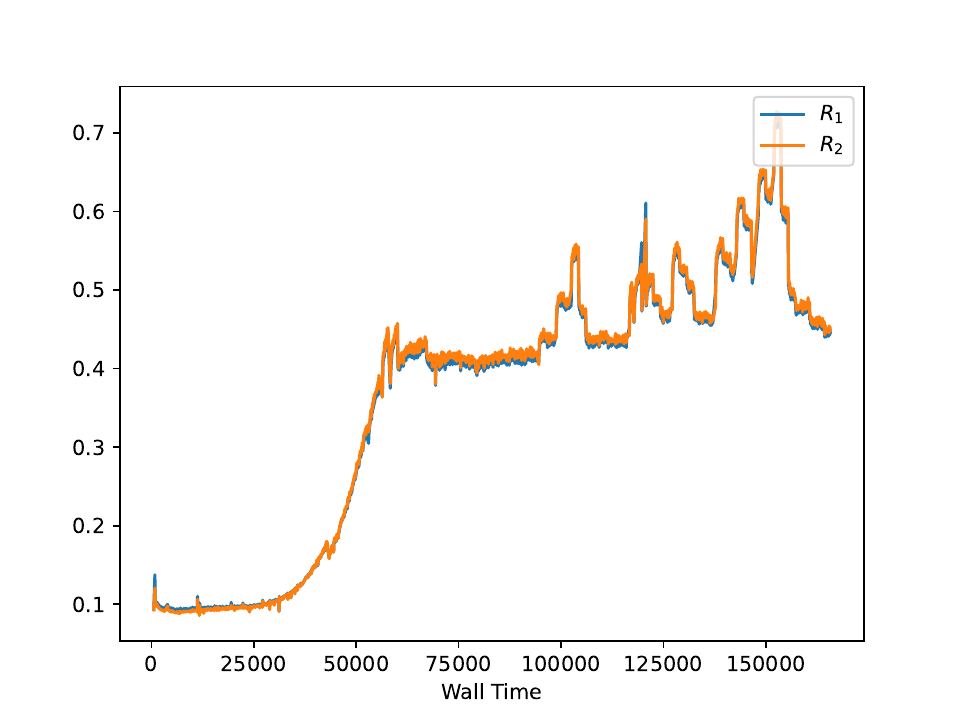}%
    \includegraphics[width=0.32\linewidth]{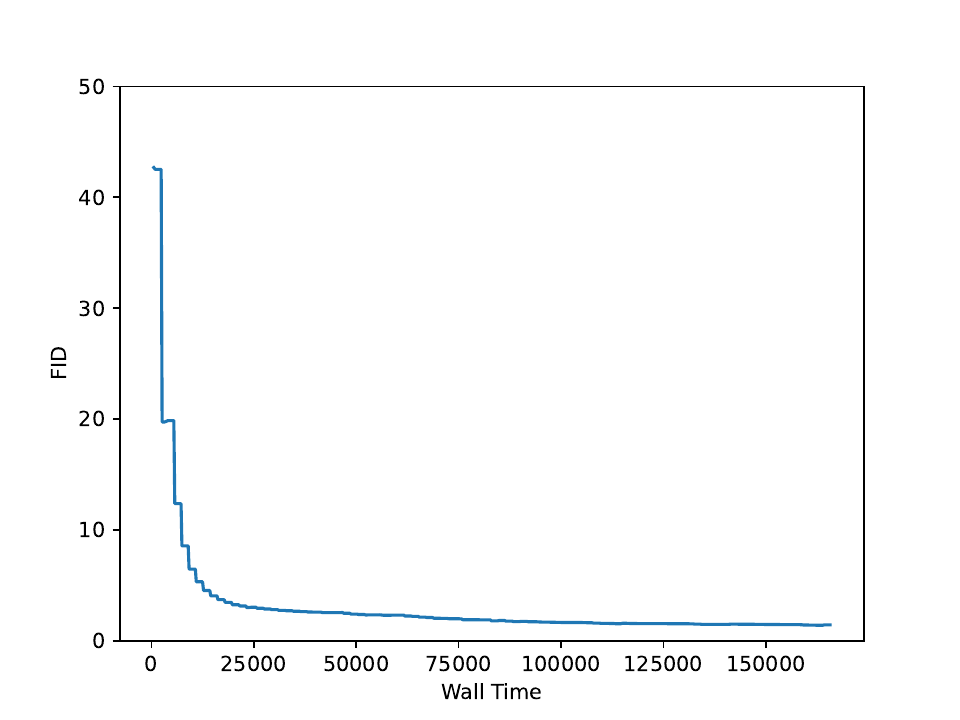}
    \caption{ImageNet-32 training curves.}
    \label{fig:imagenet32_trainingcurve}
    \vspace{-0.3cm}
\end{figure}

\begin{figure}[h!]
    \centering    
    \includegraphics[width=0.32\linewidth]{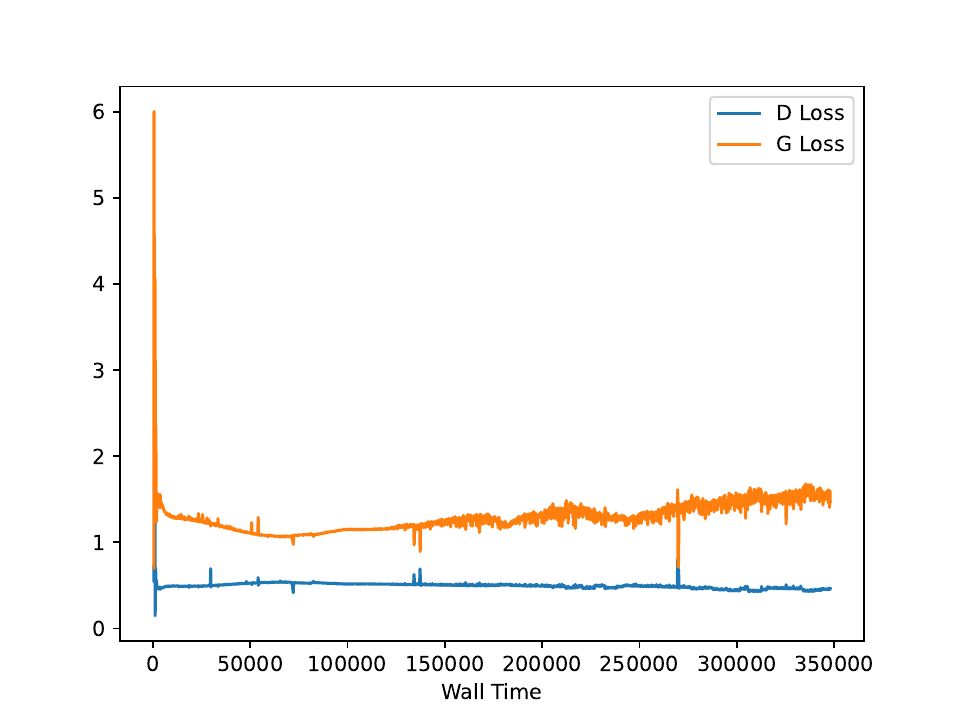}%
    \includegraphics[width=0.32\linewidth]{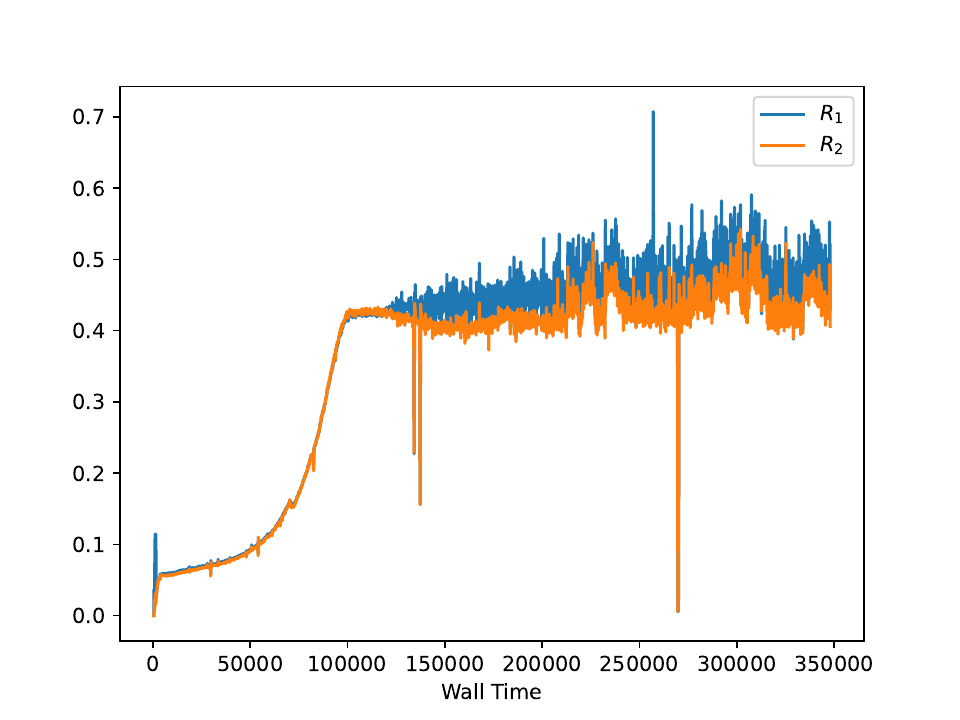}%
    \includegraphics[width=0.32\linewidth]{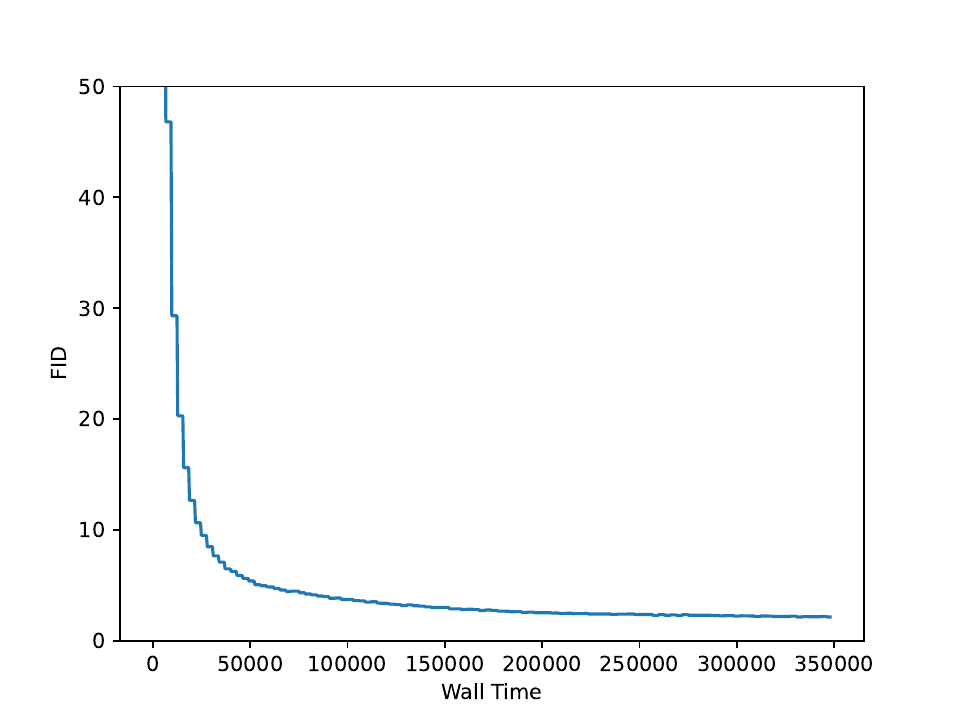}
    \caption{ImageNet-64 training curves.}
    \label{fig:imagenet64_trainingcurve}    \vspace{-0.3cm}
\end{figure}

%% file: tex/table_hyperparameters.tex
\begin{landscape}
\begin{table}[h]
\centering
\caption{\label{tab:hyperparam}Hyperparameters for each experiment. The decay factor $\beta$ of EMA can be obtained using the formula $\beta = 0.5^{\frac{\text{Minibatch size}}{\text{EMA half-life}}}$,~\eg for CIFAR-10, EMA $\beta=0.5^{\frac{512}{5\times10^6}}\approx0.9999$.}
\resizebox{\columnwidth}{!}{%
\begin{tblr}{
  column{even} = {c},
  column{3} = {c},
  column{5} = {c},
  column{7} = {c},
  cell{1}{4} = {c=2}{},
  cell{1}{6} = {c=2}{},
  hline{1-2,5,12,17,19} = {-}{},
}
Hyperparameter               & Stacked MNIST        & CIFAR-10                              & FFHQ                                  &                     & ImageNet                              &     \\
Resolution                   & $32\times32$         & $32\times32$                          & $256\times256$                        & $64\times64$                 & $32\times32$                          & $64\times64$ \\
Class conditional            & -                    & $\checkmark$                          & -                                     & -                   & $\checkmark$                          & $\checkmark$  \\
Number of GPUs               & 8                    & 8                                     & 8                                     & 8                   & 32                                    & 64 \\
Duration (Mimg)              & 10                   & 250                                   & 200                                   & 100                 & 1000                                   & 1000  \\
Burn-in (Mimg)               & 2                    & 20                                    & 20                                    & 20                  & 200                                   & 200  \\
Minibatch size               & 512                  & 512                                   & 256                                   & 256                 & 4096                                  & 4096  \\
Learning rate                & $2\times10^{-4}$       & $2\times10^{-4}\rightarrow5\times10^{-5}$ & $2\times10^{-4}\rightarrow5\times10^{-5}$ & $2\times10^{-4}\rightarrow5\times10^{-5}$                 & $2\times10^{-4}\rightarrow5\times10^{-5}$ & $2\times10^{-4}\rightarrow5\times10^{-5}$  \\
$\gamma$ for $R_1$ and $R_2$ & $1\rightarrow0.1$    & $0.05\rightarrow0.005$                & $150\rightarrow15$                    & $2\rightarrow0.2$                 & $0.5\rightarrow0.05$                  & $1\rightarrow0.1$  \\
Adam $\beta_2$               & $0.9\rightarrow0.99$ & $0.9\rightarrow0.99$                  & $0.9\rightarrow0.99$                  & $0.9\rightarrow0.99$                 & $0.9\rightarrow0.99$                  & $0.9\rightarrow0.99$  \\
EMA half-life (Mimg)         & $0\rightarrow0.5$    & $0\rightarrow5$                       & $0\rightarrow0.5$                     & $0\rightarrow0.5$                 & $0\rightarrow50$                      & $0\rightarrow50$  \\
Channels per resolution      & 768-768-768-768      & 768-768-768-768                       & 96-192-384-768-768-768-768            & 384-768-768-768-768 & 1536-1536-1536-1536                   & 1536-1536-1536-1536-1536  \\
ResBlocks per resolution     & 2-2-2-2              & 2-2-2-2                               & 2-2-2-2-2-2-2                         & 2-2-2-2-2                 & 2-2-2-2                               & 2-2-2-2-2  \\
Groups per resolution        & 96-96-96-96          & 96-96-96-96                           & 12-24-48-96-96-96-96                  & 48-96-96-96-96                 & 96-96-96-96                           & 96-96-96-96-96  \\
$G$~params                   & 20.73M               & 20.78M                                & 23.06M                                & 22.43M                 & 82.91M                                & 103.57M  \\
$D$~params                   & 20.68M               & 21.28M                                & 23.01M                                & 22.38M                 & 86.55M                                & 107.21M  \\
Dataset $x$-flips            & -                    & $\checkmark$                          & $\checkmark$                          & $\checkmark$                 & $\checkmark$                          & $\checkmark$  \\
Augment probability          & -                    & $0\rightarrow0.55$                    & $0\rightarrow0.3$                    & $0\rightarrow0.3$                 & $0\rightarrow0.5$                     & $0\rightarrow0.4$  
\end{tblr}
}
\end{table}
\end{landscape}

%% file: figures/qualitative/stacked-mnist-64x64.tex
{
\begin{figure}[h!]
    \setlength{\imgsize}{\linewidth} 
    
    \setlength{\tabcolsep}{0pt} 
    \renewcommand{\arraystretch}{0} 

    \newcommand{\qualitativeimg}[1]{%
        \includegraphics[width=\imgsize]{figures/qualitative/stacked-mnist-000008806/number-#1.jpg}%
    }
    \centering
    \includegraphics[width=\linewidth, clip, trim={0 0 768px 768px}]{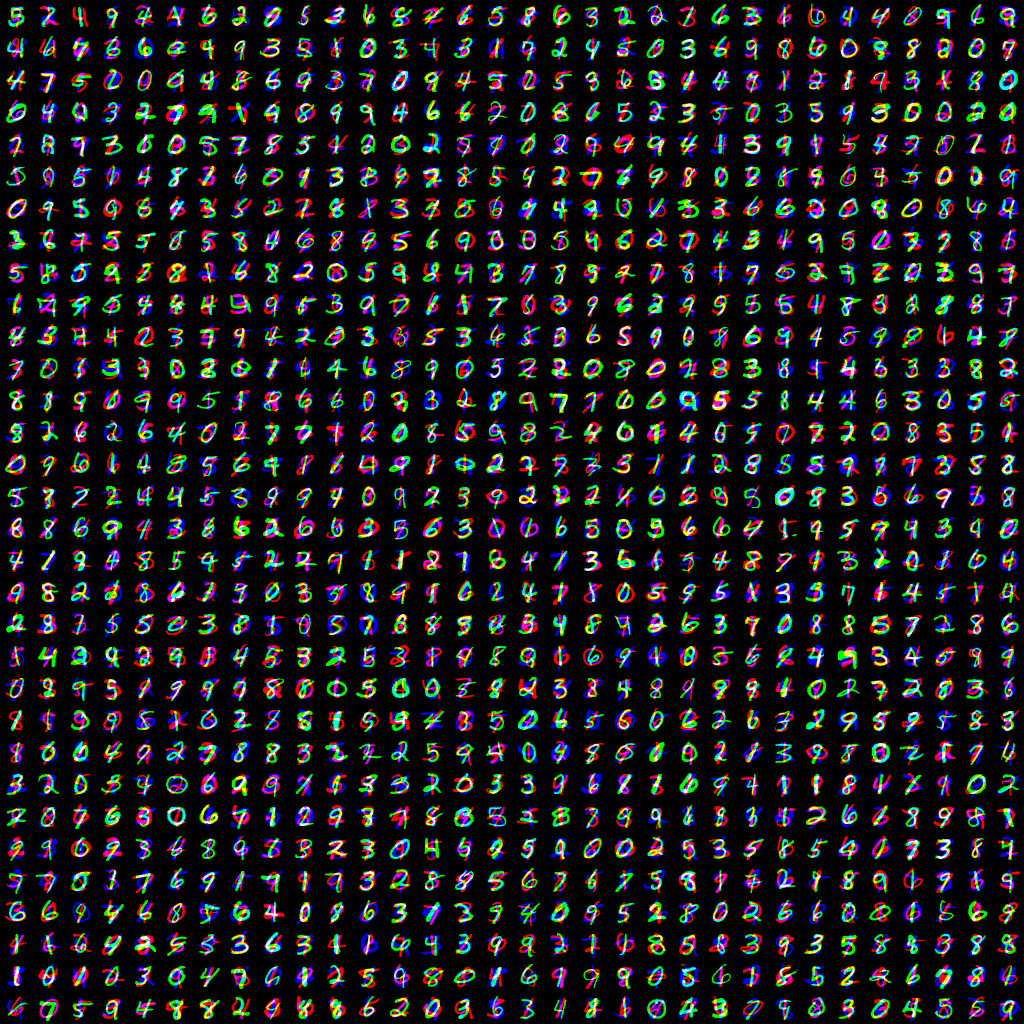}
    \caption{Qualitative examples of sample generation from our Config E on Stacked-MNIST.}
    \label{fig:stacked-mnist}
\end{figure}
}

%% file: figures/qualitative/ffhq-256-8x8.tex
{
\begin{figure}[ht!]
    \setlength{\imgsize}{0.125\linewidth} 
    
    \newcommand{\qualitativeimg}[1]{%
        \includegraphics[width=\imgsize]{figures/qualitative/ffhq-256-000139623/image-#1.jpg}%
    }
    
    \setlength{\tabcolsep}{0pt} 
    \renewcommand{\arraystretch}{0} 
    
    \centering
    \begin{tabular}{cccccccc} 
        \qualitativeimg{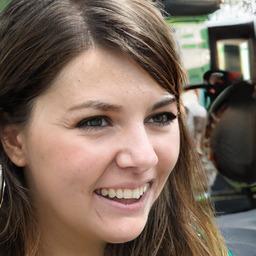} & \qualitativeimg{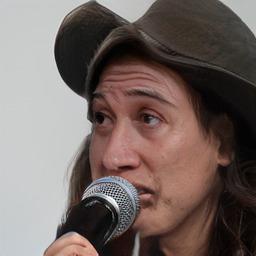} & \qualitativeimg{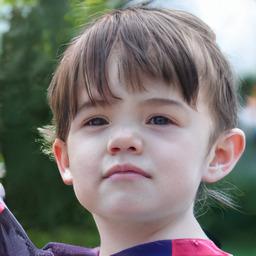} & \qualitativeimg{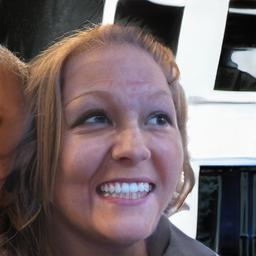} & \qualitativeimg{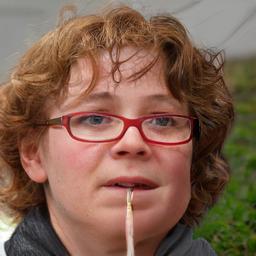} & \qualitativeimg{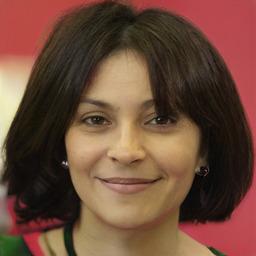} & \qualitativeimg{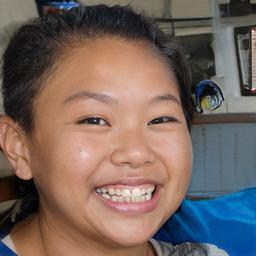} & \qualitativeimg{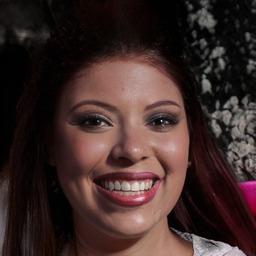} \\
        \qualitativeimg{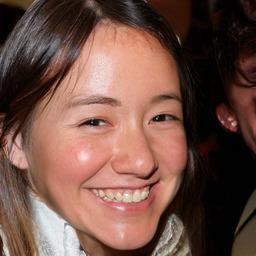} & \qualitativeimg{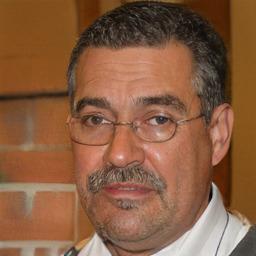} & \qualitativeimg{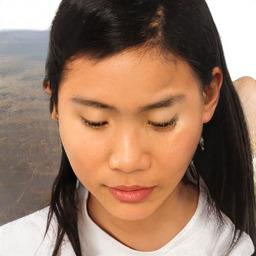} & \qualitativeimg{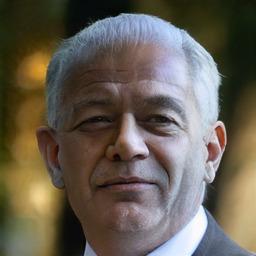} & \qualitativeimg{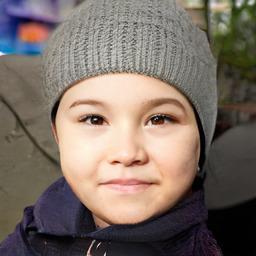} & \qualitativeimg{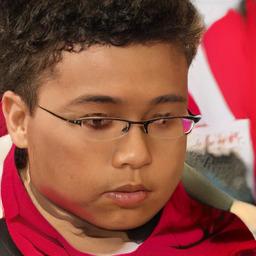} & \qualitativeimg{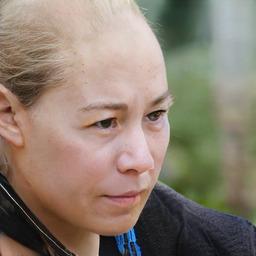} & \qualitativeimg{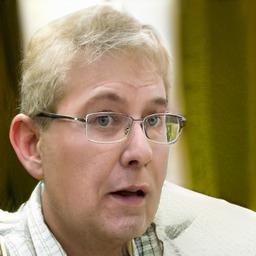} \\
        \qualitativeimg{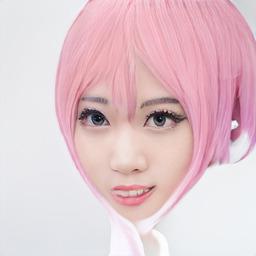} & \qualitativeimg{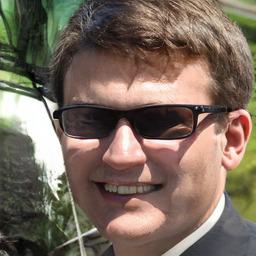} & \qualitativeimg{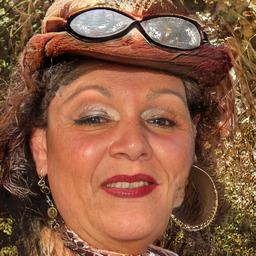} & \qualitativeimg{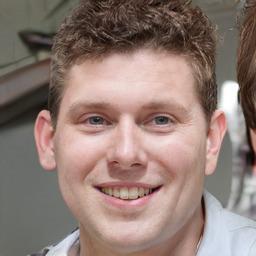} & \qualitativeimg{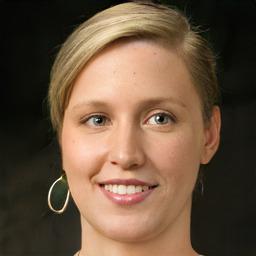} & \qualitativeimg{71} & \qualitativeimg{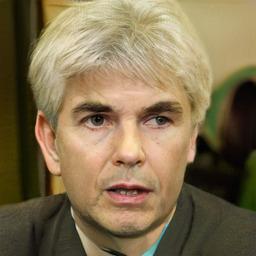} & \qualitativeimg{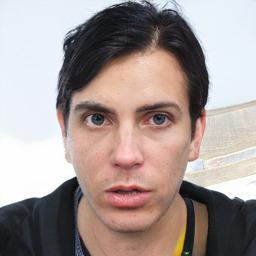} \\
        \qualitativeimg{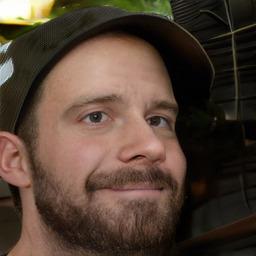} & \qualitativeimg{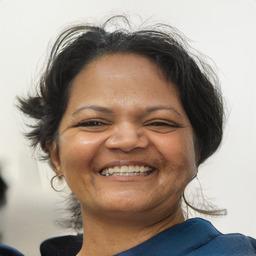} & \qualitativeimg{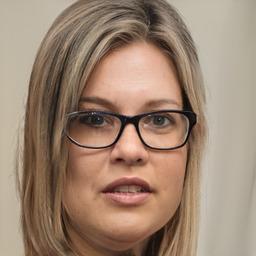} & \qualitativeimg{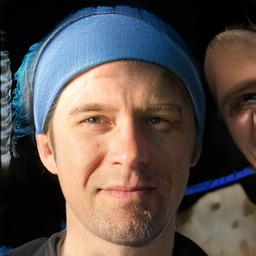} & \qualitativeimg{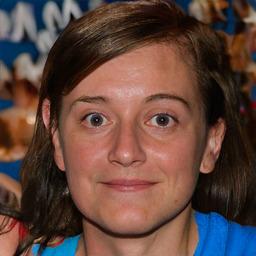} & \qualitativeimg{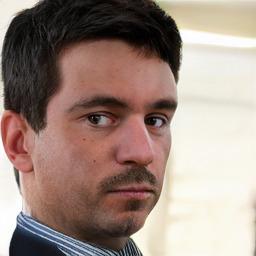} & \qualitativeimg{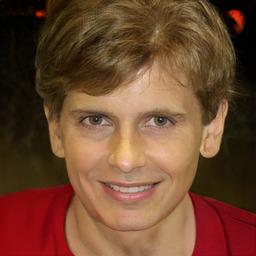} & \qualitativeimg{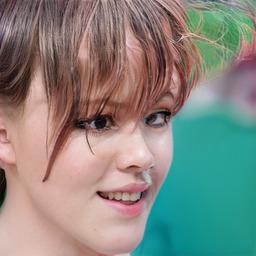} \\
        \qualitativeimg{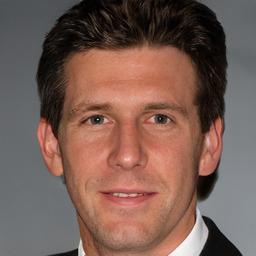} & \qualitativeimg{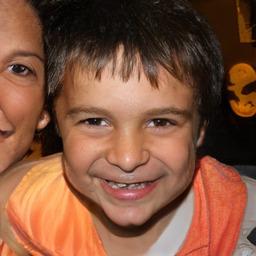} & \qualitativeimg{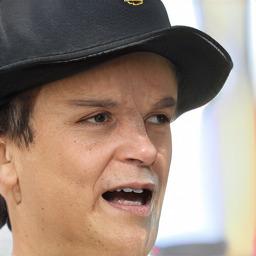} & \qualitativeimg{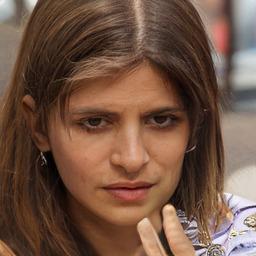} & \qualitativeimg{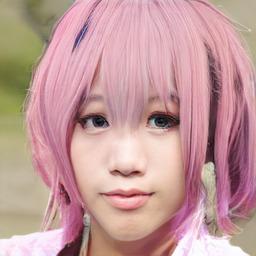} & \qualitativeimg{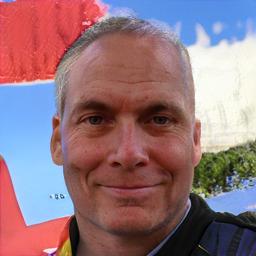} & \qualitativeimg{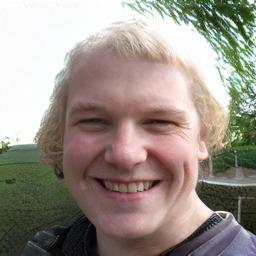} & \qualitativeimg{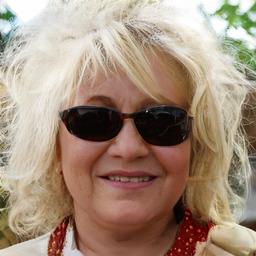} \\
        \qualitativeimg{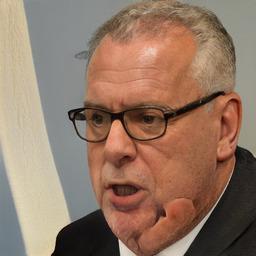} & \qualitativeimg{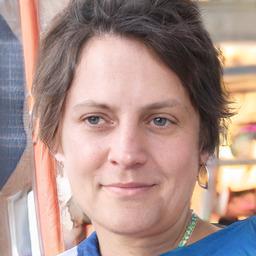} & \qualitativeimg{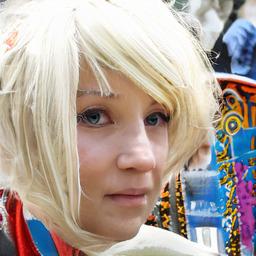} & \qualitativeimg{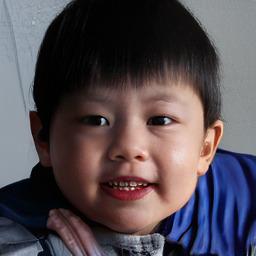} & \qualitativeimg{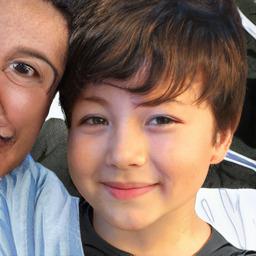} & \qualitativeimg{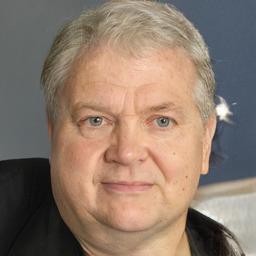} & \qualitativeimg{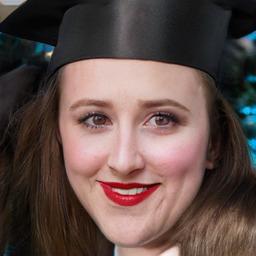} & \qualitativeimg{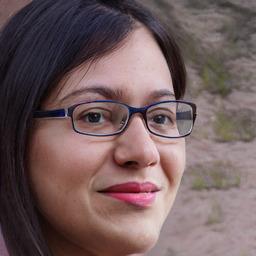} \\
        \qualitativeimg{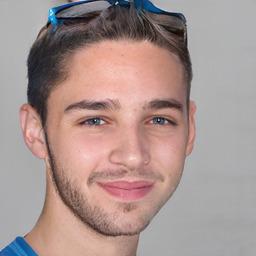} & \qualitativeimg{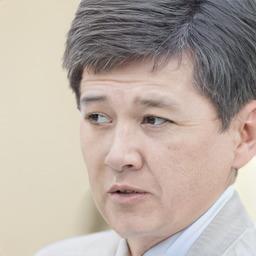} & \qualitativeimg{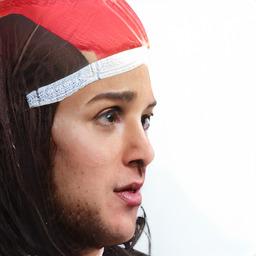} & \qualitativeimg{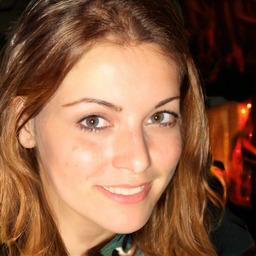} & \qualitativeimg{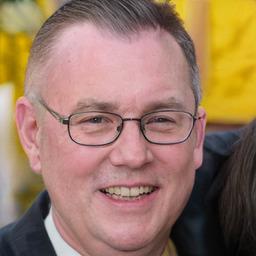} & \qualitativeimg{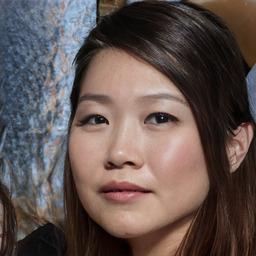} & \qualitativeimg{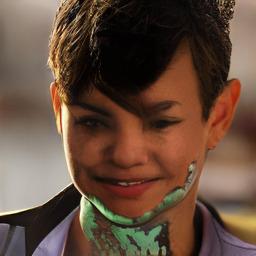} & \qualitativeimg{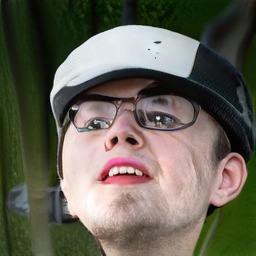} \\
        \qualitativeimg{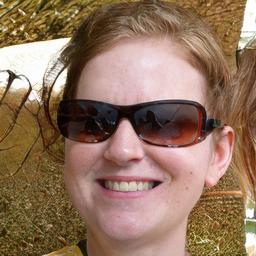} & \qualitativeimg{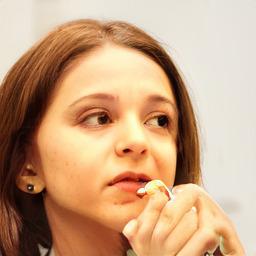} & \qualitativeimg{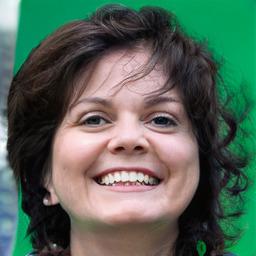} & \qualitativeimg{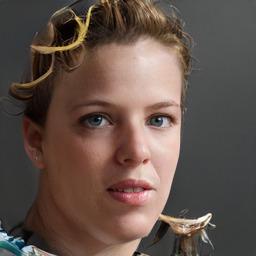} & \qualitativeimg{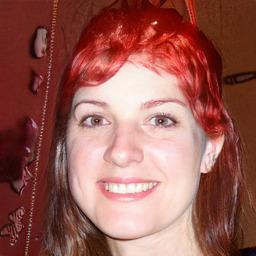} & \qualitativeimg{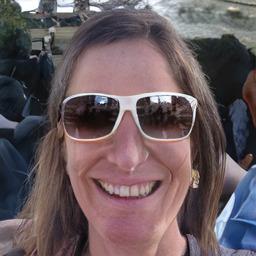} & \qualitativeimg{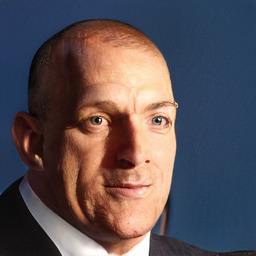} & \qualitativeimg{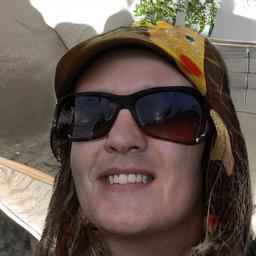} \\
    \end{tabular}
    \caption{More qualitative examples of sample generation from our Config E on FFHQ-256.}
    \label{fig:ffhq-256}
\end{figure}
}

%% file: figures/qualitative/cifar-10-32x32.tex
{
\begin{figure}[h!]
    \setlength{\imgsize}{0.2\linewidth} 
    
    \setlength{\tabcolsep}{0pt} 
    \renewcommand{\arraystretch}{0} 

    \centering
    
    \includegraphics[width=\linewidth]{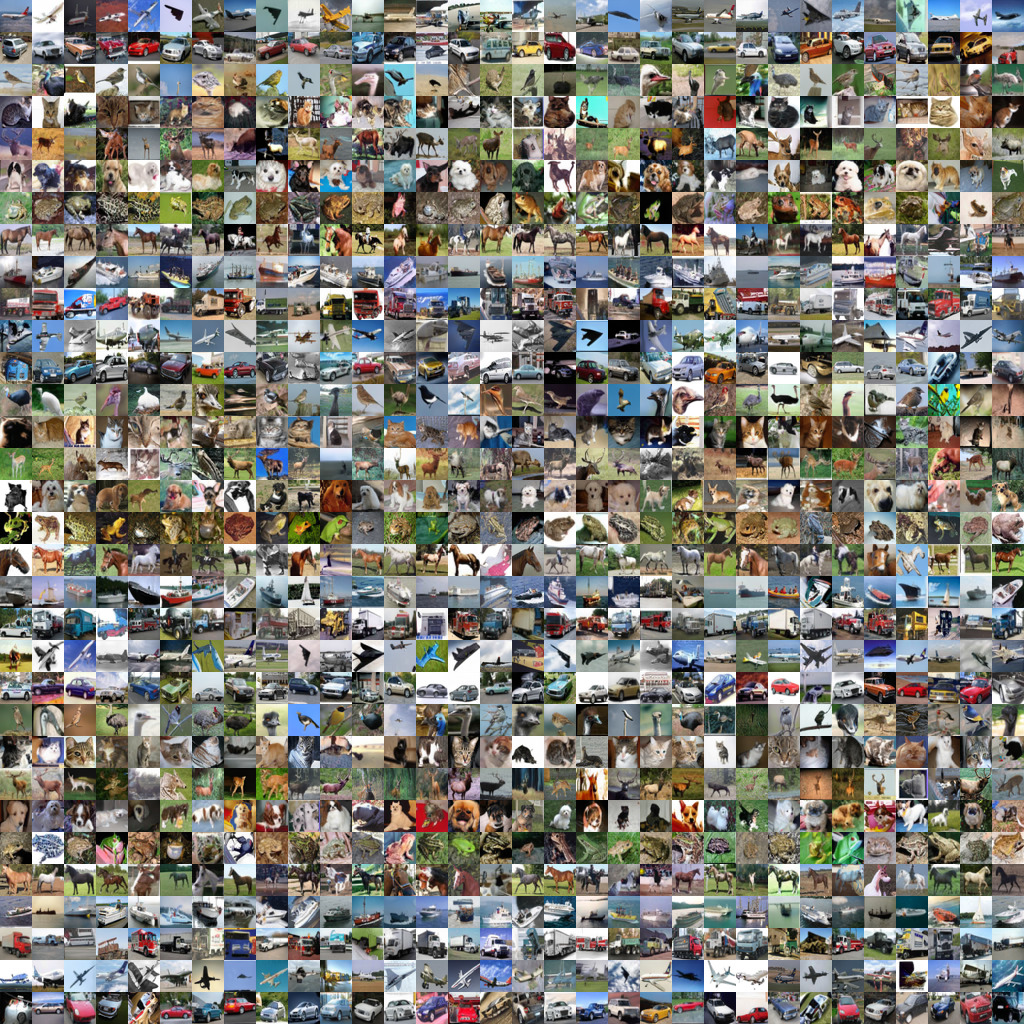}
    \caption{Qualitative examples of sample generation from our Config E on CIFAR-10.}
    \label{fig:cifar10}
\end{figure}
}

%% file: figures/qualitative/imgnet-32-32x32.tex
{
\begin{figure}[ht!]
    \setlength{\imgsize}{0.2\linewidth} 
    
    \setlength{\tabcolsep}{0pt} 
    \renewcommand{\arraystretch}{0} 

    \newcommand{\qualitativeimg}[1]{%
        \includegraphics[width=\imgsize]{figures/qualitative/stacked-mnist-000008806/number-#1.jpg}%
    }
    \centering
    
    \includegraphics[width=\linewidth]{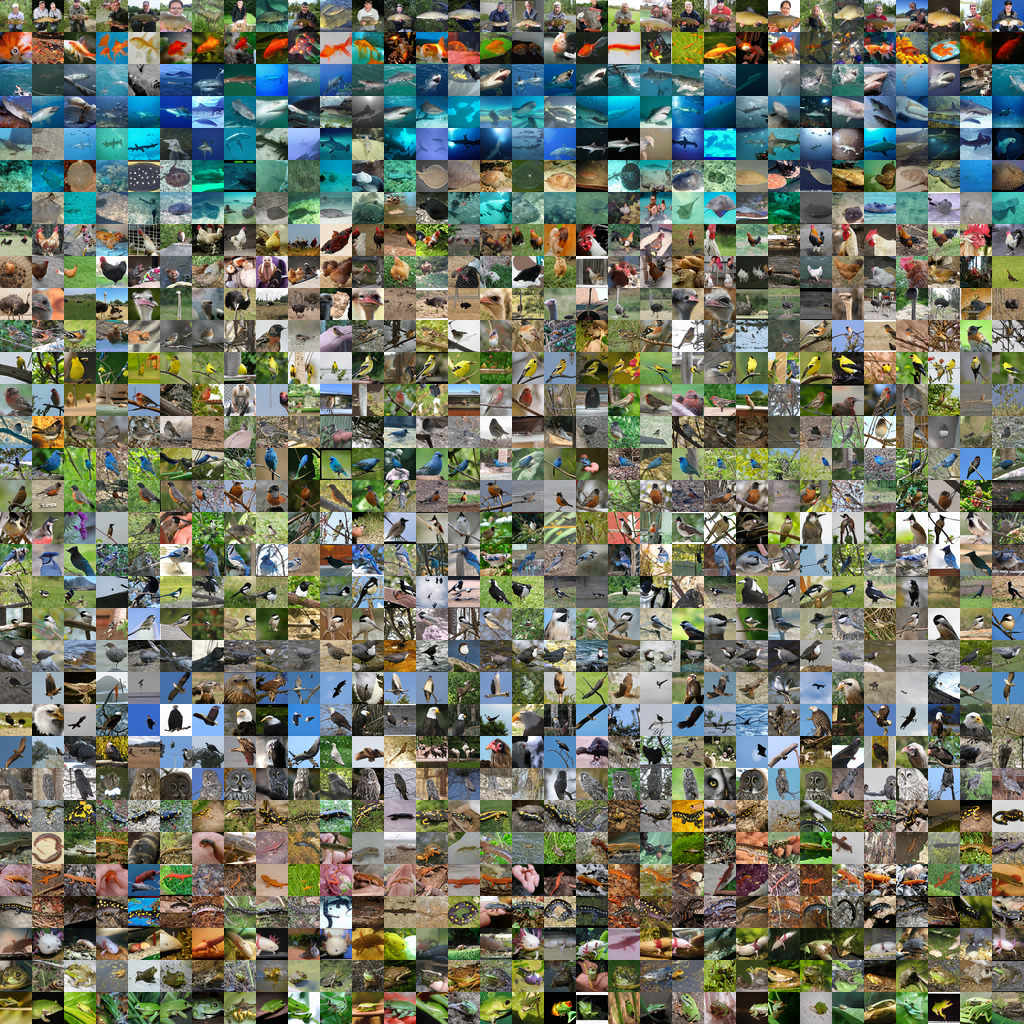}
    \caption{Qualitative examples of sample generation from our Config E on ImageNet-32.}
    \label{fig:imgnet-32}
\end{figure}
}

%% file: tex/checklist.tex
\pagebreak 
\section*{NeurIPS Paper Checklist}

\begin{enumerate}

\item {\bf Claims}
    \item[] Question: Do the main claims made in the abstract and introduction accurately reflect the paper's contributions and scope?
    \item[] Answer: \answerYes{} 
    \item[] Justification: Claim of stability is justified by Figure~\ref{fig:mnist_loss_curve} and later experimental performance. Claim of convergence properties is justified in Appendices A,B,C. Claim of SOTA GAN is experimentally justified in Section~\ref{sec:exp}. Claims are bound to specific datasets.
    \item[] Guidelines:
    \begin{itemize}
        \item The answer NA means that the abstract and introduction do not include the claims made in the paper.
        \item The abstract and/or introduction should clearly state the claims made, including the contributions made in the paper and important assumptions and limitations. A No or NA answer to this question will not be perceived well by the reviewers. 
        \item The claims made should match theoretical and experimental results, and reflect how much the results can be expected to generalize to other settings. 
        \item It is fine to include aspirational goals as motivation as long as it is clear that these goals are not attained by the paper. 
    \end{itemize}

\item {\bf Limitations}
    \item[] Question: Does the paper discuss the limitations of the work performed by the authors?
    \item[] Answer: \answerYes{} 
    \item[] Justification: Please see Section 5.
    \item[] Guidelines:
    \begin{itemize}
        \item The answer NA means that the paper has no limitation while the answer No means that the paper has limitations, but those are not discussed in the paper. 
        \item The authors are encouraged to create a separate "Limitations" section in their paper.
        \item The paper should point out any strong assumptions and how robust the results are to violations of these assumptions (e.g., independence assumptions, noiseless settings, model well-specification, asymptotic approximations only holding locally). The authors should reflect on how these assumptions might be violated in practice and what the implications would be.
        \item The authors should reflect on the scope of the claims made, e.g., if the approach was only tested on a few datasets or with a few runs. In general, empirical results often depend on implicit assumptions, which should be articulated.
        \item The authors should reflect on the factors that influence the performance of the approach. For example, a facial recognition algorithm may perform poorly when image resolution is low or images are taken in low lighting. Or a speech-to-text system might not be used reliably to provide closed captions for online lectures because it fails to handle technical jargon.
        \item The authors should discuss the computational efficiency of the proposed algorithms and how they scale with dataset size.
        \item If applicable, the authors should discuss possible limitations of their approach to address problems of privacy and fairness.
        \item While the authors might fear that complete honesty about limitations might be used by reviewers as grounds for rejection, a worse outcome might be that reviewers discover limitations that aren't acknowledged in the paper. The authors should use their best judgment and recognize that individual actions in favor of transparency play an important role in developing norms that preserve the integrity of the community. Reviewers will be specifically instructed to not penalize honesty concerning limitations.
    \end{itemize}

\item {\bf Theory Assumptions and Proofs}
    \item[] Question: For each theoretical result, does the paper provide the full set of assumptions and a complete (and correct) proof?
    \item[] Answer: \answerYes{} 
    \item[] Justification: Prior knowledge of Mescheder~\etal~\cite{r1}~is required, but this is cited appropriately to help the reader.
    \item[] Guidelines:
    \begin{itemize}
        \item The answer NA means that the paper does not include theoretical results. 
        \item All the theorems, formulas, and proofs in the paper should be numbered and cross-referenced.
        \item All assumptions should be clearly stated or referenced in the statement of any theorems.
        \item The proofs can either appear in the main paper or the supplemental material, but if they appear in the supplemental material, the authors are encouraged to provide a short proof sketch to provide intuition. 
        \item Inversely, any informal proof provided in the core of the paper should be complemented by formal proofs provided in appendix or supplemental material.
        \item Theorems and Lemmas that the proof relies upon should be properly referenced. 
    \end{itemize}

    \item {\bf Experimental Result Reproducibility}
    \item[] Question: Does the paper fully disclose all the information needed to reproduce the main experimental results of the paper to the extent that it affects the main claims and/or conclusions of the paper (regardless of whether the code and data are provided or not)?
    \item[] Answer: \answerYes{} 
    \item[] Justification: Supplemental table lists all hyperparamters, and a supplemental section describes the training configurations. 
    \item[] Guidelines:
    \begin{itemize}
        \item The answer NA means that the paper does not include experiments.
        \item If the paper includes experiments, a No answer to this question will not be perceived well by the reviewers: Making the paper reproducible is important, regardless of whether the code and data are provided or not.
        \item If the contribution is a dataset and/or model, the authors should describe the steps taken to make their results reproducible or verifiable. 
        \item Depending on the contribution, reproducibility can be accomplished in various ways. For example, if the contribution is a novel architecture, describing the architecture fully might suffice, or if the contribution is a specific model and empirical evaluation, it may be necessary to either make it possible for others to replicate the model with the same dataset, or provide access to the model. In general. releasing code and data is often one good way to accomplish this, but reproducibility can also be provided via detailed instructions for how to replicate the results, access to a hosted model (e.g., in the case of a large language model), releasing of a model checkpoint, or other means that are appropriate to the research performed.
        \item While NeurIPS does not require releasing code, the conference does require all submissions to provide some reasonable avenue for reproducibility, which may depend on the nature of the contribution. For example
        \begin{enumerate}
            \item If the contribution is primarily a new algorithm, the paper should make it clear how to reproduce that algorithm.
            \item If the contribution is primarily a new model architecture, the paper should describe the architecture clearly and fully.
            \item If the contribution is a new model (e.g., a large language model), then there should either be a way to access this model for reproducing the results or a way to reproduce the model (e.g., with an open-source dataset or instructions for how to construct the dataset).
            \item We recognize that reproducibility may be tricky in some cases, in which case authors are welcome to describe the particular way they provide for reproducibility. In the case of closed-source models, it may be that access to the model is limited in some way (e.g., to registered users), but it should be possible for other researchers to have some path to reproducing or verifying the results.
        \end{enumerate}
    \end{itemize}

\newpage
\item {\bf Open access to data and code}
    \item[] Question: Does the paper provide open access to the data and code, with sufficient instructions to faithfully reproduce the main experimental results, as described in supplemental material?
    \item[] Answer: \answerNo{} 
    \item[] Justification: There is no new data. There is no code at submission time. The authors will aim to release this by publication time, with instructions to faithfully reproduce the experiments. Code URL is included in abstract.
    \item[] Guidelines:
    \begin{itemize}
        \item The answer NA means that paper does not include experiments requiring code.
        \item Please see the NeurIPS code and data submission guidelines (\url{https://nips.cc/public/guides/CodeSubmissionPolicy}) for more details.
        \item While we encourage the release of code and data, we understand that this might not be possible, so “No” is an acceptable answer. Papers cannot be rejected simply for not including code, unless this is central to the contribution (e.g., for a new open-source benchmark).
        \item The instructions should contain the exact command and environment needed to run to reproduce the results. See the NeurIPS code and data submission guidelines (\url{https://nips.cc/public/guides/CodeSubmissionPolicy}) for more details.
        \item The authors should provide instructions on data access and preparation, including how to access the raw data, preprocessed data, intermediate data, and generated data, etc.
        \item The authors should provide scripts to reproduce all experimental results for the new proposed method and baselines. If only a subset of experiments are reproducible, they should state which ones are omitted from the script and why.
        \item At submission time, to preserve anonymity, the authors should release anonymized versions (if applicable).
        \item Providing as much information as possible in supplemental material (appended to the paper) is recommended, but including URLs to data and code is permitted.
    \end{itemize}

\item {\bf Experimental Setting/Details}
    \item[] Question: Does the paper specify all the training and test details (e.g., data splits, hyperparameters, how they were chosen, type of optimizer, etc.) necessary to understand the results?
    \item[] Answer: \answerYes{} 
    \item[] Justification: Supplemental table lists all hyperparamters, and a supplemental section describes the training configurations.
    \item[] Guidelines:
    \begin{itemize}
        \item The answer NA means that the paper does not include experiments.
        \item The experimental setting should be presented in the core of the paper to a level of detail that is necessary to appreciate the results and make sense of them.
        \item The full details can be provided either with the code, in appendix, or as supplemental material.
    \end{itemize}

\item {\bf Experiment Statistical Significance}
    \item[] Question: Does the paper report error bars suitably and correctly defined or other appropriate information about the statistical significance of the experiments?
    \item[] Answer: \answerNo{} 
    \item[] Justification: Each experiment takes many days to compute, some take weeks. We do not have the compute time to provide variance bars on training executions.
    \item[] Guidelines:
    \begin{itemize}
        \item The answer NA means that the paper does not include experiments.
        \item The authors should answer "Yes" if the results are accompanied by error bars, confidence intervals, or statistical significance tests, at least for the experiments that support the main claims of the paper.
        \item The factors of variability that the error bars are capturing should be clearly stated (for example, train/test split, initialization, random drawing of some parameter, or overall run with given experimental conditions).
        \item The method for calculating the error bars should be explained (closed form formula, call to a library function, bootstrap, etc.)
        \item The assumptions made should be given (e.g., Normally distributed errors).
        \item It should be clear whether the error bar is the standard deviation or the standard error of the mean.
        \item It is OK to report 1-sigma error bars, but one should state it. The authors should preferably report a 2-sigma error bar than state that they have a 96\% CI, if the hypothesis of Normality of errors is not verified.
        \item For asymmetric distributions, the authors should be careful not to show in tables or figures symmetric error bars that would yield results that are out of range (e.g. negative error rates).
        \item If error bars are reported in tables or plots, The authors should explain in the text how they were calculated and reference the corresponding figures or tables in the text.
    \end{itemize}

\item {\bf Experiments Compute Resources}
    \item[] Question: For each experiment, does the paper provide sufficient information on the computer resources (type of compute workers, memory, time of execution) needed to reproduce the experiments?
    \item[] Answer: \answerYes{} 
    \item[] Justification: Please see supplemental section on the experimental setting.
    \item[] Guidelines:
    \begin{itemize}
        \item The answer NA means that the paper does not include experiments.
        \item The paper should indicate the type of compute workers CPU or GPU, internal cluster, or cloud provider, including relevant memory and storage.
        \item The paper should provide the amount of compute required for each of the individual experimental runs as well as estimate the total compute. 
        \item The paper should disclose whether the full research project required more compute than the experiments reported in the paper (e.g., preliminary or failed experiments that didn't make it into the paper). 
    \end{itemize}
    
\item {\bf Code Of Ethics}
    \item[] Question: Does the research conducted in the paper conform, in every respect, with the NeurIPS Code of Ethics \url{https://neurips.cc/public/EthicsGuidelines}?
    \item[] Answer: \answerYes{} 
    \item[] Justification: Experimental settings are standard and within the norms of the community.
    \item[] Guidelines:
    \begin{itemize}
        \item The answer NA means that the authors have not reviewed the NeurIPS Code of Ethics.
        \item If the authors answer No, they should explain the special circumstances that require a deviation from the Code of Ethics.
        \item The authors should make sure to preserve anonymity (e.g., if there is a special consideration due to laws or regulations in their jurisdiction).
    \end{itemize}

\item {\bf Broader Impacts}
    \item[] Question: Does the paper discuss both potential positive societal impacts and negative societal impacts of the work performed?
    \item[] Answer: \answerYes{} 
    \item[] Justification: We mention it briefly in Section 5. The paper describes a basic machine learning methodology, and so does not address a specific application with specific societal impacts. But, GANs do have potential social impact; it is clear that face generation has a significant impact (e.g., deep fakes) and our paper does use a face database for evaluation thanks to it being a community norm.
    \item[] Guidelines:
    \begin{itemize}
        \item The answer NA means that there is no societal impact of the work performed.
        \item If the authors answer NA or No, they should explain why their work has no societal impact or why the paper does not address societal impact.
        \item Examples of negative societal impacts include potential malicious or unintended uses (e.g., disinformation, generating fake profiles, surveillance), fairness considerations (e.g., deployment of technologies that could make decisions that unfairly impact specific groups), privacy considerations, and security considerations.
        \item The conference expects that many papers will be foundational research and not tied to particular applications, let alone deployments. However, if there is a direct path to any negative applications, the authors should point it out. For example, it is legitimate to point out that an improvement in the quality of generative models could be used to generate deepfakes for disinformation. On the other hand, it is not needed to point out that a generic algorithm for optimizing neural networks could enable people to train models that generate Deepfakes faster.
        \item The authors should consider possible harms that could arise when the technology is being used as intended and functioning correctly, harms that could arise when the technology is being used as intended but gives incorrect results, and harms following from (intentional or unintentional) misuse of the technology.
        \item If there are negative societal impacts, the authors could also discuss possible mitigation strategies (e.g., gated release of models, providing defenses in addition to attacks, mechanisms for monitoring misuse, mechanisms to monitor how a system learns from feedback over time, improving the efficiency and accessibility of ML).
    \end{itemize}
    
\item {\bf Safeguards}
    \item[] Question: Does the paper describe safeguards that have been put in place for responsible release of data or models that have a high risk for misuse (e.g., pretrained language models, image generators, or scraped datasets)?
    \item[] Answer: \answerNo{} 
    \item[] Justification: There is no new data and much larger models produce higher fidelity images. The cost of training these large GANs is not prohibitive and is often done by hobbyists. As such, it is doubtful that these models will unlock any \emph{new} capabilities for mis-use or dual-use.
    \item[] Guidelines:
    \begin{itemize}
        \item The answer NA means that the paper poses no such risks.
        \item Released models that have a high risk for misuse or dual-use should be released with necessary safeguards to allow for controlled use of the model, for example by requiring that users adhere to usage guidelines or restrictions to access the model or implementing safety filters. 
        \item Datasets that have been scraped from the Internet could pose safety risks. The authors should describe how they avoided releasing unsafe images.
        \item We recognize that providing effective safeguards is challenging, and many papers do not require this, but we encourage authors to take this into account and make a best faith effort.
    \end{itemize}

\item {\bf Licenses for existing assets}
    \item[] Question: Are the creators or original owners of assets (e.g., code, data, models), used in the paper, properly credited and are the license and terms of use explicitly mentioned and properly respected?
    \item[] Answer: \answerYes{} 
    \item[] Justification: All datasets are cited.
    \item[] Guidelines:
    \begin{itemize}
        \item The answer NA means that the paper does not use existing assets.
        \item The authors should cite the original paper that produced the code package or dataset.
        \item The authors should state which version of the asset is used and, if possible, include a URL.
        \item The name of the license (e.g., CC-BY 4.0) should be included for each asset.
        \item For scraped data from a particular source (e.g., website), the copyright and terms of service of that source should be provided.
        \item If assets are released, the license, copyright information, and terms of use in the package should be provided. For popular datasets, \url{paperswithcode.com/datasets} has curated licenses for some datasets. Their licensing guide can help determine the license of a dataset.
        \item For existing datasets that are re-packaged, both the original license and the license of the derived asset (if it has changed) should be provided.
        \item If this information is not available online, the authors are encouraged to reach out to the asset's creators.
    \end{itemize}

\item {\bf New Assets}
    \item[] Question: Are new assets introduced in the paper well documented and is the documentation provided alongside the assets?
    \item[] Answer: \answerNA{} 
    \item[] Justification: No new assets are released.
    \item[] Guidelines:
    \begin{itemize}
        \item The answer NA means that the paper does not release new assets.
        \item Researchers should communicate the details of the dataset/code/model as part of their submissions via structured templates. This includes details about training, license, limitations, etc. 
        \item The paper should discuss whether and how consent was obtained from people whose asset is used.
        \item At submission time, remember to anonymize your assets (if applicable). You can either create an anonymized URL or include an anonymized zip file.
    \end{itemize}

\item {\bf Crowdsourcing and Research with Human Subjects}
    \item[] Question: For crowdsourcing experiments and research with human subjects, does the paper include the full text of instructions given to participants and screenshots, if applicable, as well as details about compensation (if any)? 
    \item[] Answer: \answerNA{} 
    \item[] Justification: No human subjects are used and no crowdsourcing is used.
    \item[] Guidelines:
    \begin{itemize}
        \item The answer NA means that the paper does not involve crowdsourcing nor research with human subjects.
        \item Including this information in the supplemental material is fine, but if the main contribution of the paper involves human subjects, then as much detail as possible should be included in the main paper. 
        \item According to the NeurIPS Code of Ethics, workers involved in data collection, curation, or other labor should be paid at least the minimum wage in the country of the data collector. 
    \end{itemize}

\item {\bf Institutional Review Board (IRB) Approvals or Equivalent for Research with Human Subjects}
    \item[] Question: Does the paper describe potential risks incurred by study participants, whether such risks were disclosed to the subjects, and whether Institutional Review Board (IRB) approvals (or an equivalent approval/review based on the requirements of your country or institution) were obtained?
    \item[] Answer: \answerNA{} 
    \item[] Justification: No human subjects are used and no crowdsourcing is used.
    \item[] Guidelines:
    \begin{itemize}
        \item The answer NA means that the paper does not involve crowdsourcing nor research with human subjects.
        \item Depending on the country in which research is conducted, IRB approval (or equivalent) may be required for any human subjects research. If you obtained IRB approval, you should clearly state this in the paper. 
        \item We recognize that the procedures for this may vary significantly between institutions and locations, and we expect authors to adhere to the NeurIPS Code of Ethics and the guidelines for their institution. 
        \item For initial submissions, do not include any information that would break anonymity (if applicable), such as the institution conducting the review.
    \end{itemize}

\end{enumerate}


%% file: main.bbl
\begin{thebibliography}{101}
\providecommand{\natexlab}[1]{#1}
\providecommand{\url}[1]{\texttt{#1}}
\expandafter\ifx\csname urlstyle\endcsname\relax
  \providecommand{\doi}[1]{doi: #1}\else
  \providecommand{\doi}{doi: \begingroup \urlstyle{rm}\Url}\fi

\bibitem[Ba et~al.(2016)Ba, Kiros, and Hinton]{ln}
Jimmy~Lei Ba, Jamie~Ryan Kiros, and Geoffrey~E Hinton.
\newblock Layer normalization.
\newblock \emph{arXiv preprint arXiv:1607.06450}, 2016.

\bibitem[Biswas et~al.(2021)Biswas, Kumar, Banerjee, and Pandey]{smu}
Koushik Biswas, Sandeep Kumar, Shilpak Banerjee, and Ashish~Kumar Pandey.
\newblock Smu: smooth activation function for deep networks using smoothing maximum technique.
\newblock \emph{arXiv preprint arXiv:2111.04682}, 2021.

\bibitem[Brock et~al.(2018)Brock, Donahue, and Simonyan]{biggan}
Andrew Brock, Jeff Donahue, and Karen Simonyan.
\newblock Large scale gan training for high fidelity natural image synthesis.
\newblock \emph{arXiv preprint arXiv:1809.11096}, 2018.

\bibitem[Brock et~al.(2021)Brock, De, Smith, and Simonyan]{nfnet}
Andy Brock, Soham De, Samuel~L Smith, and Karen Simonyan.
\newblock High-performance large-scale image recognition without normalization.
\newblock In \emph{International Conference on Machine Learning}, pp.\  1059--1071. PMLR, 2021.

\bibitem[Chollet(2017)]{xception}
Fran{\c{c}}ois Chollet.
\newblock Xception: Deep learning with depthwise separable convolutions.
\newblock In \emph{Proceedings of the IEEE conference on computer vision and pattern recognition}, pp.\  1251--1258, 2017.

\bibitem[Chrabaszcz et~al.(2017)Chrabaszcz, Loshchilov, and Hutter]{chrabaszcz2017downsampled}
Patryk Chrabaszcz, Ilya Loshchilov, and Frank Hutter.
\newblock A downsampled variant of imagenet as an alternative to the cifar datasets.
\newblock \emph{arXiv preprint arXiv:1707.08819}, 2017.

\bibitem[Dhariwal \& Nichol(2021)Dhariwal and Nichol]{adm}
Prafulla Dhariwal and Alexander Nichol.
\newblock Diffusion models beat gans on image synthesis.
\newblock \emph{Advances in neural information processing systems}, 34:\penalty0 8780--8794, 2021.

\bibitem[Dieng et~al.(2019)Dieng, Ruiz, Blei, and Titsias]{presgan}
Adji~B Dieng, Francisco~JR Ruiz, David~M Blei, and Michalis~K Titsias.
\newblock Prescribed generative adversarial networks.
\newblock \emph{arXiv preprint arXiv:1910.04302}, 2019.

\bibitem[Dosovitskiy et~al.(2020)Dosovitskiy, Beyer, Kolesnikov, Weissenborn, Zhai, Unterthiner, Dehghani, Minderer, Heigold, Gelly, et~al.]{vit}
Alexey Dosovitskiy, Lucas Beyer, Alexander Kolesnikov, Dirk Weissenborn, Xiaohua Zhai, Thomas Unterthiner, Mostafa Dehghani, Matthias Minderer, Georg Heigold, Sylvain Gelly, et~al.
\newblock An image is worth 16x16 words: Transformers for image recognition at scale.
\newblock \emph{arXiv preprint arXiv:2010.11929}, 2020.

\bibitem[Fang et~al.(2022)Fang, Sun, and Schwing]{diggan}
Tiantian Fang, Ruoyu Sun, and Alex Schwing.
\newblock Dig{GAN}: Discriminator gradient gap regularization for {GAN} training with limited data.
\newblock In Alice~H. Oh, Alekh Agarwal, Danielle Belgrave, and Kyunghyun Cho (eds.), \emph{Advances in Neural Information Processing Systems}, 2022.
\newblock URL \url{https://openreview.net/forum?id=azBVn74t_2}.

\bibitem[Gidel et~al.(2019)Gidel, Hemmat, Pezeshki, Le~Priol, Huang, Lacoste-Julien, and Mitliagkas]{ganmomentum}
Gauthier Gidel, Reyhane~Askari Hemmat, Mohammad Pezeshki, R{\'e}mi Le~Priol, Gabriel Huang, Simon Lacoste-Julien, and Ioannis Mitliagkas.
\newblock Negative momentum for improved game dynamics.
\newblock In \emph{The 22nd International Conference on Artificial Intelligence and Statistics}, pp.\  1802--1811. PMLR, 2019.

\bibitem[Gokaslan et~al.(2024)Gokaslan, Cooper, Collins, Seguin, Jacobson, Patel, Frankle, Stephenson, and Kuleshov]{gokaslan2024commoncanvas}
Aaron Gokaslan, A~Feder Cooper, Jasmine Collins, Landan Seguin, Austin Jacobson, Mihir Patel, Jonathan Frankle, Cory Stephenson, and Volodymyr Kuleshov.
\newblock Commoncanvas: Open diffusion models trained on creative-commons images.
\newblock In \emph{Proceedings of the IEEE/CVF Conference on Computer Vision and Pattern Recognition}, pp.\  8250--8260, 2024.

\bibitem[Goodfellow et~al.(2020)Goodfellow, Pouget-Abadie, Mirza, Xu, Warde-Farley, Ozair, Courville, and Bengio]{gan}
Ian Goodfellow, Jean Pouget-Abadie, Mehdi Mirza, Bing Xu, David Warde-Farley, Sherjil Ozair, Aaron Courville, and Yoshua Bengio.
\newblock Generative adversarial networks.
\newblock \emph{Communications of the ACM}, 63\penalty0 (11):\penalty0 139--144, 2020.

\bibitem[Gulrajani et~al.(2017)Gulrajani, Ahmed, Arjovsky, Dumoulin, and Courville]{wgan-gp}
Ishaan Gulrajani, Faruk Ahmed, Martin Arjovsky, Vincent Dumoulin, and Aaron~C Courville.
\newblock Improved training of wasserstein gans.
\newblock \emph{Advances in neural information processing systems}, 30, 2017.

\bibitem[He et~al.(2015)He, Zhang, Ren, and Sun]{prelu}
Kaiming He, Xiangyu Zhang, Shaoqing Ren, and Jian Sun.
\newblock Delving deep into rectifiers: Surpassing human-level performance on imagenet classification.
\newblock In \emph{Proceedings of the IEEE international conference on computer vision}, pp.\  1026--1034, 2015.

\bibitem[He et~al.(2016{\natexlab{a}})He, Zhang, Ren, and Sun]{resnet}
Kaiming He, Xiangyu Zhang, Shaoqing Ren, and Jian Sun.
\newblock Deep residual learning for image recognition.
\newblock In \emph{Proceedings of the IEEE conference on computer vision and pattern recognition}, pp.\  770--778, 2016{\natexlab{a}}.

\bibitem[He et~al.(2016{\natexlab{b}})He, Zhang, Ren, and Sun]{resnet2}
Kaiming He, Xiangyu Zhang, Shaoqing Ren, and Jian Sun.
\newblock Identity mappings in deep residual networks.
\newblock In \emph{Computer Vision--ECCV 2016: 14th European Conference, Amsterdam, The Netherlands, October 11--14, 2016, Proceedings, Part IV 14}, pp.\  630--645. Springer, 2016{\natexlab{b}}.

\bibitem[Hendrycks \& Gimpel(2016)Hendrycks and Gimpel]{gelu}
Dan Hendrycks and Kevin Gimpel.
\newblock Gaussian error linear units (gelus).
\newblock \emph{arXiv preprint arXiv:1606.08415}, 2016.

\bibitem[Heusel et~al.(2017)Heusel, Ramsauer, Unterthiner, Nessler, and Hochreiter]{fid}
Martin Heusel, Hubert Ramsauer, Thomas Unterthiner, Bernhard Nessler, and Sepp Hochreiter.
\newblock Gans trained by a two time-scale update rule converge to a local nash equilibrium.
\newblock \emph{Advances in neural information processing systems}, 30, 2017.

\bibitem[Ho et~al.(2020)Ho, Jain, and Abbeel]{ddpm}
Jonathan Ho, Ajay Jain, and Pieter Abbeel.
\newblock Denoising diffusion probabilistic models.
\newblock \emph{Advances in neural information processing systems}, 33:\penalty0 6840--6851, 2020.

\bibitem[Ioffe \& Szegedy(2015)Ioffe and Szegedy]{bn}
Sergey Ioffe and Christian Szegedy.
\newblock Batch normalization: Accelerating deep network training by reducing internal covariate shift.
\newblock In \emph{International conference on machine learning}, pp.\  448--456. pmlr, 2015.

\bibitem[Jolicoeur-Martineau(2018)]{rgan}
Alexia Jolicoeur-Martineau.
\newblock The relativistic discriminator: a key element missing from standard gan.
\newblock \emph{arXiv preprint arXiv:1807.00734}, 2018.

\bibitem[Jolicoeur-Martineau \& Mitliagkas(2019)Jolicoeur-Martineau and Mitliagkas]{ganmmc}
Alexia Jolicoeur-Martineau and Ioannis Mitliagkas.
\newblock Gradient penalty from a maximum margin perspective.
\newblock \emph{arXiv preprint arXiv:1910.06922}, 2019.

\bibitem[Jolicoeur-Martineau et~al.(2020)Jolicoeur-Martineau, Pich{\'e}-Taillefer, Combes, and Mitliagkas]{advsm}
Alexia Jolicoeur-Martineau, R{\'e}mi Pich{\'e}-Taillefer, R{\'e}mi Tachet~des Combes, and Ioannis Mitliagkas.
\newblock Adversarial score matching and improved sampling for image generation.
\newblock \emph{arXiv preprint arXiv:2009.05475}, 2020.

\bibitem[Kang et~al.(2023{\natexlab{a}})Kang, Shin, and Park]{studio}
Minguk Kang, Joonghyuk Shin, and Jaesik Park.
\newblock Studiogan: a taxonomy and benchmark of gans for image synthesis.
\newblock \emph{IEEE Transactions on Pattern Analysis and Machine Intelligence}, 2023{\natexlab{a}}.

\bibitem[Kang et~al.(2023{\natexlab{b}})Kang, Zhu, Zhang, Park, Shechtman, Paris, and Park]{gigagan}
Minguk Kang, Jun-Yan Zhu, Richard Zhang, Jaesik Park, Eli Shechtman, Sylvain Paris, and Taesung Park.
\newblock Scaling up gans for text-to-image synthesis.
\newblock In \emph{Proceedings of the IEEE/CVF Conference on Computer Vision and Pattern Recognition}, pp.\  10124--10134, 2023{\natexlab{b}}.

\bibitem[Karnewar \& Wang(2020)Karnewar and Wang]{karnewar2020msg}
Animesh Karnewar and Oliver Wang.
\newblock Msg-gan: Multi-scale gradients for generative adversarial networks.
\newblock In \emph{Proceedings of the IEEE/CVF conference on computer vision and pattern recognition}, pp.\  7799--7808, 2020.

\bibitem[Karras et~al.(2017)Karras, Aila, Laine, and Lehtinen]{pggan}
Tero Karras, Timo Aila, Samuli Laine, and Jaakko Lehtinen.
\newblock Progressive growing of gans for improved quality, stability, and variation.
\newblock \emph{arXiv preprint arXiv:1710.10196}, 2017.

\bibitem[Karras et~al.(2019)Karras, Laine, and Aila]{sg1}
Tero Karras, Samuli Laine, and Timo Aila.
\newblock A style-based generator architecture for generative adversarial networks.
\newblock In \emph{Proceedings of the IEEE/CVF conference on computer vision and pattern recognition}, pp.\  4401--4410, 2019.

\bibitem[Karras et~al.(2020{\natexlab{a}})Karras, Aittala, Hellsten, Laine, Lehtinen, and Aila]{sg2ada}
Tero Karras, Miika Aittala, Janne Hellsten, Samuli Laine, Jaakko Lehtinen, and Timo Aila.
\newblock Training generative adversarial networks with limited data.
\newblock \emph{Advances in neural information processing systems}, 33:\penalty0 12104--12114, 2020{\natexlab{a}}.

\bibitem[Karras et~al.(2020{\natexlab{b}})Karras, Laine, Aittala, Hellsten, Lehtinen, and Aila]{sg2}
Tero Karras, Samuli Laine, Miika Aittala, Janne Hellsten, Jaakko Lehtinen, and Timo Aila.
\newblock Analyzing and improving the image quality of stylegan.
\newblock In \emph{Proceedings of the IEEE/CVF conference on computer vision and pattern recognition}, pp.\  8110--8119, 2020{\natexlab{b}}.

\bibitem[Karras et~al.(2021)Karras, Aittala, Laine, H{\"a}rk{\"o}nen, Hellsten, Lehtinen, and Aila]{sg3}
Tero Karras, Miika Aittala, Samuli Laine, Erik H{\"a}rk{\"o}nen, Janne Hellsten, Jaakko Lehtinen, and Timo Aila.
\newblock Alias-free generative adversarial networks.
\newblock \emph{Advances in Neural Information Processing Systems}, 34:\penalty0 852--863, 2021.

\bibitem[Karras et~al.(2022)Karras, Aittala, Aila, and Laine]{edm}
Tero Karras, Miika Aittala, Timo Aila, and Samuli Laine.
\newblock Elucidating the design space of diffusion-based generative models.
\newblock \emph{Advances in Neural Information Processing Systems}, 35:\penalty0 26565--26577, 2022.

\bibitem[Karras et~al.(2023)Karras, Aittala, Lehtinen, Hellsten, Aila, and Laine]{edm2}
Tero Karras, Miika Aittala, Jaakko Lehtinen, Janne Hellsten, Timo Aila, and Samuli Laine.
\newblock Analyzing and improving the training dynamics of diffusion models.
\newblock \emph{arXiv preprint arXiv:2312.02696}, 2023.

\bibitem[Kim et~al.(2021)Kim, Shin, Song, Kang, and Moon]{kim2021soft}
Dongjun Kim, Seungjae Shin, Kyungwoo Song, Wanmo Kang, and Il-Chul Moon.
\newblock Soft truncation: A universal training technique of score-based diffusion model for high precision score estimation.
\newblock \emph{arXiv preprint arXiv:2106.05527}, 2021.

\bibitem[Kingma et~al.(2021)Kingma, Salimans, Poole, and Ho]{kingma2021variational}
Diederik Kingma, Tim Salimans, Ben Poole, and Jonathan Ho.
\newblock Variational diffusion models.
\newblock \emph{Advances in neural information processing systems}, 34:\penalty0 21696--21707, 2021.

\bibitem[Krizhevsky et~al.(2009)Krizhevsky, Hinton, et~al.]{krizhevsky2009learning}
Alex Krizhevsky, Geoffrey Hinton, et~al.
\newblock Learning multiple layers of features from tiny images.
\newblock \emph{Thesis}, 2009.

\bibitem[Krizhevsky et~al.(2012)Krizhevsky, Sutskever, and Hinton]{alexnet}
Alex Krizhevsky, Ilya Sutskever, and Geoffrey~E Hinton.
\newblock Imagenet classification with deep convolutional neural networks.
\newblock In F.~Pereira, C.J. Burges, L.~Bottou, and K.Q. Weinberger (eds.), \emph{Advances in Neural Information Processing Systems}, volume~25. Curran Associates, Inc., 2012.
\newblock URL \url{https://proceedings.neurips.cc/paper_files/paper/2012/file/c399862d3b9d6b76c8436e924a68c45b-Paper.pdf}.

\bibitem[Kumar et~al.(2019)Kumar, Ozair, Goyal, Courville, and Bengio]{meg}
Rithesh Kumar, Sherjil Ozair, Anirudh Goyal, Aaron Courville, and Yoshua Bengio.
\newblock Maximum entropy generators for energy-based models.
\newblock \emph{arXiv preprint arXiv:1901.08508}, 2019.

\bibitem[Kynk{\"a}{\"a}nniemi et~al.(2019)Kynk{\"a}{\"a}nniemi, Karras, Laine, Lehtinen, and Aila]{precrecall}
Tuomas Kynk{\"a}{\"a}nniemi, Tero Karras, Samuli Laine, Jaakko Lehtinen, and Timo Aila.
\newblock Improved precision and recall metric for assessing generative models.
\newblock \emph{Advances in neural information processing systems}, 32, 2019.

\bibitem[Kynk{\"a}{\"a}nniemi et~al.(2022)Kynk{\"a}{\"a}nniemi, Karras, Aittala, Aila, and Lehtinen]{kynkaanniemi2022role}
Tuomas Kynk{\"a}{\"a}nniemi, Tero Karras, Miika Aittala, Timo Aila, and Jaakko Lehtinen.
\newblock The role of imagenet classes in fr{\'e}chet inception distance.
\newblock \emph{arXiv preprint arXiv:2203.06026}, 2022.

\bibitem[Lee et~al.(2021)Lee, Chang, Jiang, Zhang, Tu, and Liu]{vitgan}
Kwonjoon Lee, Huiwen Chang, Lu~Jiang, Han Zhang, Zhuowen Tu, and Ce~Liu.
\newblock Vitgan: Training gans with vision transformers.
\newblock \emph{arXiv preprint arXiv:2107.04589}, 2021.

\bibitem[Lim et~al.(2017)Lim, Son, Kim, Nah, and Mu~Lee]{edsr}
Bee Lim, Sanghyun Son, Heewon Kim, Seungjun Nah, and Kyoung Mu~Lee.
\newblock Enhanced deep residual networks for single image super-resolution.
\newblock In \emph{Proceedings of the IEEE conference on computer vision and pattern recognition workshops}, pp.\  136--144, 2017.

\bibitem[Lim \& Ye(2017)Lim and Ye]{hingegan}
Jae~Hyun Lim and Jong~Chul Ye.
\newblock Geometric gan.
\newblock \emph{arXiv preprint arXiv:1705.02894}, 2017.

\bibitem[Lin et~al.(2021)Lin, Zhang, Ganz, Han, and Zhu]{anycostgan}
Ji~Lin, Richard Zhang, Frieder Ganz, Song Han, and Jun-Yan Zhu.
\newblock Anycost gans for interactive image synthesis and editing.
\newblock In \emph{Proceedings of the IEEE/CVF Conference on Computer Vision and Pattern Recognition}, pp.\  14986--14996, 2021.

\bibitem[Lin et~al.(2018)Lin, Khetan, Fanti, and Oh]{pacgan}
Zinan Lin, Ashish Khetan, Giulia Fanti, and Sewoong Oh.
\newblock Pacgan: The power of two samples in generative adversarial networks.
\newblock \emph{Advances in neural information processing systems}, 31, 2018.

\bibitem[Liu et~al.(2021)Liu, Lin, Cao, Hu, Wei, Zhang, Lin, and Guo]{swin}
Ze~Liu, Yutong Lin, Yue Cao, Han Hu, Yixuan Wei, Zheng Zhang, Stephen Lin, and Baining Guo.
\newblock Swin transformer: Hierarchical vision transformer using shifted windows.
\newblock In \emph{Proceedings of the IEEE/CVF international conference on computer vision}, pp.\  10012--10022, 2021.

\bibitem[Liu et~al.(2022)Liu, Mao, Wu, Feichtenhofer, Darrell, and Xie]{convnext}
Zhuang Liu, Hanzi Mao, Chao-Yuan Wu, Christoph Feichtenhofer, Trevor Darrell, and Saining Xie.
\newblock A convnet for the 2020s.
\newblock In \emph{Proceedings of the IEEE/CVF Conference on Computer Vision and Pattern Recognition}, pp.\  11976--11986, 2022.

\bibitem[Lu et~al.(2023)Lu, Poppe, et~al.]{compdiff}
Hui Lu, Ronald Poppe, et~al.
\newblock Compensation sampling for improved convergence in diffusion models.
\newblock \emph{arXiv preprint arXiv:2312.06285}, 2023.

\bibitem[Mao et~al.(2017)Mao, Li, Xie, Lau, Wang, and Paul~Smolley]{lsgan}
Xudong Mao, Qing Li, Haoran Xie, Raymond~YK Lau, Zhen Wang, and Stephen Paul~Smolley.
\newblock Least squares generative adversarial networks.
\newblock In \emph{Proceedings of the IEEE international conference on computer vision}, pp.\  2794--2802, 2017.

\bibitem[Mescheder et~al.(2017)Mescheder, Nowozin, and Geiger]{gannum}
Lars Mescheder, Sebastian Nowozin, and Andreas Geiger.
\newblock The numerics of gans.
\newblock \emph{Advances in neural information processing systems}, 30, 2017.

\bibitem[Mescheder et~al.(2018)Mescheder, Geiger, and Nowozin]{r1}
Lars Mescheder, Andreas Geiger, and Sebastian Nowozin.
\newblock Which training methods for gans do actually converge?
\newblock In \emph{International conference on machine learning}, pp.\  3481--3490. PMLR, 2018.

\bibitem[Metz et~al.(2016)Metz, Poole, Pfau, and Sohl-Dickstein]{metz2016unrolled}
Luke Metz, Ben Poole, David Pfau, and Jascha Sohl-Dickstein.
\newblock Unrolled generative adversarial networks.
\newblock In \emph{International Conference on Learning Representations}, 2016.

\bibitem[Miyato \& Koyama(2018)Miyato and Koyama]{cgans}
Takeru Miyato and Masanori Koyama.
\newblock cgans with projection discriminator.
\newblock \emph{arXiv preprint arXiv:1802.05637}, 2018.

\bibitem[Nagarajan \& Kolter(2017)Nagarajan and Kolter]{nagarajan2017gradient}
Vaishnavh Nagarajan and J~Zico Kolter.
\newblock Gradient descent gan optimization is locally stable.
\newblock \emph{Advances in neural information processing systems}, 30, 2017.

\bibitem[Ning et~al.(2023)Ning, Sangineto, Porrello, Calderara, and Cucchiara]{ning2023input}
Mang Ning, Enver Sangineto, Angelo Porrello, Simone Calderara, and Rita Cucchiara.
\newblock Input perturbation reduces exposure bias in diffusion models.
\newblock \emph{arXiv preprint arXiv:2301.11706}, 2023.

\bibitem[Nowozin et~al.(2016)Nowozin, Cseke, and Tomioka]{nowozin2016f}
Sebastian Nowozin, Botond Cseke, and Ryota Tomioka.
\newblock f-gan: Training generative neural samplers using variational divergence minimization.
\newblock \emph{Advances in neural information processing systems}, 29, 2016.

\bibitem[Peebles \& Xie(2023)Peebles and Xie]{dit}
William Peebles and Saining Xie.
\newblock Scalable diffusion models with transformers.
\newblock In \emph{Proceedings of the IEEE/CVF International Conference on Computer Vision}, pp.\  4195--4205, 2023.

\bibitem[Preechakul et~al.(2022)Preechakul, Chatthee, Wizadwongsa, and Suwajanakorn]{diffae}
Konpat Preechakul, Nattanat Chatthee, Suttisak Wizadwongsa, and Supasorn Suwajanakorn.
\newblock Diffusion autoencoders: Toward a meaningful and decodable representation.
\newblock In \emph{Proceedings of the IEEE/CVF Conference on Computer Vision and Pattern Recognition (CVPR)}, pp.\  10619--10629, June 2022.

\bibitem[Radford et~al.(2015)Radford, Metz, and Chintala]{dcgan}
Alec Radford, Luke Metz, and Soumith Chintala.
\newblock Unsupervised representation learning with deep convolutional generative adversarial networks.
\newblock \emph{arXiv preprint arXiv:1511.06434}, 2015.

\bibitem[Ramachandran et~al.(2017)Ramachandran, Zoph, and Le]{swish}
Prajit Ramachandran, Barret Zoph, and Quoc~V Le.
\newblock Searching for activation functions.
\newblock \emph{arXiv preprint arXiv:1710.05941}, 2017.

\bibitem[Rombach et~al.(2022)Rombach, Blattmann, Lorenz, Esser, and Ommer]{rombach2022high}
Robin Rombach, Andreas Blattmann, Dominik Lorenz, Patrick Esser, and Bj{\"o}rn Ommer.
\newblock High-resolution image synthesis with latent diffusion models.
\newblock In \emph{Proceedings of the IEEE/CVF conference on computer vision and pattern recognition}, pp.\  10684--10695, 2022.

\bibitem[Ronneberger et~al.(2015)Ronneberger, Fischer, and Brox]{unet}
Olaf Ronneberger, Philipp Fischer, and Thomas Brox.
\newblock U-net: Convolutional networks for biomedical image segmentation.
\newblock In \emph{Medical image computing and computer-assisted intervention--MICCAI 2015: 18th international conference, Munich, Germany, October 5-9, 2015, proceedings, part III 18}, pp.\  234--241. Springer, 2015.

\bibitem[Roth et~al.(2017)Roth, Lucchi, Nowozin, and Hofmann]{r1r2}
Kevin Roth, Aurelien Lucchi, Sebastian Nowozin, and Thomas Hofmann.
\newblock Stabilizing training of generative adversarial networks through regularization.
\newblock \emph{Advances in neural information processing systems}, 30, 2017.

\bibitem[Sadat et~al.(2024)Sadat, Buhmann, Bradley, Hilliges, and Weber]{litevae}
Seyedmorteza Sadat, Jakob Buhmann, Derek Bradley, Otmar Hilliges, and Romann~M Weber.
\newblock Litevae: Lightweight and efficient variational autoencoders for latent diffusion models.
\newblock \emph{arXiv preprint arXiv:2405.14477}, 2024.

\bibitem[Sahoo et~al.(2023)Sahoo, Gokaslan, De~Sa, and Kuleshov]{sahoo2023diffusion}
Subham~Sekhar Sahoo, Aaron Gokaslan, Chris De~Sa, and Volodymyr Kuleshov.
\newblock Diffusion models with learned adaptive noise.
\newblock \emph{arXiv preprint arXiv:2312.13236}, 2023.

\bibitem[Sandler et~al.(2018)Sandler, Howard, Zhu, Zhmoginov, and Chen]{mobnet}
Mark Sandler, Andrew Howard, Menglong Zhu, Andrey Zhmoginov, and Liang-Chieh Chen.
\newblock Mobilenetv2: Inverted residuals and linear bottlenecks.
\newblock In \emph{Proceedings of the IEEE conference on computer vision and pattern recognition}, pp.\  4510--4520, 2018.

\bibitem[Sauer et~al.(2021)Sauer, Chitta, M{\"u}ller, and Geiger]{sauer2021projected}
Axel Sauer, Kashyap Chitta, Jens M{\"u}ller, and Andreas Geiger.
\newblock Projected gans converge faster.
\newblock \emph{Advances in Neural Information Processing Systems}, 34:\penalty0 17480--17492, 2021.

\bibitem[Sauer et~al.(2022)Sauer, Schwarz, and Geiger]{sgxl}
Axel Sauer, Katja Schwarz, and Andreas Geiger.
\newblock {StyleGAN-XL}: Scaling stylegan to large diverse datasets.
\newblock In \emph{ACM SIGGRAPH 2022 conference proceedings}, pp.\  1--10, 2022.

\bibitem[Sauer et~al.(2023)Sauer, Karras, Laine, Geiger, and Aila]{sg-t}
Axel Sauer, Tero Karras, Samuli Laine, Andreas Geiger, and Timo Aila.
\newblock Stylegan-t: Unlocking the power of gans for fast large-scale text-to-image synthesis.
\newblock In \emph{International conference on machine learning}, pp.\  30105--30118. PMLR, 2023.

\bibitem[Shi et~al.(2016{\natexlab{a}})Shi, Caballero, Husz{\'a}r, Totz, Aitken, Bishop, Rueckert, and Wang]{pixshuffle}
Wenzhe Shi, Jose Caballero, Ferenc Husz{\'a}r, Johannes Totz, Andrew~P Aitken, Rob Bishop, Daniel Rueckert, and Zehan Wang.
\newblock Real-time single image and video super-resolution using an efficient sub-pixel convolutional neural network.
\newblock In \emph{Proceedings of the IEEE conference on computer vision and pattern recognition}, pp.\  1874--1883, 2016{\natexlab{a}}.

\bibitem[Shi et~al.(2016{\natexlab{b}})Shi, Caballero, Husz{\'a}r, Totz, Aitken, Bishop, Rueckert, and Wang]{subpixel}
Wenzhe Shi, Jose Caballero, Ferenc Husz{\'a}r, Johannes Totz, Andrew~P Aitken, Rob Bishop, Daniel Rueckert, and Zehan Wang.
\newblock Real-time single image and video super-resolution using an efficient sub-pixel convolutional neural network.
\newblock In \emph{Proceedings of the IEEE conference on computer vision and pattern recognition}, pp.\  1874--1883, 2016{\natexlab{b}}.

\bibitem[Simonyan \& Zisserman(2014)Simonyan and Zisserman]{vgg}
Karen Simonyan and Andrew Zisserman.
\newblock Very deep convolutional networks for large-scale image recognition.
\newblock \emph{arXiv preprint arXiv:1409.1556}, 2014.

\bibitem[Singh et~al.(2023)Singh, Shukla, and Turaga]{singh2023polynomial}
Rajhans Singh, Ankita Shukla, and Pavan Turaga.
\newblock Polynomial implicit neural representations for large diverse datasets.
\newblock In \emph{Proceedings of the IEEE/CVF Conference on Computer Vision and Pattern Recognition}, pp.\  2041--2051, 2023.

\bibitem[S{\o}nderby et~al.(2016)S{\o}nderby, Caballero, Theis, Shi, and Husz{\'a}r]{instancenoise}
Casper~Kaae S{\o}nderby, Jose Caballero, Lucas Theis, Wenzhe Shi, and Ferenc Husz{\'a}r.
\newblock Amortised map inference for image super-resolution.
\newblock \emph{arXiv preprint arXiv:1610.04490}, 2016.

\bibitem[Song et~al.(2021)Song, Meng, and Ermon]{ddim}
Jiaming Song, Chenlin Meng, and Stefano Ermon.
\newblock Denoising diffusion implicit models.
\newblock In \emph{International Conference on Learning Representations}, 2021.

\bibitem[Song \& Dhariwal(2024)Song and Dhariwal]{icm}
Yang Song and Prafulla Dhariwal.
\newblock Improved techniques for training consistency models.
\newblock In \emph{The Twelfth International Conference on Learning Representations}, 2024.

\bibitem[Song et~al.(2020)Song, Sohl-Dickstein, Kingma, Kumar, Ermon, and Poole]{sde}
Yang Song, Jascha Sohl-Dickstein, Diederik~P Kingma, Abhishek Kumar, Stefano Ermon, and Ben Poole.
\newblock Score-based generative modeling through stochastic differential equations.
\newblock \emph{arXiv preprint arXiv:2011.13456}, 2020.

\bibitem[Song et~al.(2023)Song, Dhariwal, Chen, and Sutskever]{cm}
Yang Song, Prafulla Dhariwal, Mark Chen, and Ilya Sutskever.
\newblock Consistency models.
\newblock In \emph{International Conference on Machine Learning}, pp.\  32211--32252. PMLR, 2023.

\bibitem[Srivastava et~al.(2017)Srivastava, Valkov, Russell, Gutmann, and Sutton]{srivastava2017veegan}
Akash Srivastava, Lazar Valkov, Chris Russell, Michael~U Gutmann, and Charles Sutton.
\newblock Veegan: Reducing mode collapse in gans using implicit variational learning.
\newblock \emph{Advances in neural information processing systems}, 30, 2017.

\bibitem[Sun et~al.(2020)Sun, Fang, and Schwing]{rpgan}
Ruoyu Sun, Tiantian Fang, and Alexander Schwing.
\newblock Towards a better global loss landscape of gans.
\newblock \emph{Advances in Neural Information Processing Systems}, 33:\penalty0 10186--10198, 2020.

\bibitem[Takida et~al.(2024)Takida, Imaizumi, Shibuya, Lai, Uesaka, Murata, and Mitsufuji]{takida2024san}
Yuhta Takida, Masaaki Imaizumi, Takashi Shibuya, Chieh-Hsin Lai, Toshimitsu Uesaka, Naoki Murata, and Yuki Mitsufuji.
\newblock {SAN}: Inducing metrizability of {GAN} with discriminative normalized linear layer.
\newblock In \emph{The Twelfth International Conference on Learning Representations}, 2024.
\newblock URL \url{https://openreview.net/forum?id=eiF7TU1E8E}.

\bibitem[Tao \& Wang(2020)Tao and Wang]{r1gradexpcvpr}
Song Tao and Jia Wang.
\newblock Alleviation of gradient exploding in gans: Fake can be real.
\newblock In \emph{Proceedings of the IEEE/CVF conference on computer vision and pattern recognition}, pp.\  1191--1200, 2020.

\bibitem[Thanh-Tung et~al.(2019)Thanh-Tung, Tran, and Venkatesh]{r1gradexp}
Hoang Thanh-Tung, Truyen Tran, and Svetha Venkatesh.
\newblock Improving generalization and stability of generative adversarial networks.
\newblock In \emph{International Conference on Learning Representations}, 2019.
\newblock URL \url{https://openreview.net/forum?id=ByxPYjC5KQ}.

\bibitem[Ulyanov et~al.(2016)Ulyanov, Vedaldi, and Lempitsky]{in}
Dmitry Ulyanov, Andrea Vedaldi, and Victor Lempitsky.
\newblock Instance normalization: The missing ingredient for fast stylization.
\newblock \emph{arXiv preprint arXiv:1607.08022}, 2016.

\bibitem[Vahdat et~al.(2021)Vahdat, Kreis, and Kautz]{lsgm}
Arash Vahdat, Karsten Kreis, and Jan Kautz.
\newblock Score-based generative modeling in latent space.
\newblock \emph{Advances in neural information processing systems}, 34:\penalty0 11287--11302, 2021.

\bibitem[Vaswani et~al.(2017)Vaswani, Shazeer, Parmar, Uszkoreit, Jones, Gomez, Kaiser, and Polosukhin]{trans}
Ashish Vaswani, Noam Shazeer, Niki Parmar, Jakob Uszkoreit, Llion Jones, Aidan~N Gomez, {\L}ukasz Kaiser, and Illia Polosukhin.
\newblock Attention is all you need.
\newblock \emph{Advances in neural information processing systems}, 30, 2017.

\bibitem[Wang et~al.(2018)Wang, Yu, Wu, Gu, Liu, Dong, Qiao, and Change~Loy]{esrgan}
Xintao Wang, Ke~Yu, Shixiang Wu, Jinjin Gu, Yihao Liu, Chao Dong, Yu~Qiao, and Chen Change~Loy.
\newblock Esrgan: Enhanced super-resolution generative adversarial networks.
\newblock In \emph{Proceedings of the European conference on computer vision (ECCV) workshops}, pp.\  0--0, 2018.

\bibitem[Wang et~al.(2023{\natexlab{a}})Wang, Schiff, Gokaslan, Pan, Wang, De~Sa, and Kuleshov]{wang2023infodiffusion}
Yingheng Wang, Yair Schiff, Aaron Gokaslan, Weishen Pan, Fei Wang, Christopher De~Sa, and Volodymyr Kuleshov.
\newblock Infodiffusion: Representation learning using information maximizing diffusion models.
\newblock In \emph{International Conference on Machine Learning}, pp.\  36336--36354. PMLR, 2023{\natexlab{a}}.

\bibitem[Wang et~al.(2023{\natexlab{b}})Wang, Zheng, He, Chen, and Zhou]{diffusiongan}
Zhendong Wang, Huangjie Zheng, Pengcheng He, Weizhu Chen, and Mingyuan Zhou.
\newblock Diffusion-gan: Training gans with diffusion.
\newblock In \emph{The Eleventh International Conference on Learning Representations}, 2023{\natexlab{b}}.

\bibitem[Wu \& He(2018{\natexlab{a}})Wu and He]{gn}
Yuxin Wu and Kaiming He.
\newblock Group normalization.
\newblock In \emph{Proceedings of the European conference on computer vision (ECCV)}, pp.\  3--19, 2018{\natexlab{a}}.

\bibitem[Wu \& He(2018{\natexlab{b}})Wu and He]{groupnorm}
Yuxin Wu and Kaiming He.
\newblock Group normalization.
\newblock In \emph{Proceedings of the European conference on computer vision (ECCV)}, pp.\  3--19, 2018{\natexlab{b}}.

\bibitem[Xiao et~al.(2020)Xiao, Kreis, Kautz, and Vahdat]{vaebm}
Zhisheng Xiao, Karsten Kreis, Jan Kautz, and Arash Vahdat.
\newblock Vaebm: A symbiosis between variational autoencoders and energy-based models.
\newblock \emph{arXiv preprint arXiv:2010.00654}, 2020.

\bibitem[Xiao et~al.(2021)Xiao, Kreis, and Vahdat]{ddgan}
Zhisheng Xiao, Karsten Kreis, and Arash Vahdat.
\newblock Tackling the generative learning trilemma with denoising diffusion gans.
\newblock \emph{arXiv preprint arXiv:2112.07804}, 2021.

\bibitem[Xie et~al.(2017)Xie, Girshick, Doll{\'a}r, Tu, and He]{resnext}
Saining Xie, Ross Girshick, Piotr Doll{\'a}r, Zhuowen Tu, and Kaiming He.
\newblock Aggregated residual transformations for deep neural networks.
\newblock In \emph{Proceedings of the IEEE conference on computer vision and pattern recognition}, pp.\  1492--1500, 2017.

\bibitem[Yin et~al.(2024)Yin, Gharbi, Zhang, Shechtman, Durand, Freeman, and Park]{dmd}
Tianwei Yin, Micha{\"e}l Gharbi, Richard Zhang, Eli Shechtman, Fredo Durand, William~T Freeman, and Taesung Park.
\newblock One-step diffusion with distribution matching distillation.
\newblock In \emph{Proceedings of the IEEE/CVF Conference on Computer Vision and Pattern Recognition}, pp.\  6613--6623, 2024.

\bibitem[Yu et~al.(2022)Yu, Luo, Zhou, Si, Zhou, Wang, Feng, and Yan]{metaformer}
Weihao Yu, Mi~Luo, Pan Zhou, Chenyang Si, Yichen Zhou, Xinchao Wang, Jiashi Feng, and Shuicheng Yan.
\newblock Metaformer is actually what you need for vision.
\newblock In \emph{Proceedings of the IEEE/CVF conference on computer vision and pattern recognition}, pp.\  10819--10829, 2022.

\bibitem[Zhang et~al.(2022)Zhang, Gu, Zhang, Bao, Chen, Wen, Wang, and Guo]{zhang2022styleswin}
Bowen Zhang, Shuyang Gu, Bo~Zhang, Jianmin Bao, Dong Chen, Fang Wen, Yong Wang, and Baining Guo.
\newblock Styleswin: Transformer-based gan for high-resolution image generation.
\newblock In \emph{Proceedings of the IEEE/CVF conference on computer vision and pattern recognition}, pp.\  11304--11314, 2022.

\bibitem[Zhang et~al.(2019)Zhang, Dauphin, and Ma]{fixup}
Hongyi Zhang, Yann~N Dauphin, and Tengyu Ma.
\newblock Fixup initialization: Residual learning without normalization.
\newblock \emph{arXiv preprint arXiv:1901.09321}, 2019.

\bibitem[Zhang(2019)]{blurpool}
Richard Zhang.
\newblock Making convolutional networks shift-invariant again.
\newblock In \emph{International conference on machine learning}, pp.\  7324--7334. PMLR, 2019.

\bibitem[Zhao et~al.(2021)Zhao, Singh, Lee, Zhang, Odena, and Zhang]{zhao2021improved}
Zhengli Zhao, Sameer Singh, Honglak Lee, Zizhao Zhang, Augustus Odena, and Han Zhang.
\newblock Improved consistency regularization for gans.
\newblock In \emph{Proceedings of the AAAI conference on artificial intelligence}, volume~35, pp.\  11033--11041, 2021.

\end{thebibliography}
